\documentclass{statsoc}
\usepackage[a4paper]{geometry}
\usepackage{graphicx}
\usepackage[textwidth=8em,textsize=small]{todonotes}
\usepackage{natbib}

\usepackage[utf8]{inputenc} 
\usepackage[T1]{fontenc}    
\usepackage{hyperref}       
\usepackage{url}            
\usepackage{booktabs}       
\usepackage{amsfonts}       
\usepackage{lipsum}
\usepackage{ntheorem} 
\usepackage{nicefrac}       
\usepackage{microtype}      
\usepackage{xcolor}         
\usepackage{setspace}
\usepackage{enumitem}
\setitemize{noitemsep,topsep=0pt,parsep=0pt,partopsep=0pt}
\setenumerate{noitemsep,topsep=0pt,parsep=0pt,partopsep=0pt}
\usepackage{floatrow}
\usepackage{caption}
\newfloatcommand{capbtabbox}{table}[][\FBwidth]
\usepackage[parfill]{parskip}
\usepackage{amsmath}
\usepackage{psfrag,epsf,color}
\usepackage{comment}
\RequirePackage{algorithm}
\RequirePackage{algorithmic}
\usepackage{amssymb}
\usepackage{multirow}
\usepackage{array}
\usepackage{bm}
\usepackage{booktabs}
\usepackage{float}
\usepackage{bbding}
\usepackage{pifont}
\newcommand{\xmark}{\ding{55}}
\usepackage{wasysym}
\usepackage{xcolor}
\definecolor{ao}{rgb}{0.0, 0.5, 0.0}
\usepackage{etoolbox}
\makeatletter
\patchcmd{\@makecaption}
  {\parbox}
  {\advance\@tempdima-\fontdimen2} 
  {}{}
\makeatother

\makeatletter
\renewcommand\hyper@natlinkbreak[2]{#1}
\makeatother

\newcommand{\Mean}{{\mathbb{E}}}

\def\independenT#1#2{\mathrel{\rlap{$#1#2$}\mkern2mu{#1#2}}}

\newcommand{\bB}{{\mathbb B}}
\newcommand{\bG}{{\mathbb G}}

\newcommand{\bbeta}{\bm{\beta}}

\newcommand{\cC}{{\mathcal C}}
\newcommand{\cN}{{\mathcal N}}

\newcommand{\bstheta}{\theta}

\newcommand{\bsone}{\boldsymbol{1}}

\usepackage{xargs}
\newcommandx{\ssum}[5][1=\widehat{C}_g, 2=\tau, 3 = , 4 = \tau, 5 = \widehat{C}_g]{\ensuremath{\frac{1}{#4|#5|} \sum \limits_{i\in #1}\sum\limits_{t=T-#2}^{T #3}
  }}
  
  \newcommandx{\ossum}[3][1=\widehat{\cC}_k, 2=\tau, 3 = ]{\ensuremath{ \sum \limits_{i\in #1}\sum\limits_{t=T-#2}^{#3}
  }}
  
  \newcommandx{\ossumo}[3][1=\cC_k, 2=T-\tau+1, 3 = T]{\ensuremath{ \sum \limits_{i\in #1}\sum\limits_{t=#2}^{#3}
  }}

\newcommandx{\cssum}[3][1=\widehat{\tau}_i, 2=\tau, 3 = ]{\ensuremath{\frac{1}{N} \sum \limits_{i=1}^N \frac{1}{#1}\sum\limits_{t=T-#2}^{T #3}
  }}  

\newcommandx{\cssumn}[3][1=\widehat{\tau}_i, 2=\tau, 3 = ]{\ensuremath{\sum \limits_{i=1}^N \frac{1}{#1}\sum\limits_{t=T-#2}^{T #3}
  }}

  \newcommandx{\ssumt}[5][1=\widehat{\cC}_k, 2=\tau, 3 = , 4 = \tau, 5 = \widehat{C}_g]{\ensuremath{\frac{1}{#4|#5|} \sum \limits_{i\in #1}\sum\limits_{t=#2}^{#3}
  }}
  
  \newcommandx{\ossumt}[3][1=\widehat{\cC}_k, 2=\tau, 3 = ]{\ensuremath{ \sum \limits_{i\in #1}\sum\limits_{t=#2}^{ #3}
  }}

\newcommandx{\cssumt}[3][1=\widehat{\tau}_i, 2=\tau, 3 = ]{\ensuremath{\frac{1}{N} \sum \limits_{i=1}^N \frac{1}{#1}\sum\limits_{t=#2}^{#3}
  }}

\newcommandx{\lc}[4][1 = \widehat{\cC}_k,2 =\tau, 3=, 4=\bstheta_g]{
\ensuremath{\ell(#4;#1, [T-#2, T #3])
}
}

\newcommandx{\lce}[4][1 = \widehat{\cC}_k,2 =\tau, 3=, 4=\bstheta_g]{
\ensuremath{
\ell^0(#4;#1,[T-#2,T#3])
}}

\newcommandx{\lcfirst}[4][1 = \widehat{\cC}_k,2 =T-\tau, 3=, 4=\bstheta_g]{
\ensuremath{
\ell^{'}(#4;#1, [T-#2, T #3])
}
}
\newcommandx{\lcsecond}[4][1 = \widehat{\cC}_k,2 =T-\tau, 3=, 4=\bstheta_g]{
\ensuremath{
\ell^{''}_{#1}(#4, D_{[T-#2, T #3]})
}
}

\newcommandx{\lct}[4][1 = \widehat{\cC}_k,2 =t_1, 3=t_2, 4=\bstheta]{
\ensuremath{\ell(#4;#1, [#2, #3])
}
}

\newcommandx{\lcte}[4][1 = \widehat{\cC}_k,2 =t_1, 3=t_2, 4=\bstheta]{
\ensuremath{
\ell_0(#4;#1,[#2,#3])
}}

\newcommandx{\lctfirst}[4][1 = \widehat{\cC}_k,2 =t_1, 3=t_2, 4=\bstheta_g]{
\ensuremath{
\ell^{'}(#4;#1, [#2, #3])
}
}
\newcommandx{\lctsecond}[4][1 = \widehat{\cC}_k,2 =t_1, 3=t_2, 4=\bstheta_g]{
\ensuremath{
\ell^{''}(#4;#1, [#2, #3])
}
}
\newcommandx{\chc}[1][1=C_g]{
\ensuremath{
|#1||\widehat{\cC}_k|^{-1}
}}

  \newcommand{\triple}{
  \ensuremath{S_{it}|S_{it-1},A_{it-1}
  }}
   \newcommandx{\truple}[1][1=t-1]{
  \ensuremath{S_{i#1},A_{i#1}
  }}
  
  \newcommand{\tk}{\ensuremath{\bstheta_k}}
\newcommand{\tkt}{\ensuremath{\bstheta_k^0}}
  
  \newcommandx{\btgit}{\ensuremath{\bstheta_{k_i^0
  }}}
  
    \newcommandx{\btgih}[1][1=\widehat{k}_i]{\ensuremath{\widehat{\bstheta}_{#1}}}
    \newcommandx{\tgi}{\ensuremath{\widehat{\bstheta}_{k_i}}}
    
\newcommandx{\btgh}[1][1=k]{\ensuremath{\widehat{\bstheta}_{#1}}}  
  
    \newcommandx{\btgt}[1][1=k]{\ensuremath{\bstheta^0_{#1}}}

    \newcommandx{\bsob}[1][1=\widehat{k}_{i}(\bstheta) \neq k_{i}^{0}\right]{\ensuremath{\bsone \{ #1\} }}
\usepackage{xr}
\newtheorem{thm}{Theorem}
\newtheorem{asmp}{Assumption}
\newtheorem*{asmp-non}{Assumption}
\newtheorem{coro}{Corollary}
\newtheorem{lemma}{Lemma}[section]

\DeclareMathOperator*{\argmax}{arg\,max}

\usepackage{silence}
\newcommand*\samethanks[1][\value{footnote}]{\footnotemark[#1]}
\title[Doubly Inhomogeneous Reinforcement Learning]{Doubly Inhomogeneous Reinforcement Learning}

\author[L. Hu, M. Li, C. Shi, Z. Wu, and P. Fryzlewicz]{Liyuan~Hu$^{a,\thanks{The first two authors contributed equally to this paper and are alphabetically ordered.}}$ , 
Mengbing~Li$^{b,\samethanks}$, 
Chengchun~Shi$^{a,\thanks{Address for correspondence: Chengchun Shi, LSE, Houghton Street, London WC2A 2AE}}$ , 
Zhenke~Wu$^b$,
Piotr~Fryzlewicz$^a$ \\ \\
$^a$London School of Economics and Political Science, London, UK \\
$^b$University of Michigan, Ann Arbor, USA 
\\ \\
}
\email{\texttt{l.hu11@lse.ac.uk}, \texttt{mengbing@umich.edu}, \texttt{c.shi7@lse.ac.uk}, \\ \texttt{zhenkewu@umich.edu}, \texttt{p.fryzlewicz@lse.ac.uk}}

\usepackage{xr}
\makeatletter
\newcommand*{\addFileDependency}[1]{
  \typeout{(#1)}
  \@addtofilelist{#1}
  \IfFileExists{#1}{}{\typeout{No file #1.}}
}
\makeatother
\newcommand*{\myexternaldocument}[1]{
    \externaldocument{#1}
    \addFileDependency{#1.tex}
    \addFileDependency{#1.aux}
}
\myexternaldocument{supplement-cs}

\begin{document}
\begin{abstract}
This paper studies reinforcement learning (RL) in doubly inhomogeneous environments defined by temporal non-stationarity and subject heterogeneity, which often challenge high-quality sequential decision making. We propose an algorithm to determine the ``best data rectangles'' that display similar dynamics over time and population to facilitate policy learning.
Our approach
alternates between change point detection and cluster identification,
and is
a general wrapper method that integrates
a wide range of off-the-shelf algorithms for clustering, change point detection and policy learning. Theoretical analysis shows that our method achieves faster convergence of the estimated change points and clusters along with lower regrets compared to traditional RL algorithms that ignore temporal non-stationarity or subject heterogeneity, or both. More importantly, under certain conditions, our method attains the ``oracle'' property that both the change point detection and clustering errors can be exactly zero as if the best data rectangles were known in advance. 
Empirically, we demonstrate the usefulness of our method through simulations and analysis of a mobile health dataset.
\end{abstract}

\section{Introduction} \label{sec:intro}
Reinforcement learning (RL) is a powerful machine learning technique to train an intelligent agent to make sequential decisions for maximizing the long-term rewards it receives \citep{sutton2018reinforcement}. It has gained considerable popularity and has become one of the most important frontiers in artificial intelligence with a number of successful applications in game playing  \citep{mnih2015human}, 
ridesharing \citep{qin2025reinforcement}, healthcare \citep{chakraborty2013statistical,kosorok2019precision}, training of large language models \citep{ouyang2022training},
among many others \citep{li2019reinforcement}. 
Many existing state-of-the-art RL algorithms require the following two fundamental assumptions: 1) temporal stationarity: the Markov transition function within each data trajectory does not experience temporal changes, and 2) subject homogeneity: all data trajectories share the same transition function,  possibly with synchronized temporal changes. 
While both assumptions are plausible in some applications (e.g., video games), they can be violated in other applications. For instance, non-stationarity has been encountered in traffic signal control \citep{padakandla2020reinforcement}, mobile health \citep{liao2020personalized}, neuroscience \citep{zhang2020mixed}, ridesharing \citep{wan2021} and infectious disease control \citep{jiang2022modelling}. 
Heterogeneity across different subjects is commonly seen in precision medicine \citep{wei2023adaptive}, mobile health \citep{hu2020personalized}, public health \citep{huling2018risk} and urgent care \citep{chen2024reinforcement}. 
Different from prior work in the RL literature that either investigated doubly homogeneous environments -- requiring both assumptions to hold -- or studied singly inhomogeneous environments with only temporal non-stationarity or subject heterogeneity, 
we consider doubly inhomogeneous environments where both assumptions can be simultaneously violated. 

Doubly inhomogeneous environments are pervasive in the emerging field of interventional mobile health (mHealth) which needs RL for improving intervention strategies. Customized apps coupled with wearable devices can track subjects' daily behaviors and psychological states and provide timely prompts designed to benefit users. The Intern Health Study \citep[IHS;][]{necamp2020assessing} is one such example focused on the cohort of first-year training physicians (``medical interns'') in the United States. These medical interns often face difficult medical decisions and have challenging shift schedules to maintain regular exercise and sleep habits. 
To promote their general well-being, they are encouraged to install an IHS app that provides tips and insights on physical activity, sleep, and mood. 
Temporal non-stationarity is common in such interventional mHealth applications. User habituation to the prompts may occur, resulting in subjects being less responsive to the contents of the suggestions as the internship continues, calling for treatment policy adaptation over time \citep{klasnja2019efficacy}. Additionally, potential between-subject heterogeneity calls for optimal interventional policies that may be different across subjects rather than a single optimal policy. In the context of IHS, training physicians work under a wide range of specialties, shift schedules, and workloads. This 
not only causes between-subject differences in their daily physical activity, sleep and mood, but also potentially different effects of mobile prompts between the subjects.
Failure to recognise such double inhomogeneity in interventional mHealth may lead to ineffective policies that overburden users, which may result in disengagement from any potential interventions or even app deletion \citep{nahum2022engagement}.

This paper focuses on developing RL algorithms in doubly inhomogeneous environments. Throughout this paper, \textit{non-stationarity} refers to the inhomogeneity over time whereas \textit{heterogeneity} refers to the inhomogeneity across different subjects. In addition, \textit{double inhomogeneity} refers to both types of inhomogeneities: over time and population. We focus on offline estimation from a retrospective dataset, which is essential in various real applications  \citep{levine2020offline,jin2021pessimism}. Specifically, 
given a historical dataset from multiple subjects up to certain time, we aim to learn an optimal ``warm-up'' policy, i.e., a policy 
that recommends future actions for these subjects to maximize their long-term rewards, until their Markov transition functions change. Meanwhile, such an offline learning procedure can be sequentially applied to update the warm-up policy at pre-specified time points 
as data accumulate for online policy learning \citep{luckett2020estimating}. 

Policy learning is challenging in the presence of both inhomogeneities. The major challenge lies in effectively borrowing information over time and population. 
On the one hand, due to non-stationarity, it is desirable to use more recent observations to learn the warm-up policy rather than relying on all the past data \citep{padakandla2020reinforcement,alegre2021minimum,wei2021non,li2022TestingStationarity,wang2023robust}.  
On the other hand, due to subject heterogeneity, it is sub-optimal to apply a one-size-fits-all policy to all subjects for action recommendation \citep{shi2018maximin,chen2024reinforcement}. A straightforward approach to addressing double inhomogeneity is to develop subject- and/or time-specific policies based on each data trajectory or data at each given time. 
However, the resulting estimated optimal policy 
can be very noisy due to the limited sample size. 

\subsection{Contributions}
We summarise our contributions as follows:
\begin{enumerate}
    \item \textbf{Methodologically}, we propose an algorithm for determining the ``best data rectangles'' -- collections of largest possible rectangles consisting of two-dimensional data points indexed over time and population in which the transition dynamics are consistent across all time points and subjects (see Figure \ref{fig:dynamics_composite} for illustrations). This approach enables the application of any existing state-of-the-art RL algorithm designed for optimal policy learning in doubly homogeneous environments, to the estimated data rectangles for doubly inhomogeneous policy learning. To our knowledge, this is the first proposal in the statistics literature for policy learning in doubly inhomogeneous RL environments. It  leverages information over time and population, enhancing the efficiency of the estimated policy in the presence of double inhomogeneity. Meanwhile, our proposal demonstrates the power of classical statistical tools, such as change point detection and clustering, in addressing practical challenges in modern machine learning.

    \item \textbf{Theoretically}, we demonstrate that in doubly inhomogeneous environments, the proposed algorithm outperforms 
    both doubly homogeneous and singly inhomogeneous RL algorithms, 
    in terms of (i) reduced change point detection error, (ii) more accurate clustering, and (iii) lower policy regret (the difference in the expected cumulative reward between a learned policy and the optimal policy). We also prove the ``oracle'' property of the proposed algorithm, which shows that its regret is asymptotically the same as that of the oracle algorithm which works as if the best data rectangles were known.  
    
    \item \textbf{Empirically}, we show that by applying the optimal policy learned by our double-inhomogeneity-aware method, 
    medical interns can  achieve a 10\% increase in their average daily step counts. The learned policy can be combined with other types of prompts in the IHS study app to promote general well-being of the training physicians \citep{wang2023effectiveness}. When combined with an online batch learning scheme as illustrated in Section \ref{sec:sim:real:online}, our finding highlights the practical feasibility and utility of recognising double inhomogeneity in interventional mobile health studies.

\end{enumerate}

\subsection{Related work}
Existing policy learning or evaluation algorithms can be divided into three categories, depending on the environments in which they are used:  
\begin{enumerate}
    \item \textbf{Doubly homogeneous algorithms}: The first category studies doubly homogeneous environments, assuming both temporal stationarity and subject heterogeneity. As commented in the introduction, most RL algorithms fall into this category. Recently, a number of such algorithms have been proposed in the statistics literature \citep{ertefaie2018constructing,luckett2020estimating,liao2022batch,yang2022toward,chen2023steel,hu2023off,liu2023online,ramprasad2023online,wang2023projected,li2024settling,shi2024statistically,zhou2022estimating}.
    \item \textbf{Singly inhomogeneous algorithms}: The second category focuses on either non-stationarity \citep[see e.g.,][]{cheung2020reinforcement,fei2020dynamic,pmlr-v139-mao21b,wei2021non,zhong2021optimistic} or heterogeneity \cite[see e.g.,][]{hu2020personalized,chen2024reinforcement}, but not both. The latter class of heterogeneous learning algorithms is also related to a rich line of literature on transfer and federated learning in RL or bandits \citep[see e.g.,][]{zhu2020transfer,han2021federated,chen2022transferred,jin2022federated,yang2023federated}. While the aforementioned non-stationary (or heterogeneous) RL algorithms can be applied in a subject-wise (resp. time-wise) manner to accommodate double inhomogeneity, our theoretical study reveals that they tend to incur larger regret than the proposed double inhomogeneous algorithm (see Section \ref{sec:theory} for details).
    \item \textbf{Doubly inhomogeneous algorithms}: The third category allows the environment to be doubly inhomogeneous. In the literature, only a few recent studies have explored these settings, and both their focus and methodologies  differ from ours. For instance, \cite{tomkins2021intelligentpooling} and \cite{dwivedi2022counterfactual} proposed to employ classical random effects models to handle subject heterogeneity and temporal nonstationarity. They focused on bandit settings where the states are independent over time -- an assumption that is clearly violated in RL. More recently, \cite{bian2023off} investigated the RL setting but focused on policy evaluation. While policy evaluation serves as an intermediate step in policy learning algorithms such as policy iteration and actor-critic \citep{sutton2018reinforcement}, it does not directly address the problem of estimating the optimal policy.
\end{enumerate}
In addition to RL, our proposal is related to existing clustering \citep[see][for a recent review]{saxena2017review} and change point detection \citep[see e.g.,][]{friedrich2008complexity,cho2012multiscale,killick2012optimal,frick2014multiscale,fryzlewicz2014wild,wang2018high,li2022network,zhao2023high} algorithms, as it alternates between the two procedures to identify the best data rectangles for policy learning. Meanwhile, our theoretical and empirical analyses reveal that such an alternating algorithm is more statistically efficient than applying a clustering algorithm per time (Time-wise) or a change point detection algorithm per subject (Subject-wise) for handling double inhomogeneity, in the sense that it achieves smaller change point detection errors, clustering errors and regret (see Theorems \ref{thmrate} and \ref{thmregret} for detailed justifications). 
\subsection{Organisation of the paper}
The rest of the paper is structured as follows. Section \ref{sec:prelim} provides background on RL and describes doubly inhomogeneous environments. Section \ref{sec:method} presents our proposal for doubly inhomogeneous policy learning. Section \ref{sec:theory} establishes the theoretical properties of the proposed doubly inhomogeneous learning, showcases its advantages over traditional doubly homogeneous and singly inhomogeneous learning, and demonstrates its oracle property. Sections \ref{sec:simu} and \ref{sec:realdata} further illustrate the usefulness of our algorithm using simulations and real data analysis, respectively.

\section{Preliminaries: background on RL and double inhomogeneity}\label{sec:prelim}
We begin by describing the observed data, the concept of policy, and the standard stationarity and homogeneity assumptions in doubly homogeneous RL environments. We next provide detailed examples to illustrate doubly inhomogeneous environments.
\subsection{Data, policy, stationarity \& homogeneity} \label{sec:prelim:notation}
\textit{\textbf{Data}}. 
We focus on an offline setting  
given a pre-collected dataset $\{(S_{i,t},A_{i,t},R_{i,t}):1 \le i \le N, 0 \le t \le T\}$ where $(S_{i,t},A_{i,t},R_{i,t})$ denotes the state-action-reward triplet from the $i$th subject at time $t$, $N$ denotes the number of subjects, and $T$ denotes the termination time. These trajectories are assumed to be independent across different subjects. In our motivating IHS data application, the state are the covariates measured from the medical interns, including their daily average self-reported mood score, sleep minutes and step counts. The action is a binary treatment intervention applied to these interns, which determines whether or not to send a certain prompt to a given intern. The reward measures a given intern's outcome, 
which is the step count on the following day. 

The data generating process can be described as follows: \vspace{-0.5em}
\begin{itemize}
    \item At each time $t$, we measure the states from all subjects to obtain $\{S_{i,t}: 1\le i\le N\}$.
    \item Next, we assign each subject $i$ a treatment $A_{i,t}$.
    \item In response, each subject transits into a potentially different state $S_{i,t+1}$ at the next time point, and we observe their outcome $R_{i,t}$, which is a deterministic function of $(S_{i,t},A_{i,t},S_{i,t+1})$.
    \item This process repeats until we arrive at the termination time $T$.
\end{itemize}

\textit{\textbf{Policy}}. 
A policy determines how the decision maker assigns actions to each subject at each time. Formally speaking, a \textit{subject-specific history-dependent policy} $\pi$ is a list of decision rules $\{\pi_{i,t}\}_{i,t}$ such that each $\pi_{i,t}$ takes the $i$th subject's observed data history prior to time $t$ and $S_{i,t}$ as input (denoted by $\bar{S}_{i,t}$), and outputs a probability distribution on the actions space (denoted by $\pi_{i,t}(\bullet|\bar{S}_{i,t})$). When each $\pi_{i,t}$ depends on $\bar{S}_{i,t}$ only through $S_{i,t}$ and the dependence is stationary over time, we refer to the resulting policy as a \textit{stationary} policy. When $\pi_{i,t}$ is independent of $i$, we refer to the resulting policy as a \textit{homogeneous} policy. When $\pi_{i,t}$ is independent of both $i$ and $t$, we refer to the resulting policy as a \textit{doubly homogeneous} policy as it does not change over time or across subjects. 

\textit{\textbf{Double homogeneity}}. Let $p_{i,t}(\bullet|a,s)$ denote the probability density/mass function of $S_{i,t+1}$ given $(A_{i,t}=a, S_{i,t}=s)$. 
Under the following Markov assumption (MA), global stationarity assumption (GSA) and global homogeneity assumption (GHA), 
there exists an optimal doubly homogeneous policy that maximizes each subject's expected cumulative rewards \citep{Puterman1994}.\vspace{-0.5em}


\begin{asmp-non}[MA]
$S_{i,t+1}$ is independent of the past data history given $S_{i,t}$ and $A_{i,t}$ for any $i$, $t$.
\end{asmp-non}
\vspace{-1em}
\begin{asmp-non}[GSA]
$p_{i,t}$ is constant as a function of $t$.
\end{asmp-non}
\vspace{-1em}
\begin{asmp-non}[GHA] 
$p_{i,t}$ is constant as a function of $i$.
\end{asmp-non}
\vspace{-1em}

The above three assumptions are fundamental to RL and are commonly imposed in the literature \citep[see e.g.,][]{sutton2018reinforcement}. They substantially simplify the calculation of the optimal policy. 
When these assumptions hold, it suffices to search the optimal policy within the class of doubly homogeneous policies rather than the much broader class of subject-specific and/or history-dependent policies. Among these assumptions, MA is likely satisfied when we concatenate measurements over time to construct the state and can be tested based on the observed data \citep{chen2012testing,shi2020does,zhou2023testing}. 
The latter two assumptions -- GSA and GHA -- are required in traditional doubly homogeneous environments. Nonetheless, as commented earlier, both assumptions are often not met in mHealth. When either assumption is violated, the optimal policy is no longer doubly homogeneous and we summarise the form of the resulting optimal policy in Table \ref{tab:policy}. 

\begin{table}
\caption{\label{tab:policy}Forms of the optimal policy in different environments. Refer to Section \ref{sec:prelim} for details.}
\centering
\small
\vspace{-0.05in}
\begin{tabular}{|c|c|c|c|}
    \hline 
    GSA {\color{ao}{\checkmark}} GHA {\color{ao}{\checkmark}} & GSA {\color{ao}{\checkmark}} GHA {\color{red}\xmark} & GSA {\color{red}\xmark} GHA {\color{ao}{\checkmark}} & GSA {\color{red}\xmark} GHA {\color{red}\xmark} \\ 
    \hline
    doubly homogeneous & stationary & homogeneous & subject-specific history-dependent \\
    \hline
\end{tabular}
\end{table}

\subsection{Doubly inhomogeneous environments} \label{sec:prelim:environment}
This paper focuses on doubly inhomogeneous environments where both GSA and GHA are violated. Under MA, we assume that data can be grouped into clusters defined by their Markov transition function $p_{i,t}$. Specifically, any two data points $(i_1,t_1)$ and $(i_2,t_2)$ at different times $t_1\neq t_2$ or from different subjects $i_1\neq i_2$ belong to the same cluster if their transition functions are identical, i.e., $p_{i_1,t_1}=p_{i_2,t_2}$. Below, we provide some concrete examples to illustrate double inhomogeneity. 

\begin{figure}[t]
\centering
\begin{minipage}[c]{0.26\textwidth}
    \centering
    \includegraphics[width=\textwidth]{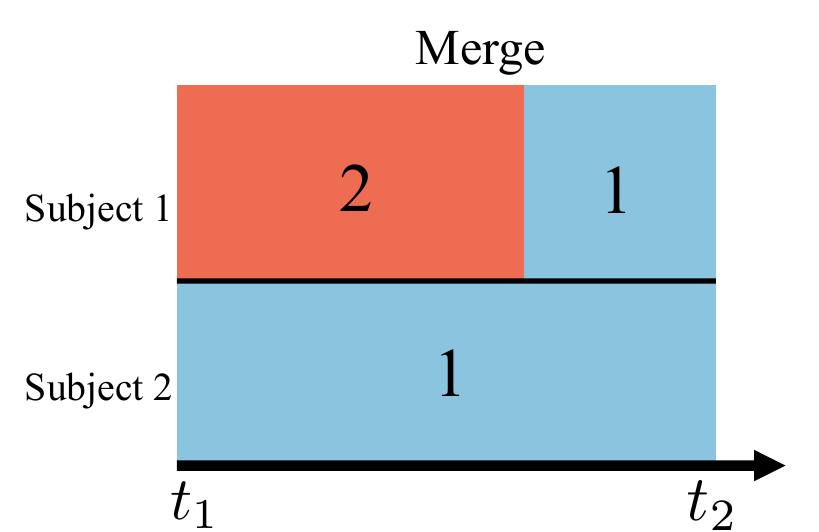} 
\end{minipage}
\begin{minipage}[c]{0.20\textwidth}
    \centering
    \includegraphics[width=\textwidth]{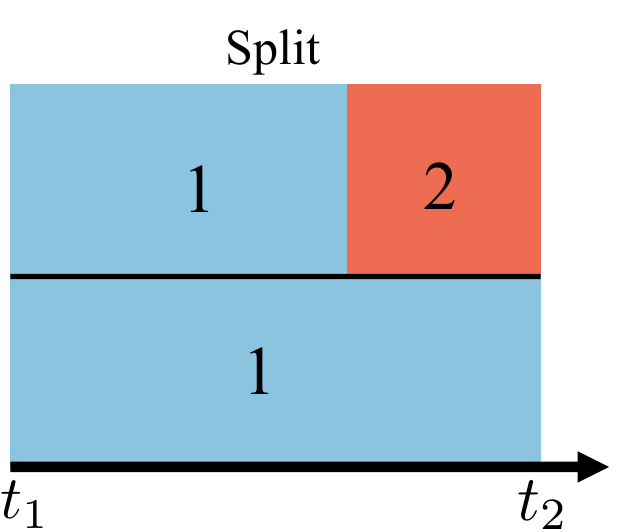} 
\end{minipage}
\begin{minipage}[c]{0.20\textwidth}
    \centering
    \includegraphics[width=\textwidth]{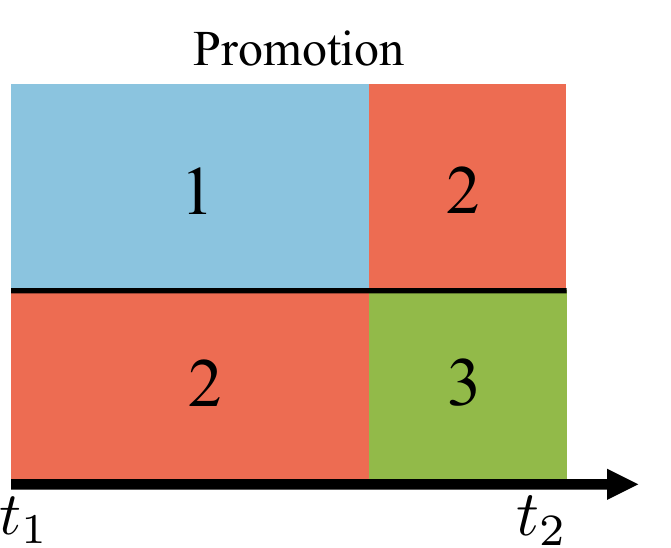} 
\end{minipage}
\begin{minipage}[c]{0.20\textwidth}
    \centering
    \includegraphics[width=\textwidth]{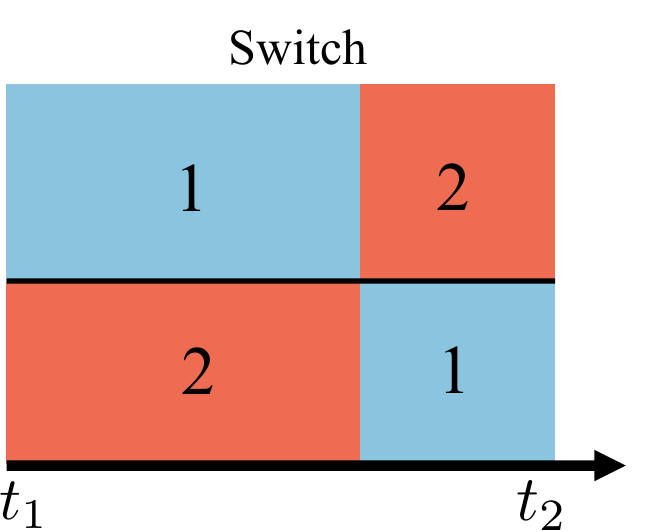} 
\end{minipage}
\begin{minipage}[c]{0.26\textwidth}
    \centering
    \includegraphics[width=\textwidth]{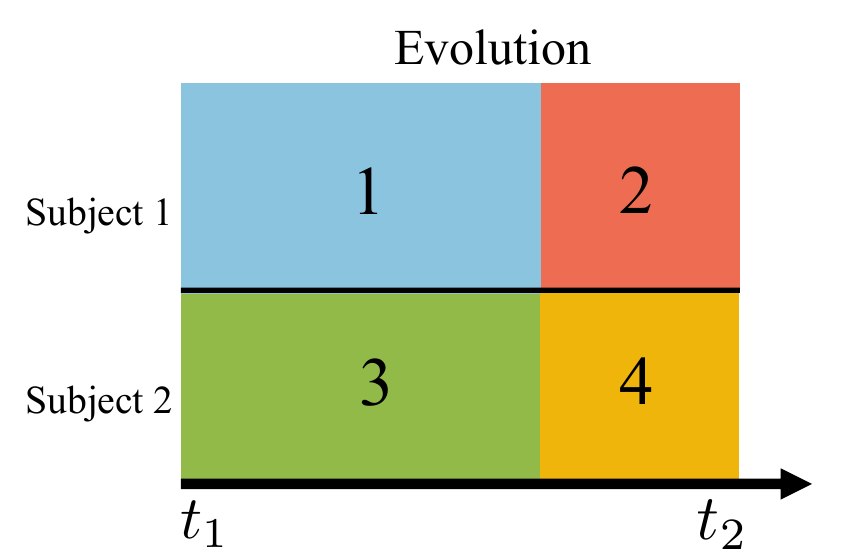} 
\end{minipage}
\begin{minipage}[c]{0.20\textwidth}
    \centering
    \includegraphics[width=\textwidth]{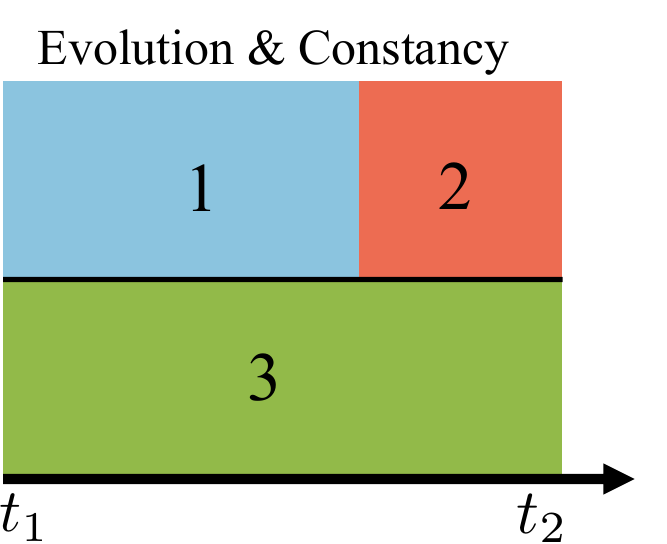} 
\end{minipage}
\begin{minipage}[c]{0.20\textwidth}
    \centering
    \includegraphics[width=\textwidth]{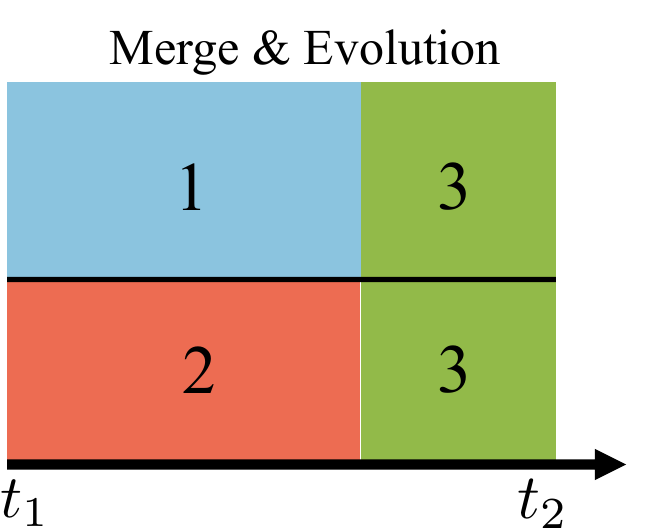} 
\end{minipage}
\begin{minipage}[c]{0.20\textwidth}
    \centering
    \includegraphics[width=\textwidth]{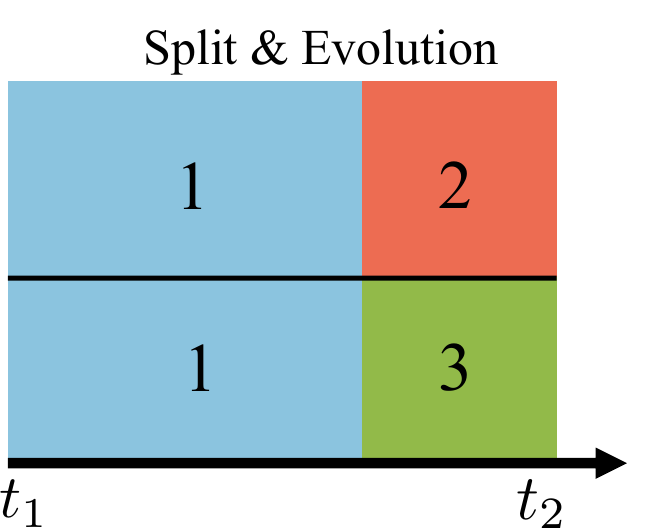} 
\end{minipage}
\caption{Eight basic building blocks of doubly inhomogeneous environments with two subjects (one in each row) and a single change point. Different transition functions are represented by different colours.}
\label{fig:dynamics_illustration}
\end{figure}

\begin{figure}[t]
    \centering 
    \includegraphics[width=0.6\textwidth]{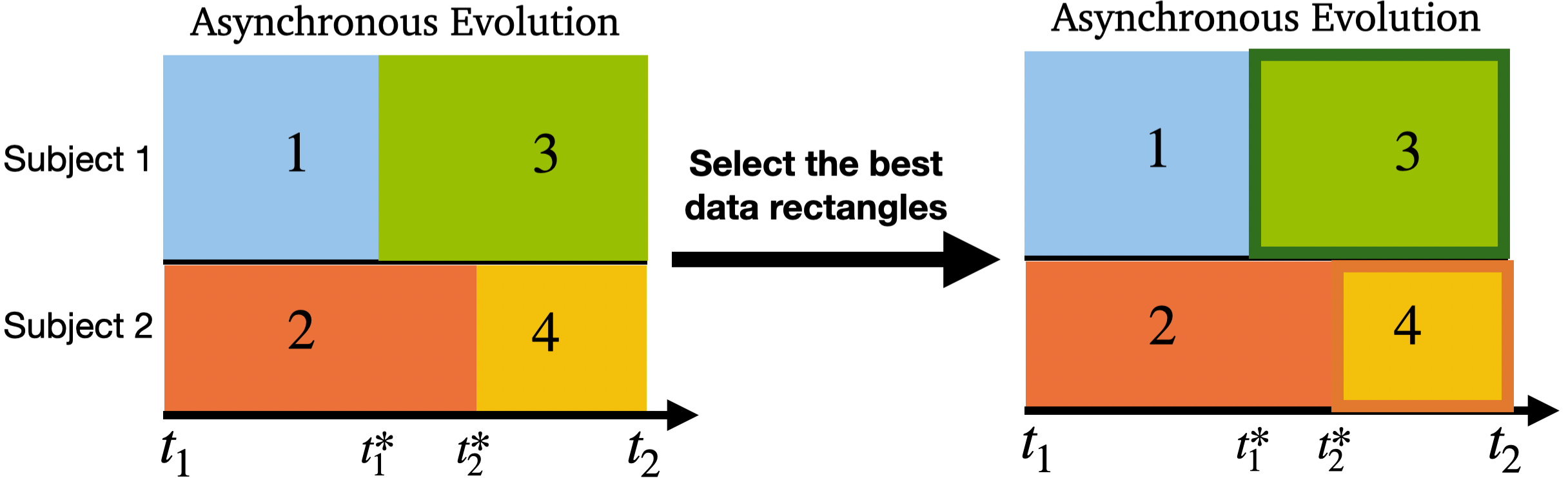} 
    \includegraphics[width=\textwidth]{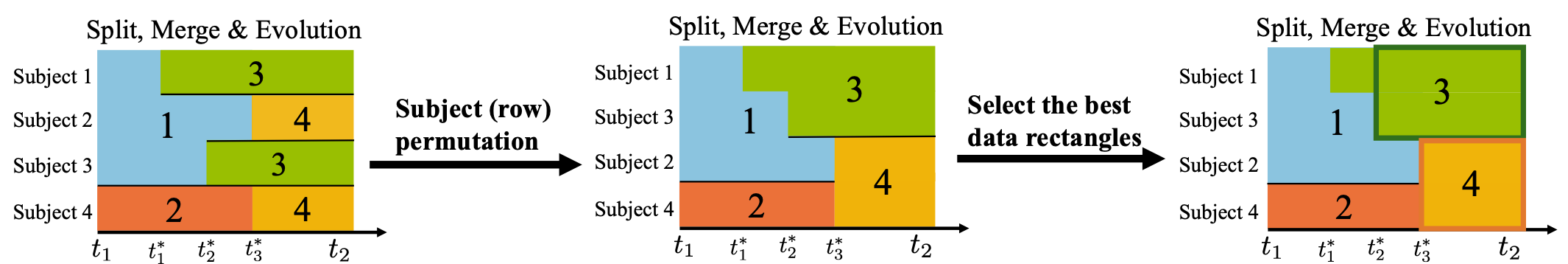}
    \vspace{-0.1in}
    \singlespacing
    \caption{Two additional doubly inhomogeneous environments. The top panel visualises an asynchronous evolution example where two subjects evolve at different time points asynchronously. The bottom panel visualises a split, merge and evolution example where an initial cluster first splits, then parts of it merge with another, evolving subsequently into a new one. In both panels, the best data rectangles with same dynamics over time and population are highlighted with bold borders.
    In particular in the bottom panel, subjects 1 and 3 evolve to a new shared dynamic at different time points and form a cluster. The best data rectangle of this cluster begins at the most recent change point $T - t_2^*$.
    }
    \label{fig:dynamics_composite}
\end{figure}

For simplicity, we consider the case with $N=2$ subjects. To account for non-stationarity, assume that there exists a single change point in the system such that either one or both subjects' transition functions can change at that time. This yields a total of 8 basic scenarios depicted in Figure \ref{fig:dynamics_illustration}, including merge, split, promotion, switch, evolution, and their combinations. 
These examples serve as the basic building blocks that can be used to construct more complicated scenarios with more than two subjects undergoing one or more change points. For instance, the asynchronous evolution example in the top panel of Figure \ref{fig:dynamics_composite} can be viewed as the composition of two separate ``Evolution \& Constancy'' configurations where two subjects evolve at time points $t_1^*$ and $t_2^*$ asynchronously. Meanwhile, the bottom panel of 
Figure \ref{fig:dynamics_composite} visualises a slightly more complicated example with four subjects. At the initial time $t_1$, each of the two clusters consists of one or more subjects sharing the same transition function. At time $t_1^*$, the first cluster splits into two, creating a third cluster. Next, the third subject in Cluster 1 merges with that in Cluster 3 at time $t_2^*$. Finally, the second subject merges with that in Cluster 2 at time $t_3^*$ and their merged transition function evolves further, resulting in a fourth cluster. 

In such environments, our objective lies in learning a warm-up policy for each of the $N$ subjects to maximize their long-term rewards starting from time $T$, until a further change gets detected. As demonstrated in our  numerical study, such a procedure can be repeatedly applied for updating the optimal policy at each current time as data accumulate. In the next section, we detail our methodology to identify the best data rectangles for efficient policy learning.


\section{Doubly inhomogeneous policy learning}\label{sec:method}
\subsection{Motivation for doubly inhomogeneous learning and limitations of existing algorithms}\label{sec:motivation}
Before describing the proposed algorithm, we present its motivation and discuss the limitations of existing doubly homogeneous and singly inhomogeneous algorithms in doubly inhomogeneous environments. We use the 2020 IHS study as a primary example. The dataset includes 1024 medical interns, collected over 83 days across 12 specialties. During this period, each subject was randomized daily to receive or not to receive activity suggestions. Daily mood scores were self-reported in the study app, while step count and sleep duration (in minutes) were tracked using wearables (Fitbit). We focus on optimizing policies to improve their long-term step counts. As commented in the introduction, user overburden or habituation to prompts, together with the variability among interns due to different specialties and workloads, results in a potentially doubly inhomogeneous environment. Meanwhile, analyses of the 2018 cohort have identified changes of dynamics over both time and population \citep{li2022TestingStationarity,wang2023robust}\footnote{Methodologically, these works assume subjects within each specialty share same dynamics to handle subject heterogeneity. That is, they group subjects into known clusters. We consider more general and realistic scenarios with unknown clusters.}, suggesting that similar double inhomogeneity may also exist in the 2020 cohort. 

\begin{figure}[t]
  \includegraphics[width=0.4\linewidth]{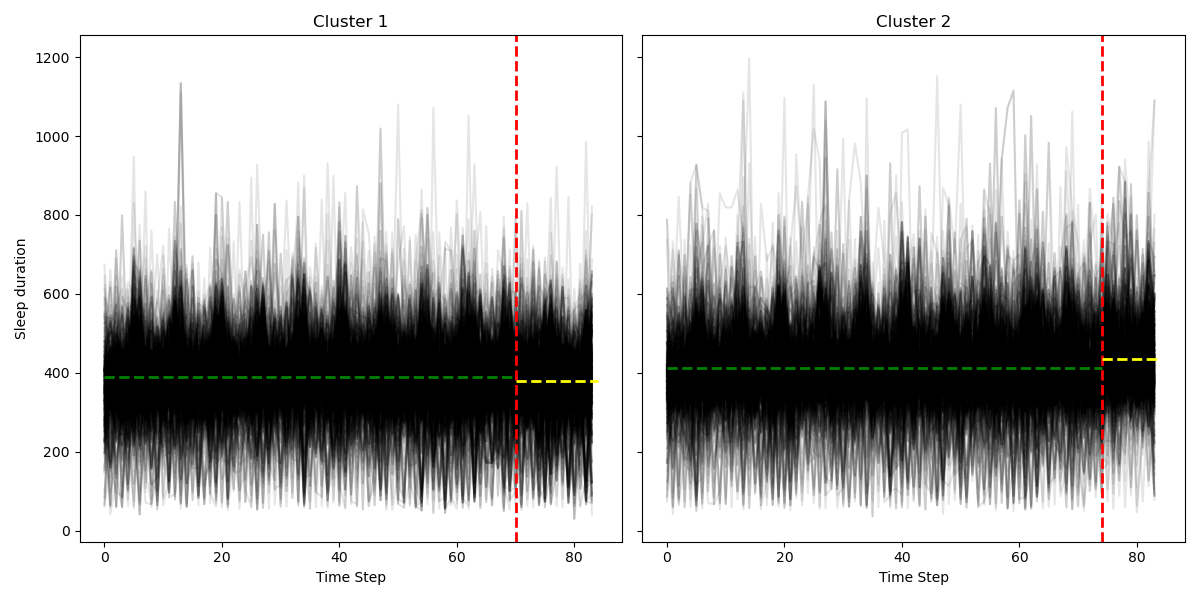}
\includegraphics[width=0.55\textwidth]{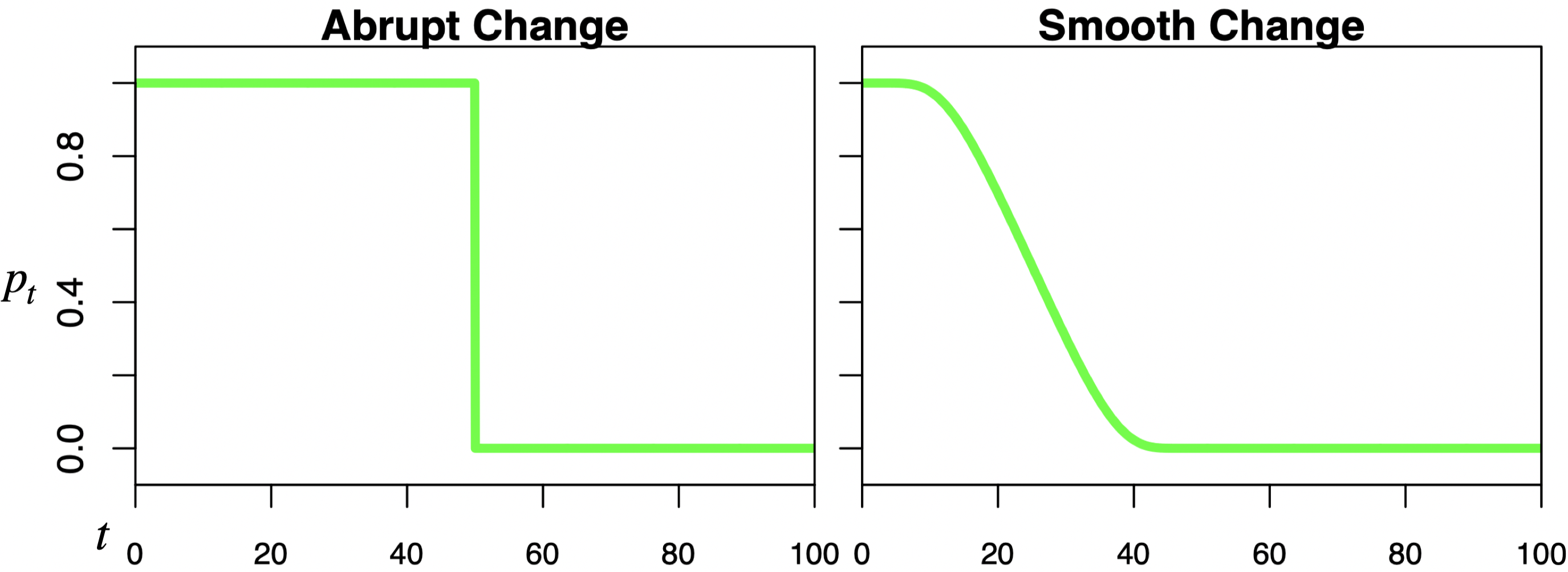} 
  \caption{Left panel: The individual sleep duration trajectories for the detected two clusters in the IHS dataset, along with their cluster-specific change points. The red vertical lines visualise the change points. The green and yellow horizontal lines report the cluster-specific average sleep duration before and after the change points. Right panel: Examples of abrupt and smooth change points occurring at 50 and 40, respectively.}%
  \label{fig:realdata:proposed_sleep_cpcluster}
\end{figure}

Our analysis confirms the existence of double inhomogeneity. Specifically, the proposed method identifies two clusters: the first one contains 520 subjects among whom the most recent change point occurred at day 70; the second one contains 504 subjects among whom the most recent change point occurred at day 74.
The left panel in Figure \ref{fig:realdata:proposed_sleep_cpcluster} visualises the individual sleep duration trajectories for the two clusters, along with their cluster-specific change points. 
An obvious decrease in the mean sleep duration for the first cluster happens after day 70, whereas the second cluster experiences an increase after day 74. 
We further fit two linear regression models to the data from the two clusters after their respective change points and find that they have different treatment effects. Specifically, the treatment effect is positive for the first cluster but negative for the second. These discrepancies suggest the existence of two groups of interns experiencing different changes around day 70. 

However, existing doubly homogeneous or singly inhomogeneous learning algorithms typically ignore at least one type of inhomogeneities inherent in the data, leading to sub-optimal policy learning. For instance, applying the clustering algorithm to the whole dataset assuming the stationary assumption (denoted by ``Stationary'') fails to identify any cluster structure, since the difference is likely washed out by aggregating all historical data over time. On the other hand, although the ``Homogeneous'' algorithm, which applies change point detection to the entire cohort, does identify a common change point at day 74 for all subjects, it ignores the different treatment effects and changes in sleep durations experienced by the two subgroups of interns. By merging the two clusters, these effects are likely averaged out. As reported in Table \ref{tab:realdata_opevalues}, overlooking these effects result in policies that perform 
worse than the one derived from the proposed doubly inhomogeneous learning.

\begin{table}[t]
\caption{
Average daily step counts 
under different policies.
}%
  \label{tab:realdata_opevalues}
  \begin{tabular}{|c|c|c|c|c|c|}
            \hline
            Policy & Proposed & Doubly Homogeneous & Homogeneous & Stationary & Behavior \\
            \hline
            Step counts&  
            8378.03 
            & 8333.25
            & 8308.32
            & 8333.25
            & 7596.30
            \\
            \hline
        \end{tabular}
\end{table}

\subsection{The main idea: identifying best data rectangles}
We present the idea of our proposal in this section. Motivated by our data analysis, we introduce the following conditions that form the basis of the proposed algorithm. Specifically, to relax global stationarity and homogeneity, we assume the environments are only ``locally'' stationary and homogeneous at the endpoint time $T$. 

\begin{asmp}[Local Stationarity at the Endpoint, LSE]\label{ass:LHE}
For each subject $i$, there exists a largest integer $\tau_i >0$, such that $p_{i,t}$ is a constant function of $t$ for any $T-\tau_i\le t\le T$. 
\end{asmp}
\begin{asmp}[Local Homogeneity at the Endpoint, LHE]\label{ass:LSE}
There exists a minimal finite number $K$ of disjoint subject clusters $\cup_{k=1}^K \mathcal{C}_k = \{1, \ldots, N\}$, where $\mathcal{C}_k \subseteq \{1, \ldots, N\}$, such that within each cluster $\mathcal{C}_k$, $p_{i,T}$ is a constant function of $i$.\vspace{-1em}
\end{asmp}
These two assumptions are likely to hold in our motivating IHS data example. As illustrated in Section \ref{sec:motivation}, there are likely two clusters at the endpoint where the subjects experienced changes at days 70 and 74, respectively. Both assumptions are mild and can be easily satisfied in general doubly inhomogeneous environments, as we elaborate below: 
\begin{itemize}
	\item Both assumptions are satisfied in all the examples introduced in Section \ref{sec:prelim:environment}.
	\item LSE allows the change at time $T-\tau_i$ to be either an abrupt change, or a smooth change (the transition function changes smoothly until certain time and remains stationary afterwards), as shown in the right panel of Figure \ref{fig:realdata:proposed_sleep_cpcluster}.
	\item Different from works in transfer and federated learning where data are typically classified into predefined clusters based on their attributes such as location \citep[see e.g.,][]{tang2016fused,li2022transfer}, LHE does not require the cluster membership to be known in advance, nor does it require the knowledge of the number of clusters $K$.
	\item We do not impose any homogeneity or stationarity assumption before the change point. For instance, for any two subjects that belong to the same most recent cluster, they are not required to share the same transition function prior to the change; see the ``Merge'' and ``Merge \& Evolution'' examples in Figure \ref{fig:dynamics_illustration}. 
\end{itemize}
As commented earlier, Assumptions \ref{ass:LHE} and \ref{ass:LSE} are fundamental to our proposal. Specifically, they imply that within the $k$-th cluster, $p_{i,t}$ is a constant function of $(i,t)$ for any $i\in \mathcal{C}_k, T-\tau_i\le t\le T$. Once $\{\mathcal{C}_k\}_k$ and $\{\tau_i\}_i$ are identified, we can apply existing doubly homogeneous RL algorithms such as fitted Q-iteration \citep[FQI,][]{Ernst2005,riedmiller2005} to the data subset $\{O_{i,t}=(S_{i,t},A_{i,t},R_{i,t},S_{i,t+1}): i\in \mathcal{C}_k, t\ge  T-\tau_i \}$ with the same transition function to derive the optimal warm-up policy for subjects within the $k$-th cluster. This motivates us to identify $\{\mathcal{C}_k\}_k$ and $\{\tau_{i}\}_i$ to borrow information to facilitate the policy learning. Nonetheless, this is challenging as subjects in the same most recent cluster do not necessarily share the same most recent change point. In other words, there is no guarantee that $\tau_{i}=\tau_{j}$ for any $i,j\in \mathcal{C}_k$. As a compromise, instead of identifying subject-specific change points $\{T-\tau_{i}: i\in \mathcal{C}_k\}$ within each cluster $\mathcal{C}_k$, we focus on estimating the most recent change point within $\mathcal{C}_k$, given by $T-\min \{\tau_{i}: i\in \mathcal{C}_k\}$. For an illustration, refer to the bottom panel of Figure \ref{fig:dynamics_composite}. In the third cluster, although Subjects 1 and 3 have different change points, our focus is on identifying the most recent one, corresponding to that of Subject 3. 
For any $k$ and $i\in \mathcal{C}_k$, let $\tau_i^*=\min \{\tau_{j}: j\in \mathcal{C}_k\}$. 
Observations fall into each region $\{(i,t): i\in \mathcal{C}_k, t\ge T-\tau_i^*\}$ share the same transition function and form a data rectangle over time and population (after a permutation of subjects).
Our objective is to identify these data rectangles, or equivalently, these $\mathcal{C}_k$s and $\tau_i^*$s. 
Given their estimators $\{\widehat{\tau}_i^*\}_i$ and $\{\widehat{\mathcal{C}}_k\}_k$, we apply existing doubly homogeneous RL algorithms to each estimated data rectangle to learn cluster-specific optimal policies for doubly inhomogeneous policy learning.

\subsection{The detailed algorithm}\label{sec:detailalg}
In this section, we detail the proposed algorithm, 
which is simple to describe. It consists of the following four steps:   \vspace{-0.3em}
\begin{itemize}
    \item \textbf{Input} Observation $\{(S_{i,t},A_{i,t},R_{i,t}):1 \le i \le N, 0 \le t \le T\}$, and initial estimators of $\{\tau_i^*\}_i$.
    
    \item \textbf{Step 1: Clustering}. Given the estimated most recent change points, cluster  subjects based on their recent observations following the change points.
    \item \textbf{Step 2: Most recent change point detection}. Given the estimated cluster memberships, update the most recent change point for each cluster.
    \item \textbf{Step 3: Iterating}. Alternate Steps 1 and 2 a few times to compute $\{\widehat{\tau}_i^{*}\}_i$ and $\{\widehat{\mathcal{C}}_k\}_k$.
    \item \textbf{Step 4: Policy learning}. Apply a base doubly homogeneous RL algorithm for policy learning to each data rectangle $\{O_{i,t}: i\in \widehat{\mathcal{C}}_k, t\ge T-\widehat{\tau}_i^*\}$ to compute an optimal policy $\widehat{\pi}_k$. 
    \item \textbf{Output} the warm-up policy $\widehat{\pi}$, which assigns $\widehat{\pi}_k$ to subjects within the $k$-th cluster $\widehat{\mathcal{C}}_k$. 
\end{itemize} 
In the clustering step, we further develop an information criterion to adaptively determine the number of clusters.  
Our algorithm is general in that we can plug-in any consistent most recent change point detection, clustering and policy learning algorithms designed for Markov decision processes \citep[MDPs,][]{Puterman1994} in Steps 1 -- 4. Below, we provide a concrete model-based method for change point detection and clustering by parameterising the transition function $p_{i,t}$ using some model $p(S'|A,S;\theta)$. Meanwhile, ``model-free'' clustering \citep{chen2024reinforcement} and change point detection algorithms \citep{li2022TestingStationarity} are equally applicable. These algorithms model the Q-function (the expected cumulative reward starting from a given initial state-action pair) and are able to consistently identify clusters or time intervals with different Q-functions. 

We denote the initial estimators of $\{\tau_i^*\}_i$ as $\{\tau_i^0\}_i$. 
Our theory (see Theorem \ref{thmrate} in Section \ref{sec:theory}) suggests that the error in estimating the data rectangles induced by these estimators are minimal, provided that their distances to the endpoint $T$ are no larger than those of the oracle change points. As a result, we recommend to use initial most recent change points that are close to $T$ to maintain stationarity within these intervals.  
We next elaborate three crucial components of our algorithm: most recent change point detection, clustering and the information criterion. 


\noindent \textit{\textbf{Most recent change point detection}}. For a given collection of cluster memberships $\{\widehat{\mathcal{C}}_k\}_k$, we aim to estimate $\min\{\tau_{j}:j\in \mathcal{C}_k\}$ for each $k$. Toward that end, we adopt a hypothesis testing framework for change point detection. For a given $\tau>1$, we aim to test the null hypothesis $\mathcal{H}_{0,\tau}^{(k)}: p_{i,t}$ is constant as a function of $t$ for any $i\in \mathcal{C}_k$ 
and any $t\in [T-\tau, T]$. To test $\mathcal{H}_{0,\tau}^{(k)}$, for a candidate change point location $u$, we divide the interval into two sub-intervals $[T-\tau, u)\cup [u,T]$ and test whether the transition function is constant across these two intervals. Specifically, we construct the following log-likelihood ratio test statistic $\textrm{LR}(\widehat{\mathcal{C}}_k, [T-\tau, T], u)$, given by
\begin{eqnarray*}
    \sum_{i\in |\widehat{\mathcal{C}}_k|} \sum_{t< u} \log \frac{p(S_{i,t}|A_{i,t-1},S_{i,t-1}, \widehat{\theta}_{\widehat{\mathcal{C}}_k,[T-\tau, T]})}{p(S_{i,t}|A_{i,t-1},S_{i,t-1}, \widehat{\theta}_{\widehat{\mathcal{C}}_k,[T-\tau, u-1]})}+\sum_{i\in |\widehat{\mathcal{C}}_k|}\sum_{t\ge u} \log \frac{p(S_{i,t}|A_{i,t-1},S_{i,t-1}, \widehat{\theta}_{\widehat{\mathcal{C}}_k,[T-\tau, T]})}{p(S_{i,t}|A_{i,t-1},S_{i,t-1}, \widehat{\theta}_{\widehat{\mathcal{C}}_k,[u, T]})},
\end{eqnarray*}
where $\widehat{\theta}_{\mathcal{C},[t_1,t_2]}$ denotes the estimated model parameter based on the data rectangle $\{O_{i,t}: i\in \mathcal{C}, t_1\le t\le t_2\}$. We next take its maximum over all possible candidate locations to construct the test statistic $\max_{T-\tau<u<T}\textrm{LR}(\widehat{\mathcal{C}}_k, [T-\tau, T], u)$ 
and reject the null if it exceeds a certain threshold. 
The choice of the threshold is discussed in detail in Section \ref{sec:sim:implementation:cp} of the Supplementary Materials. 
The resulting test can be viewed as a generalization of the classical Page-Lorden CUSUM test \citep{page1954continuous,lorden1971procedures} to MDPs. We sequentially apply such a procedure to test $\mathcal{H}_{0,\tau}^{(k)}$ for $\tau=\tau_0,\tau_0+1,\cdots$, until $\mathcal{H}_{0,\tau}^{(k)}$ gets rejected for some integer $\tau$. 
For any $i \in \widehat{\mathcal{C}}_k$, we set $\widehat{\tau}_i^*$ to the largest $\tau$ such that $\mathcal{H}_{0,\tau}^{(k)}$ is not rejected. Note that $\widehat{\tau}_i^*$'s are cluster-specific, i.e., $\widehat{\tau}_i^*=\widehat{\tau}_j^*$ for any $i,j\in \widehat{\mathcal{C}}_k$. 


\noindent \textit{\textbf{Clustering}}. For a given collection of $\{\widehat{\tau}_i^*\}_i$, 
we aim to update the cluster memberships $\{\widehat{\mathcal{C}}_k\}_k$ based on the set of 
most recent observations following these identified change points, i.e., $\{O_{i,t}: 1\le i\le N, t\ge T-\widehat{\tau}_i^*\}$. 
Specifically, given the number of clusters $K$, we compute $\{\widehat{\mathcal{C}}_{k}\}_k$ by solving the following optimisation:
\begin{eqnarray*}
    \{\widehat{\mathcal{C}}_k\}=\argmax_{\{\mathcal{C}_k\}_k} \sum_{k=1}^K \sum_{i\in \mathcal{C}_k} \frac{1}{\widehat{\tau}_i^*} \sum_{t=T-\widehat{\tau}_i^*+1}^T \log p(S_{i,t}|A_{i,t-1},S_{i,t-1}; \widehat{\theta}_{\mathcal{C}_k}),
\end{eqnarray*}
where $\widehat{\theta}_{\mathcal{C}_k}$ denotes the within-cluster maximum likelihood estimator for $\theta$, computed based on the data subset $\{O_{i,t}: i\in \mathcal{C}_k, t\ge T-\widehat{\tau}_i^*\}$. Notice that we maximise a weighted maximum likelihood function with the $i$-th weight inversely proportional to $\widehat{\tau}_i^*$. Such a scaling allows us to equally weight all subjects in the objective function so that 
the clustering errors are not dominated by subjects having much larger $\widehat{\tau}^*$'s than others.


\noindent \textit{\textbf{Information criterion}}. 
Since the number of clusters $K$ is unknown, we develop an information criterion (IC) to adaptively determine this hyper-parameter. Given a set of candidate numbers of clusters $\{K_j\}_j$, we first apply the clustering algorithm with each $K_j$ to learn the corresponding cluster memberships and then select the one that maximizes the following IC:
\begin{eqnarray}\label{eq:ic}
    \sum_{k=1}^{K_j} \sum_{i\in \mathcal{C}_k(K_j)} \frac{1}{\widehat{\tau}_i} \sum_{t=T-\widehat{\tau}_i+1}^T \log p(S_{i,t}|A_{i,t-1},S_{i,t-1}; \widehat{\theta}_{\mathcal{C}_k(K_j)}(K_j))-\frac{K_j N}{T}\log (NT).
\end{eqnarray}
Using similar arguments in proving the consistency of Bayesian information criterion \citep[BIC, see e.g.,][]{Schwarz1978}, we show that the proposed IC can consistently identify the number of clusters; see Theorem \ref{thmrate} in Section \ref{sec:theory} for details. Additionally, we focus on the most recent change point, 
regardless of the number of change points that have occurred in the past. Our IC is designed to determine the number of clusters, rather than the number of change points.

\section{Theory}\label{sec:theory}
We first analyse the estimated clusters and change points computed by Step 3 of our algorithm (see Section \ref{sec:detailalg}). We next establish the regret of our estimated optimal policy, defined as the difference in the expected cumulative reward between the optimal policy and the estimated one. Additionally, we analytically compare our proposal against the \textbf{Doubly Homogeneous} algorithm which ignores double inhomogeneity and the following singly inhomogeneous algorithms: 
\begin{enumerate}
    \item The \textbf{Homogeneous} algorithm applies change point detection to the whole cohort and the base RL algorithm (e.g., FQI) to observations following the detected change point for policy learning;
    \item The \textbf{Stationary} algorithm applies clustering to the whole cohort and the base RL algorithm to each cluster to learn cluster-specific policies;
    \item The \textbf{Subject-wise} algorithm applies change point detection to each individual trajectory and the base RL algorithm to observations following each individual change point to learn individual-specific policies;
    \item The \textbf{Time-wise} algorithm applies clustering to observations at the endpoint and the base RL algorithm to each cluster to learn cluster-specific policies.
\end{enumerate}
We also compare against an \textbf{Oracle} algorithm which works under the assumption that the change point locations and clusters are known in advance.
In Sections \ref{sec:simu} and \ref{sec:realdata}, we will contrast these algorithms empirically. 
\subsection{Clustering and change point detection errors}\label{sec:cpcluster}
\begin{table}
\caption{\label{tab:rate_convergence} Clustering and change point detection (CPD) errors when $N$ and $T$ have different divergence properties. 
The ``CPD error'' refers to the change point detection error
and ``non-negligible'' means that the error does not decay to zero as $NT\to \infty$. The errors shown are under high probabilities.}
\centering
\small
\begin{tabular}{lllll}
\toprule
Iteration                 &     Proposed                         & $T\rightarrow \infty$                   & $T\rightarrow \infty$                  & $T$ fixed                            \\
                          &                              & $N\rightarrow \infty$                   & $N$ fixed                              & $N\rightarrow \infty$                \\ \midrule
$1^{st}$                  & clustering error & $0$ & $0$   & non-negligible  \\
                          & CPD error             & $0$  & $ \displaystyle O\left(\frac{\log^3 (N T)}{  NT s_{cp}^{2}}\right)$ & non-negligible                           \\
\multirow{2}{*}{$2^{nd}$} &clustering error & $0$             & $0$            & non-negligible                        \\
                          & CPD error             & $ 0$  & $\displaystyle O\left(\frac{\log^3 (N T)}{  NT  s_{cp}^{2}}\right)$   & non-negligible                           \\
$\dots$                   &                              & $\dots$                                 & $\dots$                                & $\dots$                             \\ \bottomrule
\end{tabular}
\end{table}

Our theoretical findings can be summarised in Table \ref{tab:rate_convergence}.
First, we  show that, as the termination time $T\to\infty$, given a set of initial estimators $\{\tau_{i}^0\}_i$ that are sufficiently close to the endpoint $T$ in distance, 
the change point estimators achieve an optimal rate of convergence. Specifically, during each iteration, for sufficiently large $T$, the estimated most recent change points satisfy the following with probability at least $1-O(N^{-1}T^{-1})$:
\begin{equation}\label{eqn:cperr}
    \max_i \frac{|\widehat{\tau}_i^*-\tau_i^*|}{\tau_i^*}=O\left[\frac{\log^3(NT)}{NT s_{cp}^2}\right],
\end{equation}
where $s_{cp}$ denotes the signal strength, measuring the difference in the system dynamics before and after the change (see the detailed definition in  Section \ref{sec:notation} of the Supplementary Materials). 
When $N$ is finite, this rate is consistent with those in the change point detection literature up to logarithmic factors \citep[see e.g.,][]{cho2012multiscale,fryzlewicz2014wild,wang2018high}. To elaborate on this rate, we 
compare it against those attained by the \textbf{Subject-wise} and \textbf{Homogeneous} algorithms:
\begin{itemize}
    \item Similar analysis reveals that the Subject-wise algorithm achieves a change point detection error of the order $T^{-1}s_{cp}^{-2}$ up to logarithmic factors. This rate is much slower than \eqref{eqn:cperr} in settings with a large $N$, demonstrating the advantage of borrowing information across subjects. 
    \item Additionally, for sufficiently large $N$, the change point detection error can be smaller than $T^{-1}$, yielding an exactly zero change point error. This demonstrates the ``oracle'' property of our algorithm, which works as if the oracle change points and clusters were known to us. Notice that such a property is generally impossible to attain by the Subject-wise algorithm.
    \item The Homogeneous algorithm generally fails to consistently identify change points, 
    due to its overlooking of the heterogeneity across subjects. 
\end{itemize}

Second, 
the clustering error (e.g., the proportion of incorrectly clustered subjects) equals \textit{exactly} zero with probability at least $1-O(N^{-1}T^{-1})$, demonstrating the advantage of borrowing information over time. 
In contrast, neither the \textbf{Stationary} algorithm nor the \textbf{Time-wise} algorithm are able to achieve an \textit{asymptotically} zero clustering error as the total number of observations $NT\to \infty$. For the Stationary algorithm, this is because it ignores the nonstationarity over time. 
As for the Time-wise algorithm, its inconsistency is due to that it only uses observations at the endpoint, and it is generally impossible to obtain a diminishing clustering error without repeated observations \citep[i.e., when $T=1$,][]{Marriott1975,ECTA11319}. 

Next, we formally summarize our results in the following theorems. 
To save space, we relegate some additional regularity conditions 
to Section \ref{sec:addassump} of the  Supplementary Materials. Specifically, Assumption \ref{as:tau} is concerned with the change point location and cluster sizes. Assumption \ref{as:compact} requires both the state space and the parameter space to be compact. 
Assumption \ref{as:f'' contineous} assumes the boundedness and differentiability of the transition function $p$. Assumption \ref{as:signal} imposes certain signal strengths conditions to guarantee the consistency of the proposed algorithm. 
Assumption \ref{as:ergodic} requires the underlying MDP to satisfy certain mixing conditions. Assumption \ref{as:samplesize} requires $T$ that should not be too small relative to $N$. 
For any $z\in \mathbb{R}$, we use $z_+$ to denote $\max(z,0)$.

\begin{thm}\label{thmrate}
Suppose MA, Assumptions \ref{ass:LHE}, \ref{ass:LSE}, and \ref{as:tau}--\ref{as:samplesize} in the Supplementary Materials hold and that as $T\to \infty$, the initial estimators satisfy $\max_i (\tau_i^0-\tau_i^*)_+/\tau_i^*\ll T^{-1/2}\sqrt{\log(NT)}$ and $\min_i \tau_i^0\ge \kappa T$ for some constant $\kappa>0$; 
Then at each iteration, the following holds: 
\begin{itemize}
    \item The proposed IC is consistent in identifying the true number of clusters as $T\to \infty$.
    \item For sufficiently large $T$, the clustering error equals exactly zero, with probability at least $1-O(N^{-1}T^{-1})$.
    \item For sufficiently large $T$, the estimated change points satisfy \eqref{eqn:cperr}, with probability at least $1-O(N^{-1}T^{-1})$.
\end{itemize}
\end{thm}

\begin{coro}\label{thmrate2}
Suppose the conditions in Theorem \ref{thmrate} hold. Suppose that $T$ is sufficiently large and $N$ satisfies $N\gg s_{cp}^{-2} \log^3 (NT)$. Then at each iteration, with  probability at least $1-O(N^{-1}T^{-1})$, both the change point detection error and the clustering error equal zero. 
\end{coro}

Finally, we discuss the assumptions on the initial change point estimators in 
Theorem \ref{thmrate}. Notice that we only require the \textit{overestimation} error of each initial estimator (e.g., $(\tau_i^0-\tau_i^*)_+$) to satisfy a certain rate. As long as $\tau_i^0$ is proportional to $T$, the \textit{underestimation} error (e.g., $(\tau_i^*-\tau_i^0)_+$) will not affect the final estimators' rates of convergence. 

\subsection{Regret analysis}\label{sec:regret}
As commented in Section \ref{sec:prelim:environment}, in offline settings, we aim to identify an optimal warm-up policy for the $N$ subjects to maximize their cumulative rewards after time $T$. Specifically, for a given warm-up policy $\pi$, we measure its quality by the expected cumulative reward under $\pi$, defined as
\begin{eqnarray*}
    J(\pi)=\frac{1}{N}\sum_{i=1}^N \sum_{t=T+1}^{\infty} \gamma^{t-T-1} \Mean^{\pi} (R_{i,t}),
\end{eqnarray*}
where $\Mean^{\pi} (R_{i,t})$ denotes the $i$th subject's expected reward at time $t$ assuming they follow policy $\pi$ from time $T+1$ onwards, and $\gamma$ denotes the discount factor. 

In this section, we demonstrate how the clustering and change point detection errors translate into sub-optimalities of the estimated optimal policy. In particular, we apply FQI --- a classical doubly homogeneous Q-learning algorithm --- to the estimated data rectangles for policy learning. Meanwhile, other RL algorithms are equally applicable. Our theoretical analysis is concerned with the regret of warm-up policies $\pi$ computed via FQI, which quantifies the gap between the best possible expected cumulative rewards $\sup_{\pi^*} J(\pi^*)$ and $J(\pi)$ where the supremum is taken over the set of all possible subject-specific history-dependent policies $\pi^*$. By definition, a smaller regret indicates a better policy. 

\begin{table}[t]
\small
	\caption{Regrets of estimated optimal polices computed via various algorithms when $N$ and $T$ have different properties. The errors shown are under high probabilities.}\label{tab:regret}
	\begin{tabular}{ccccc}
	\toprule
	& Oracle & Proposed & Subject-wise & Other algorithms\\ \midrule
	$T\to \infty$ \& $N$ fixed & $\displaystyle O\Big(\frac{R_{\max}\log(NT)}{\sqrt{NT}(1-\gamma)^3}\Big)$& $\displaystyle O\Big(\frac{R_{\max}\log^{3/2}(NT)}{\sqrt{NT}(1-\gamma)^3s_{cp}}\Big)$ & $\displaystyle O\Big(\frac{R_{\max}\log^{3/2}(T)}{\sqrt{T}(1-\gamma)^3s_{cp}}\Big)$ & non-negligible \\
 $T\to \infty$ \& $N\to \infty$ &$\displaystyle O\Big(\frac{R_{\max}\log(NT)}{\sqrt{NT}(1-\gamma)^3}\Big)$ & $\displaystyle O\Big(\frac{R_{\max}\log(NT)}{\sqrt{NT}(1-\gamma)^3}\Big)$ &  $\displaystyle O\Big(\frac{R_{\max}\log^{3/2}(T)}{\sqrt{T}(1-\gamma)^3s_{cp}}\Big)$ & non-negligible \\
 $T$ fixed $\&$ $N$ fixed & $\displaystyle O\Big(\frac{R_{\max}\log(NT)}{\sqrt{NT}(1-\gamma)^3}\Big)$ & non-negligible & non-negligible & non-negligible\\ \bottomrule
	\end{tabular}
\end{table}

Regrets of various estimated policies are summarised in Table \ref{tab:regret}. In particular, Theorem \ref{thmregret} below shows that for sufficiently large $T$, the regret of the warm-up policy computed by the proposed algorithm is upper bounded by
\begin{eqnarray}\label{eqn:regret}
    O\Big[\frac{R_{\max}\log^{3/2}(NT)}{(1-\gamma)^3\sqrt{NT}s_{cp}}\Big],
\end{eqnarray}
with probability at least $1-O(N^{-1}T^{-1})$, where $R_{\max}$ denotes the upper bound on the immediate reward, specified in Assumption \ref{as:reward} in the Supplementary Materials. Notice that the regret bound depends additionally on $s_{cp}$, which measures the difference in the transition function before and after the change point. 

Additionally, when $N$ is also sufficiently large and satisfies $N\gg s_{cp}^{-2} \log^3 (NT)$, the upper bound can be further sharpened to 
\begin{eqnarray}\label{eqn:regret1}
    O\Big[\frac{R_{\max}\log(NT)}{(1-\gamma)^3\sqrt{NT}}\Big].
\end{eqnarray}
However, the estimated policy is not guaranteed to be consistent with a finite $T$ as $N\to \infty$, since both the change point detection and clustering errors can be non-negligible; see Table \ref{tab:rate_convergence}. 

We next analytically compare our algorithm against the rest of the algorithms one by one:
\begin{itemize}
    \item \textbf{Oracle}: Using similar arguments to the proof of Theorem \ref{thmregret}, one can show that the Oracle algorithm achieves of a regret of \eqref{eqn:regret1}. 
    This regret is the same to ours in settings where both $N$ and $T$ are sufficiently large, which demonstrates the oracle property of our algorithm. This is expected as our algorithm achieves zero clustering and change point detection errors in these settings. In other settings, our algorithm's regret is larger than that of the Oracle algorithm by a factor of $s_{cp}^{-1}\log^{1/2}(NT)$, 
    demonstrating the price we pay to accommodate double inhomogeneity.
    \item \textbf{Subject-wise}: Similar to Theorem \ref{thmregret}, one can show that the Subject-wise algorithm achieves a regret of $O(T^{-1/2}R_{\max}(1-\gamma)^{-3}s_{cp}^{-1})$, up to logarithmic factors, with probability at least $1-O(T^{-1}N^{-1})$. In comparison, the rate at which the regret of our algorithm decays to zero is faster by a factor of $N^{-1/2}$, demonstrating the advantage of borrowing information over subjects.
    \item \textbf{Time-wise}: The regret of the Time-wise algorithm is not guaranteed to decay to zero, since its clustering error is non-negligible (i.e., it does not decay to zero as $N T\to \infty$), as discussed in Section \ref{sec:cpcluster}. Even in settings where the cluster memberships were known, similar to Theorem \ref{thmregret}, one could show that the resulting algorithm would achieve a regret of $O(N^{-1/2}R_{\max}(1-\gamma)^{-3}s_{cp}^{-1})$, up to logarithmic factors, with probability at least $1-O(N^{-1})$. This rate is still slower than \eqref{eqn:regret} when $T\gg s_{cp}^{-2}$, since the algorithm only uses the observations at the last time point to learn the optimal policy. 
    \item \textbf{Others}: The regrets of the remaining three algorithms -- Homogeneous, Stationary and Doubly Homogeneous -- are all non-negligible, as shown in Theorem \ref{thmregret}, since they all ignore at least one type of inhomogeneity. 
\end{itemize}
We next formally establish these results, which require some additional assumptions imposed in Section \ref{sec:addassump} of the Supplementary Materials. In particular, notice that FQI is an iterative algorithm. Assumption \ref{as:FQIiter} requires the number of iterations to be much larger than $\log(NT)/\log(\gamma^{-1})$. This assumption is mild, as the number of iterations is user-specified, and we can choose a sufficiently large value to satisfy this assumption. Assumption \ref{as:reward} requires all immediate rewards to be bounded, and Assumption \ref{as:complete} requires completeness of the Q-function class used in FQI.
Both are frequently imposed in the RL literature \citep[see e.g.,][]{chen2019information,fan2020theoretical,uehara2021finite}. Finally, Assumption \ref{as:VCclass} requires the estimated Q-function to belong to the Vapnik-Chervonenkis (VC) class \citep{van1996}, which relaxes some existing conditions that assume a finite hypothesis class \citep{chen2019information}.  

Recall from Section \ref{sec:prelim:environment} that in doubly inhomogeneous environments, the clusters can merge, split, evolve, etc, when a new change point occurs. We denote by $T^*$ the next change point after $T$. That is, for each subject, its transition function remains constant from time $T$ to $T^*-1$.
\begin{thm}\label{thmregret}
    Suppose the conditions in Theorem \ref{thmrate} hold. Suppose Assumptions \ref{as:FQIiter} -- \ref{as:VCclass} hold. Then for sufficiently large $T$, the regret of the estimated policy computed via the proposed doubly inhomogeneous algorithm is upper bounded by 
    \begin{eqnarray}\label{eqn:thmregret}
        O\Big[\frac{R_{\max}\log(NT)}{(1-\gamma)^3\sqrt{NT}s_{cp}}\Big]+\frac{2R_{\max}\Mean (\gamma^{T^*-T})}{\gamma(1-\gamma)},
    \end{eqnarray}
    with probability at least $1-O(N^{-1}T^{-1})$. In particular:\vspace{-1em}
    \begin{itemize}
        \item When $T^*-T\gg \log T/\log(\gamma^{-1})$, the proposed algorithms achieves a regret of \eqref{eqn:regret1};
        \item When $N\gg s_{cp}^2\log^3(NT)$, the first term in \eqref{eqn:thmregret} can be further sharpened to \eqref{eqn:regret}. 
    \end{itemize}
    Meanwhile, for the Homogeneous, Stationary and Doubly Homogeneous algorithms, there exists at least one MDP such that their estimated optimal policy's regret does not decay to zero as $NT\to \infty$. 
\end{thm}
We remark that the condition $T^*-T\gg \log T/\log(\gamma^{-1})$ in Theorem \eqref{thmregret} is likely to hold, as it only requires the new change point to occur after $\log(T)$ many observations, a number much smaller than $T$. Assuming this condition holds leads to the regret bounds in \eqref{eqn:regret} and \eqref{eqn:regret1}. When this condition is violated, we can sequentially apply our algorithm to update the warm-up policy, to further reduce the regret.

\section{Simulation experiments}\label{sec:simu}
We conduct simulation experiments to evaluate the finite sample performance of our algorithm and compare it against existing doubly homogeneous or singly inhomogeneous algorithms in terms of offline change point detection and clustering errors (Section \ref{sec:sim:real:offline}) as well as online policy regrets (Section \ref{sec:sim:real:online}). Thoughout this section, we simulate a semi-synthetic data environment mimicking the motivating IHS dataset to illustrate the utilities of the proposed algorithm. To save space, 
some implementation details are relegated to  Section \ref{sec:sim:real:additional} of the Supplementary Materials 
 \footnote{See 
\url{https://github.com/zaza0209/DIRL}
for a Python implementation of our algorithm. }. 

\subsection{Offline change point detection and clustering} \label{sec:sim:real:offline}
\noindent \textit{\textbf{Data generating process}}. 
The offline data is generated based on our IHS data analysis in Section \ref{sec:motivation}, which identified two distinct clusters of interns and associated change points.
The state vector comprises three variables from IHS: the cubic root of daily step count, the square root of daily sleep minutes, and daily mood score. The transformations were conducted to make the distributions less skewed and help with estimations. The actions are binary: $A_{it} = 1$ (with probability 0.5) indicates the subject is randomized to receive activity messages at time $t$, whereas $A_{it} = -1$ (with probability 0.5) indicates other types of messages (mood or sleep) or no messages at all. The step count in the following day is used as the reward.

We estimate the transition 
functions for each of the four data rectangles identified from the IHS dataset (see e.g., Figure \ref{fig:realdata:proposed_sleep_cpcluster}) separately and use them for generating the offline data. Specifically, denote the transition functions for the first cluster before and after the change point as $\mathcal{P}_1$ and $\mathcal{P}_2$, respectively. Similarly, denote the transition functions for the second cluster before and after the change point as $\mathcal{P}_3$ and $\mathcal{P}_4$.
We simulate an offline dataset over the time interval $[0, 99]$ for 60 subjects. For the first 20 subjects, they follow $\mathcal{P}_1$ from $t=0$ to $t=49$ and switch to $\mathcal{P}_2$ at $t=50$. Similarly, the next 20 subjects follow $\mathcal{P}_3$ from $t=0$ to $t=49$ and then switch to $\mathcal{P}_4$ at $t=50$. To introduce further complexity in the environment, we add a third cluster with the last 20 subjects, who follow $\mathcal{P}_4$ throughout the entire period. Additionally, we vary the signal strengths ($s_{cp}$ \& $s_{cl}$) to be strong, moderate, and weak to create three simulation scenarios.

Overall, the simulation environment can be viewed as a complicated ``Evolution \& Constancy'' example, with two best data rectangles in the offline dataset. Our goal in this section is to identify these data rectangles.

\noindent \textit{\textbf{Competing methods and evaluation metrics}}. We compare the proposed algorithms against four singly inhomogeneous algorithms: (i) Time-wise; (ii) Subject-wise; (iii) Homogeneous; (iv) Stationary. We consider two evaluation criteria for comparison. To investigate the empirical performance of the estimated change points $\{\widehat{\tau}_i^*\}_i$, we define the CPD error as $\sum_{i=1}^N |\widehat{\tau}_i^* -\tau_i^*|(N\tau_i^*)^{-1}$, such that a larger error implies worse change point estimation. We also use the Adjusted Rand Index \citep[ARI,][]{hubert1985comparing} to measure the quality of estimated cluster memberships $\{\widehat{C}_k\}_k$. ARI is a chance-corrected measure between $0$ and $1$ with values near $1$ indicating higher degrees of concordance.

\noindent \textit{\textbf{Results}}.
Table \ref{tab:semisyn offline all methods} summarises the results, aggregated over 40 simulations. Under strong signals, the proposed algorithm's CPD errors are close to zero and their ARIs reach 1. As the signal strength decays, our algorithm produces larger CPD errors and smaller ARIs. This is consistent with our theoretical findings in Theorem \ref{thmrate}. In addition, we find that the proposed IC correctly identifies the number of clusters in most simulations with strong and moderate signals. This empirically verifies the consistency of our IC in settings where the signal is 
moderately strong. 

In contrast, the Time-wise algorithm struggles to consistently identify clusters because it relies on observations at the endpoint only for clustering. The Subject-wise algorithm also fails to consistently detect change points, even with strong signals, since it conducts change point detection separately for each individual trajectory. The Homogeneous algorithm does not identify any change points at all, as it ignores the heterogeneity across subjects, resulting in CPD errors 1.0. Finally, the Stationary algorithm performs well in clustering as reflected by high ARIs, despite the fact that it ignores non-stationarity. This is because the transition function of the last cluster $\mathcal{P}_4$ is stationary and aligns with that of the second cluster after their change point at $t=50$, which improves clustering performance even when non-stationarity is ignored.

\begin{table}[t]
\centering
\small
\begin{tabular}{lcccccc}
\hline
\multirow{2}{*}{Method} & \multicolumn{2}{c}{Strong} & \multicolumn{2}{c}{Moderate} & \multicolumn{2}{c}{Weak}\\ \cline{2-7}  
& CPD error & ARI & CPD error & ARI & CPD error & ARI \\\hline
 Proposed     & 0.318 (0.019) 
  & 1.000 (0.000) & 0.823 (0.036) 
 & 1.000 (0.000) & 1.000 (0.000) 
& 0.691 (0.084) \\
   Time-wise &  - & 0.014 (0.006) & - & 0.014 (0.006) & - & 0.000 (0.000) \\
   Subject-wise        & 0.804 (0.001) & -   & 0.805 (0.001) &- & 0.804 (0.001) & - \\
  Homogeneous       & 1.000 (0.000) & - & 1.000 (0.000) & - & 1.000 (0.000) & - \\
  Stationary  & - & 1.000 (0.000) & - &  1.000 (0.000)  & - & 0.796 (0.074) \\ \hline
\end{tabular}
\caption{Offline semi-synthetic simulation results: change point detection errors and ARIs of various algorithms, with standard errors in parentheses. ``-'' means not applicable.}
\label{tab:semisyn offline all methods}
\end{table}


\subsection{Online policy learning} \label{sec:sim:real:online}
\noindent \textit{\textbf{Data generating process and policy learning}}. We demonstrate the utility of the proposed policy learning algorithm in online settings. In each simulation, we first generate an offline dataset whose data distribution process is described in Section \ref{sec:sim:real:offline}.  
Next, we apply our algorithm 
to identify the most recent change points and cluster memberships, and then estimate the optimal warm-up policy based on FQI using the estimated data rectangles. We then employ the estimated optimal policy for action selection and generate a new data batch for the next 25 time points, and apply our algorithm again to update the estimated warm-up policy. This procedure is repeated to regularly update the estimated optimal policy assuming the online data arrive in batch every 25 time points, until the terminal time $400$ is reached. When generating the new data batches, the first two clusters are allowed to merge, split or switch, and the change points follow a Poisson process with a rate of $1/60$ so that we expect one change point every 60 time points. The third cluster stays constant throughout. See Section \ref{sec:sim:real:evaluation:additional} of the Supplementary Materials for more details about the setting and the policy learning algorithm. 

\noindent \textbf{\textit{Competing methods and policy value}}. We compare the proposed policy learning algorithm against (i) Oracle; (ii) Doubly Homogeneous; (iii) Homogeneous and (iv) Stationary. We did not compare against Time-wise and Subject-wise since they produced the optimal policy only based on 10 -- 20 observations, yielding highly unreliable policies. Nor did we compare against transfer or federated RL algorithms, since they require us to know the cluster memberships \textit{a priori}. For a fair comparison, all the aforementioned algorithms use FQI as the base policy learning algorithm, which is why we did not compare against the specialized non-stationary or heterogeneous RL algorithms  \citep{alegre2021minimum,chen2024reinforcement}. For each policy computed by either the proposed algorithm or the baseline algorithm, we averaged their cumulative reward over time and population to evaluate their policy value.  
\begin{table}[t]
\centering
\small
\caption{Online data analysis results: value differences between the proposed policy and the baseline policies, with standard errors in parentheses.}\label{tab:semisyn online average}
\begin{tabular}{lccc}
 \toprule
Policy       & Strong        & Moderate      & Weak          \\
\midrule
Proposed - Oracle       & -6.89 (3.59)  & -13.97 (2.65)  & -14.38 (2.57)  \\
Proposed - Doubly Homogeneous         & 52.05 (3.57)   & 30.12 (3.21)   & 10.02 (2.71)   \\
Proposed - Homogeneous  & 56.09 (4.69)   & 27.82 (3.23)   & 6.14 (2.77)   \\
Proposed - Stationary   & 25.12 (2.91)   & 6.54 (3.20)   & -3.59 (2.17)   \\
 \bottomrule
\end{tabular}
\end{table}

\noindent \textbf{\textit{Results}}. Table \ref{tab:semisyn online average} displays the value differences between the proposed policy and the baseline policies, aggregated over 40 replications. It can be seen that our policy achieves lower average rewards than the policy computed by the Oracle algorithm. This is expected since Oracle knows the true change points and cluster memberships. On the other hand, the average rewards of other baseline policies are significantly smaller than ours in most cases, demonstrating the danger of ignoring either non-stationarity or heterogeneity.

\section{Application to the Intern Health Study }\label{sec:realdata}
In this section, we apply the proposed policy learning algorithm to the 2020 IHS dataset. The state, action, and reward variables are the same as those described in Section \ref{sec:simu}. Without the knowledge of the oracle data generating process, we implement a cross-validation procedure to evaluate various warm-up policies using real data. Specifically, 
we perform change point detection, clustering and policy learning using the data from the first $T_{train}$ days, and  
evaluate the estimated policy using the data from the following $T_{train}+1$-th day. We assume the cluster structure on day $T_{train}+1$ remains consistent with that detected in the training set, implying no change point occurs between $T_{train}$ and $T_{train}+1$. We vary $T_{train}$ from 62 to 82 days and report the average policy value for policies computed by (i) the Proposed algorithm; (ii) Doubly Homogeneous; (iii) Homogeneous; (iv) Stationary. Similar to the analysis in Section \ref{sec:sim:real:online}, we did not compare against Time-wise and Subject-wise since their estimated policies are learned from a limited amount of data.  
Additional implementation details are provided in Section \ref{sec:sim:realIHS2020} of the Supplementary Materials.

It can be seen from Table \ref{tab:realdata_opevalues} 
that the proposed algorithm achieves the highest estimated policy value.  
In particular, our estimated optimal policy increases the daily average step count by approximately 45 steps compared to the DH algorithm, which produces a learned policy that does not change with time or subject. 
The stationary method, which fails to identify any cluster structures, performs similarly to DH.
Although the Homogeneous method detected similar change points to our proposal, the policy learned after these change points results in the lowest estimated step counts. This is due to the presence of heterogeneous clusters. As mentioned in Section \ref{sec:motivation}, the two clusters exhibit different treatment effects. Thus, their optimal policies are likely different as well. Finally, when compared to the behavior policy, demonstrates a substantial improvement, leading to a 10\% increase in average daily step counts. 
These findings further confirm the existence of double inhomogeneity, and highlight the necessity of identifying both types of inhomogeneity for more effective policy learning.

\bibliographystyle{rss}
\bibliography{mycite}

\begin{thebibliography}{87}
\expandafter\ifx\csname natexlab\endcsname\relax\def\natexlab#1{#1}\fi
\expandafter\ifx\csname url\endcsname\relax
  \def\url#1{\texttt{#1}}\fi
\expandafter\ifx\csname urlprefix\endcsname\relax\def\urlprefix{URL: }\fi

\bibitem[{Alegre et~al.(2021)Alegre, Bazzan and da~Silva}]{alegre2021minimum}
Alegre, L.~N., Bazzan, A.~L. and da~Silva, B.~C. (2021) Minimum-delay
  adaptation in non-stationary reinforcement learning via online
  high-confidence change-point detection.
\newblock In \textit{Proceedings of the 20th International Conference on
  Autonomous Agents and MultiAgent Systems}, 97--105.

\bibitem[{Bian et~al.(2025)Bian, Shi, Qi and Wang}]{bian2023off}
Bian, Z., Shi, C., Qi, Z. and Wang, L. (2025) Off-policy evaluation in doubly
  inhomogeneous environments.
\newblock \textit{Journal of the American Statistical Association}, To appear.

\bibitem[{Bonhomme and Manresa(2015)}]{ECTA11319}
Bonhomme, S. and Manresa, E. (2015) Grouped patterns of heterogeneity in panel
  data.
\newblock \textit{Econometrica}, \textbf{83}, 1147--1184.

\bibitem[{Bradley(2005)}]{Bradley2005}
Bradley, R.~C. (2005) Basic properties of strong mixing conditions. {A} survey
  and some open questions.
\newblock \textit{Probability Surveys}, \textbf{2}, 107--144.

\bibitem[{Casella and Berger(2024)}]{casella2024statistical}
Casella, G. and Berger, R. (2024) \textit{Statistical inference}.
\newblock CRC Press.

\bibitem[{Chakraborty and Moodie(2013)}]{chakraborty2013statistical}
Chakraborty, B. and Moodie, E.~E. (2013) \textit{Statistical Methods for
  Dynamic Treatment Regimes: Reinforcement Learning, Causal Inference, and
  Personalized Medicine}, vol.~76.
\newblock Springer Science \& Business Media.

\bibitem[{Chen and Hong(2012)}]{chen2012testing}
Chen, B. and Hong, Y. (2012) Testing for the {Markov} property in time series.
\newblock \textit{Econometric Theory}, \textbf{28}, 130--178.

\bibitem[{Chen et~al.(2022)Chen, Jordan and Li}]{chen2022transferred}
Chen, E.~Y., Jordan, M.~I. and Li, S. (2022) Transferred {Q}-learning.
\newblock \textit{arXiv preprint arXiv:2202.04709}.

\bibitem[{Chen et~al.(2024)Chen, Song and Jordan}]{chen2024reinforcement}
Chen, E.~Y., Song, R. and Jordan, M.~I. (2024) Reinforcement learning in latent
  heterogeneous environments.
\newblock \textit{Journal of the American Statistical Association}, 1--32.

\bibitem[{Chen and Jiang(2019)}]{chen2019information}
Chen, J. and Jiang, N. (2019) Information-theoretic considerations in batch
  reinforcement learning.
\newblock In \textit{International Conference on Machine Learning}, 1042--1051.
  PMLR.

\bibitem[{Chen and Christensen(2015)}]{ChenChristensen2015}
Chen, X. and Christensen, T.~M. (2015) {Optimal uniform convergence rates and
  asymptotic normality for series estimators under weak dependence and weak
  conditions}.
\newblock \textit{Journal of Econometrics}, \textbf{188}, 447--465.
\newblock
  \urlprefix\url{https://ideas.repec.org/a/eee/econom/v188y2015i2p447-465.html}.

\bibitem[{Chen et~al.(2023)Chen, Qi and Wan}]{chen2023steel}
Chen, X., Qi, Z. and Wan, R. (2023) Steel: Singularity-aware reinforcement
  learning.
\newblock \textit{arXiv preprint arXiv:2301.13152}.

\bibitem[{Chernozhukov et~al.(2014)Chernozhukov, Chetverikov and
  Kato}]{cherno2014}
Chernozhukov, V., Chetverikov, D. and Kato, K. (2014) Gaussian approximation of
  suprema of empirical processes.
\newblock \textit{Ann. Statist.}, \textbf{42}, 1564--1597.

\bibitem[{Cheung et~al.(2020)Cheung, Simchi-Levi and
  Zhu}]{cheung2020reinforcement}
Cheung, W.~C., Simchi-Levi, D. and Zhu, R. (2020) Reinforcement learning for
  non-stationary {M}arkov decision processes: The blessing of (more) optimism.
\newblock In \textit{International Conference on Machine Learning}, 1843--1854.
  PMLR.

\bibitem[{Cho and Fryzlewicz(2012)}]{cho2012multiscale}
Cho, H. and Fryzlewicz, P. (2012) Multiscale and multilevel technique for
  consistent segmentation of nonstationary time series.
\newblock \textit{Statistica Sinica}, 207--229.

\bibitem[{Dwivedi et~al.(2022)Dwivedi, Tian, Tomkins, Klasnja, Murphy and
  Shah}]{dwivedi2022counterfactual}
Dwivedi, R., Tian, K., Tomkins, S., Klasnja, P., Murphy, S. and Shah, D. (2022)
  Counterfactual inference for sequential experiments.
\newblock \textit{arXiv preprint arXiv:2202.06891}.

\bibitem[{Dzhaparidze and Van~Zanten(2001)}]{dzhaparidze2001bernstein}
Dzhaparidze, K. and Van~Zanten, J. (2001) On bernstein-type inequalities for
  martingales.
\newblock \textit{Stochastic processes and their applications}, \textbf{93},
  109--117.

\bibitem[{Ernst et~al.(2005)Ernst, Geurts and Wehenkel}]{Ernst2005}
Ernst, D., Geurts, P. and Wehenkel, L. (2005) Tree-based batch mode
  reinforcement learning.
\newblock \textit{Journal of Machine Learning Research}, \textbf{6}, 503--556.

\bibitem[{Ertefaie and Strawderman(2018)}]{ertefaie2018constructing}
Ertefaie, A. and Strawderman, R.~L. (2018) Constructing dynamic treatment
  regimes over indefinite time horizons.
\newblock \textit{Biometrika}, \textbf{105}, 963--977.

\bibitem[{Fan et~al.(2020)Fan, Wang, Xie and Yang}]{fan2020theoretical}
Fan, J., Wang, Z., Xie, Y. and Yang, Z. (2020) A theoretical analysis of deep
  {Q}-learning.
\newblock In \textit{Learning for Dynamics and Control}, 486--489. PMLR.

\bibitem[{Fei et~al.(2020)Fei, Yang, Wang and Xie}]{fei2020dynamic}
Fei, Y., Yang, Z., Wang, Z. and Xie, Q. (2020) Dynamic regret of policy
  optimization in non-stationary environments.
\newblock In \textit{Advances in Neural Information Processing Systems},
  6743--6754.

\bibitem[{Frick et~al.(2014)Frick, Munk and Sieling}]{frick2014multiscale}
Frick, K., Munk, A. and Sieling, H. (2014) Multiscale change point inference.
\newblock \textit{Journal of the Royal Statistical Society Series B},
  \textbf{76}, 495--580.

\bibitem[{Friedrich et~al.(2008)Friedrich, Kempe, Liebscher and
  Winkler}]{friedrich2008complexity}
Friedrich, F., Kempe, A., Liebscher, V. and Winkler, G. (2008) Complexity
  penalized {M}-estimation: fast computation.
\newblock \textit{Journal of Computational and Graphical Statistics},
  \textbf{17}, 201--224.

\bibitem[{Fryzlewicz(2014)}]{fryzlewicz2014wild}
Fryzlewicz, P. (2014) Wild binary segmentation for multiple change-point
  detection.
\newblock \textit{The Annals of Statistics}, \textbf{42}, 2243--2281.

\bibitem[{Han et~al.(2021)Han, Hou, Cho, Duan and Cai}]{han2021federated}
Han, L., Hou, J., Cho, K., Duan, R. and Cai, T. (2021) Federated adaptive
  causal estimation ($face$) of target treatment effects.
\newblock \textit{arXiv preprint arXiv:2112.09313}.

\bibitem[{Hogg and Craig(1995)}]{hogg1995introduction}
Hogg, R.~V. and Craig, A.~T. (1995) Introduction to mathematical statistics.
\newblock \textit{Englewood Hills, New Jersey}.

\bibitem[{Hu et~al.(2021)Hu, Qian, Cheng and Cheung}]{hu2020personalized}
Hu, X., Qian, M., Cheng, B. and Cheung, Y.~K. (2021) Personalized policy
  learning using longitudinal mobile health data.
\newblock \textit{Journal of the American Statistical Association},
  \textbf{116}, 410--420.

\bibitem[{Hu and Wager(2023)}]{hu2023off}
Hu, Y. and Wager, S. (2023) Off-policy evaluation in partially observed
  {Markov} decision processes under sequential ignorability.
\newblock \textit{The Annals of Statistics}, \textbf{51}, 1561--1585.

\bibitem[{Hubert and Arabie(1985)}]{hubert1985comparing}
Hubert, L. and Arabie, P. (1985) Comparing partitions.
\newblock \textit{Journal of Classification}, \textbf{2}, 193--218.

\bibitem[{Huling et~al.(2018)Huling, Yu, Liang and Smith}]{huling2018risk}
Huling, J.~D., Yu, M., Liang, M. and Smith, M. (2018) Risk prediction for
  heterogeneous populations with application to hospital admission prediction.
\newblock \textit{Biometrics}, \textbf{74}, 557--565.

\bibitem[{Jiang et~al.(2022)Jiang, Zhao and Shao}]{jiang2022modelling}
Jiang, F., Zhao, Z. and Shao, X. (2022) Modelling the covid-19 infection
  trajectory: A piecewise linear quantile trend model.
\newblock \textit{Journal of the Royal Statistical Society Series B},
  \textbf{84}, 1589--1607.

\bibitem[{Jin et~al.(2022)Jin, Peng, Yang, Wang and Zhang}]{jin2022federated}
Jin, H., Peng, Y., Yang, W., Wang, S. and Zhang, Z. (2022) Federated
  reinforcement learning with environment heterogeneity.
\newblock In \textit{International Conference on Artificial Intelligence and
  Statistics}, 18--37. PMLR.

\bibitem[{Jin et~al.(2021)Jin, Yang and Wang}]{jin2021pessimism}
Jin, Y., Yang, Z. and Wang, Z. (2021) Is pessimism provably efficient for
  offline {RL}?
\newblock In \textit{International Conference on Machine Learning}, 5084--5096.
  PMLR.

\bibitem[{Killick et~al.(2012)Killick, Fearnhead and
  Eckley}]{killick2012optimal}
Killick, R., Fearnhead, P. and Eckley, I.~A. (2012) Optimal detection of
  changepoints with a linear computational cost.
\newblock \textit{Journal of the American Statistical Association},
  \textbf{107}, 1590--1598.

\bibitem[{Klasnja et~al.(2019)Klasnja, Smith, Seewald, Lee, Hall, Luers, Hekler
  and Murphy}]{klasnja2019efficacy}
Klasnja, P., Smith, S., Seewald, N.~J., Lee, A., Hall, K., Luers, B., Hekler,
  E.~B. and Murphy, S.~A. (2019) Efficacy of contextually tailored suggestions
  for physical activity: A micro-randomized optimization trial of {H}eartsteps.
\newblock \textit{Annals of Behavioral Medicine}, \textbf{53}, 573--582.

\bibitem[{Kosorok and Laber(2019)}]{kosorok2019precision}
Kosorok, M.~R. and Laber, E.~B. (2019) Precision medicine.
\newblock \textit{Annual review of statistics and its application}, \textbf{6},
  263--286.

\bibitem[{Levine et~al.(2020)Levine, Kumar, Tucker and Fu}]{levine2020offline}
Levine, S., Kumar, A., Tucker, G. and Fu, J. (2020) Offline reinforcement
  learning: Tutorial, review, and perspectives on open problems.
\newblock \textit{arXiv preprint arXiv:2005.01643}.

\bibitem[{Li et~al.(2024)Li, Shi, Chen, Chi and Wei}]{li2024settling}
Li, G., Shi, L., Chen, Y., Chi, Y. and Wei, Y. (2024) Settling the sample
  complexity of model-based offline reinforcement learning.
\newblock \textit{The Annals of Statistics}, \textbf{52}, 233--260.

\bibitem[{Li et~al.(2022{\natexlab{a}})Li, Berrett and Yu}]{li2022network}
Li, M., Berrett, T. and Yu, Y. (2022{\natexlab{a}}) Network change point
  localisation under local differential privacy.
\newblock \textit{Advances in Neural Information Processing Systems},
  \textbf{35}, 15013--15026.

\bibitem[{Li et~al.(2022{\natexlab{b}})Li, Shi, Wu and
  Fryzlewicz}]{li2022TestingStationarity}
Li, M., Shi, C., Wu, Z. and Fryzlewicz, P. (2022{\natexlab{b}}) Testing
  stationarity and change point detection in reinforcement learning.
\newblock \textit{arXiv preprint arXiv:2203.01707}.

\bibitem[{Li et~al.(2022{\natexlab{c}})Li, Cai and Li}]{li2022transfer}
Li, S., Cai, T.~T. and Li, H. (2022{\natexlab{c}}) Transfer learning for
  high-dimensional linear regression: Prediction, estimation and minimax
  optimality.
\newblock \textit{Journal of the Royal Statistical Society Series B},
  \textbf{84}, 149--173.

\bibitem[{Li(2019)}]{li2019reinforcement}
Li, Y. (2019) Reinforcement learning applications.
\newblock \textit{arXiv preprint arXiv:1908.06973}.

\bibitem[{Liao et~al.(2020)Liao, Greenewald, Klasnja and
  Murphy}]{liao2020personalized}
Liao, P., Greenewald, K., Klasnja, P. and Murphy, S. (2020) Personalized
  {H}eartsteps: A reinforcement learning algorithm for optimizing physical
  activity.
\newblock \textit{Proceedings of the ACM on Interactive, Mobile, Wearable and
  Ubiquitous Technologies}, \textbf{4}, 1--22.

\bibitem[{Liao et~al.(2022)Liao, Qi, Wan, Klasnja and Murphy}]{liao2022batch}
Liao, P., Qi, Z., Wan, R., Klasnja, P. and Murphy, S.~A. (2022) Batch policy
  learning in average reward {Markov} decision processes.
\newblock \textit{Annals of statistics}, \textbf{50}, 3364.

\bibitem[{Liu et~al.(2023)Liu, Tu, Zhang and Chen}]{liu2023online}
Liu, W., Tu, J., Zhang, Y. and Chen, X. (2023) Online estimation and inference
  for robust policy evaluation in reinforcement learning.
\newblock \textit{arXiv preprint arXiv:2310.02581}.

\bibitem[{Lorden(1971)}]{lorden1971procedures}
Lorden, G. (1971) Procedures for reacting to a change in distribution.
\newblock \textit{The annals of mathematical statistics}, 1897--1908.

\bibitem[{Luckett et~al.(2020)Luckett, Laber, Kahkoska, Maahs, Mayer-Davis and
  Kosorok}]{luckett2020estimating}
Luckett, D.~J., Laber, E.~B., Kahkoska, A.~R., Maahs, D.~M., Mayer-Davis, E.
  and Kosorok, M.~R. (2020) Estimating dynamic treatment regimes in mobile
  health using {V}-learning.
\newblock \textit{Journal of the American Statistical Association},
  \textbf{115}, 692--706.

\bibitem[{Mao et~al.(2021)Mao, Zhang, Zhu, Simchi-Levi and
  Basar}]{pmlr-v139-mao21b}
Mao, W., Zhang, K., Zhu, R., Simchi-Levi, D. and Basar, T. (2021) Near-optimal
  model-free reinforcement learning in non-stationary episodic {MDP}s.
\newblock In \textit{Proceedings of the 38th International Conference on
  Machine Learning}, 7447--7458. PMLR.

\bibitem[{Marriott(1975)}]{Marriott1975}
Marriott, F. H.~C. (1975) 389: Separating mixtures of normal distributions.
\newblock \textit{Biometrics}, \textbf{31}, 767--769.

\bibitem[{Meitz and Saikkonen(2019)}]{meitz2019subgeometric}
Meitz, M. and Saikkonen, P. (2019) Subgeometric ergodicity and beta-mixing.
\newblock \textit{arXiv preprint arXiv:1904.07103}.

\bibitem[{Mendelson et~al.(2008)Mendelson, Pajor and
  Tomczak-Jaegermann}]{mendelson2008uniform}
Mendelson, S., Pajor, A. and Tomczak-Jaegermann, N. (2008) Uniform uncertainty
  principle for bernoulli and subgaussian ensembles.
\newblock \textit{Constructive Approximation}, \textbf{28}, 277--289.

\bibitem[{Mnih et~al.(2015)Mnih, Kavukcuoglu, Silver, Rusu, Veness, Bellemare,
  Graves, Riedmiller, Fidjeland, Ostrovski et~al.}]{mnih2015human}
Mnih, V., Kavukcuoglu, K., Silver, D., Rusu, A.~A., Veness, J., Bellemare,
  M.~G., Graves, A., Riedmiller, M., Fidjeland, A.~K., Ostrovski, G. et~al.
  (2015) Human-level control through deep reinforcement learning.
\newblock \textit{Nature}, \textbf{518}, 529--533.

\bibitem[{Nahum-Shani et~al.(2022)Nahum-Shani, Shaw, Carpenter, Murphy and
  Yoon}]{nahum2022engagement}
Nahum-Shani, I., Shaw, S.~D., Carpenter, S.~M., Murphy, S.~A. and Yoon, C.
  (2022) Engagement in digital interventions.
\newblock \textit{American Psychologist}.

\bibitem[{NeCamp et~al.(2020)NeCamp, Sen, Frank, Walton, Ionides, Fang, Tewari
  and Wu}]{necamp2020assessing}
NeCamp, T., Sen, S., Frank, E., Walton, M.~A., Ionides, E.~L., Fang, Y.,
  Tewari, A. and Wu, Z. (2020) Assessing real-time moderation for developing
  adaptive mobile health interventions for medical interns: Micro-randomized
  trial.
\newblock \textit{Journal of Medical Internet Research}, \textbf{22}, e15033.

\bibitem[{Ouyang et~al.(2022)Ouyang, Wu, Jiang, Almeida, Wainwright, Mishkin,
  Zhang, Agarwal, Slama, Ray et~al.}]{ouyang2022training}
Ouyang, L., Wu, J., Jiang, X., Almeida, D., Wainwright, C., Mishkin, P., Zhang,
  C., Agarwal, S., Slama, K., Ray, A. et~al. (2022) Training language models to
  follow instructions with human feedback.
\newblock \textit{Advances in neural information processing systems},
  \textbf{35}, 27730--27744.

\bibitem[{Padakandla et~al.(2020)Padakandla, Prabuchandran and
  Bhatnagar}]{padakandla2020reinforcement}
Padakandla, S., Prabuchandran, K. and Bhatnagar, S. (2020) Reinforcement
  learning algorithm for non-stationary environments.
\newblock \textit{Applied Intelligence}, \textbf{50}, 3590--3606.

\bibitem[{Page(1954)}]{page1954continuous}
Page, E.~S. (1954) Continuous inspection schemes.
\newblock \textit{Biometrika}, \textbf{41}, 100--115.

\bibitem[{Puterman(1994)}]{Puterman1994}
Puterman, M.~L. (1994) \textit{Markov decision processes: discrete stochastic
  dynamic programming}.
\newblock Wiley Series in Probability and Mathematical Statistics: Applied
  Probability and Statistics. John Wiley \& Sons, Inc., New York.

\bibitem[{Qin et~al.(2025)Qin, Tang, Li, Zhu and Ye}]{qin2025reinforcement}
Qin, Z., Tang, X., Li, Q., Zhu, H. and Ye, J. (2025) \textit{Reinforcement
  Learning in the Ridesharing Marketplace}.
\newblock Springer.

\bibitem[{Ramprasad et~al.(2023)Ramprasad, Li, Yang, Wang, Sun and
  Cheng}]{ramprasad2023online}
Ramprasad, P., Li, Y., Yang, Z., Wang, Z., Sun, W.~W. and Cheng, G. (2023)
  Online bootstrap inference for policy evaluation in reinforcement learning.
\newblock \textit{Journal of the American Statistical Association},
  \textbf{118}, 2901--2914.

\bibitem[{Riedmiller(2005)}]{riedmiller2005}
Riedmiller, M. (2005) Neural fitted q iteration--first experiences with a data
  efficient neural reinforcement learning method.
\newblock In \textit{European Conference on Machine Learning}, 317--328.
  Springer.

\bibitem[{Saxena et~al.(2017)Saxena, Prasad, Gupta, Bharill, Patel, Tiwari, Er,
  Ding and Lin}]{saxena2017review}
Saxena, A., Prasad, M., Gupta, A., Bharill, N., Patel, O.~P., Tiwari, A., Er,
  M.~J., Ding, W. and Lin, C.-T. (2017) A review of clustering techniques and
  developments.
\newblock \textit{Neurocomputing}, \textbf{267}, 664--681.

\bibitem[{Schwarz(1978)}]{Schwarz1978}
Schwarz, G. (1978) Estimating the dimension of a model.
\newblock \textit{The Annals of Statistics}, \textbf{6}, 461--464.

\bibitem[{Shi et~al.(2024)Shi, Luo, Le, Zhu and Song}]{shi2024statistically}
Shi, C., Luo, S., Le, Y., Zhu, H. and Song, R. (2024) Statistically efficient
  advantage learning for offline reinforcement learning in infinite horizons.
\newblock \textit{Journal of the American Statistical Association},
  \textbf{119}, 232--245.

\bibitem[{Shi et~al.(2018)Shi, Song, Lu and Fu}]{shi2018maximin}
Shi, C., Song, R., Lu, W. and Fu, B. (2018) Maximin projection learning for
  optimal treatment decision with heterogeneous individualized treatment
  effects.
\newblock \textit{Journal of Royal Statistical Society: Series B}, \textbf{80},
  681--702.

\bibitem[{Shi et~al.(2020)Shi, Wan, Song, Lu and Leng}]{shi2020does}
Shi, C., Wan, R., Song, R., Lu, W. and Leng, L. (2020) Does the {M}arkov
  decision process fit the data: Testing for the {M}arkov property in
  sequential decision making.
\newblock In \textit{Proceedings of the 37th International Conference on
  Machine Learning}, 8807--8817. PMLR.

\bibitem[{Shi et~al.(2022)Shi, Zhang, Lu and Song}]{shi2020statistical}
Shi, C., Zhang, S., Lu, W. and Song, R. (2022) Statistical inference of the
  value function for reinforcement learning in infinite horizon settings.
\newblock \textit{Journal of Royal Statistical Society: Series B}, \textbf{84}.

\bibitem[{Sutton and Barto(2018)}]{sutton2018reinforcement}
Sutton, R.~S. and Barto, A.~G. (2018) \textit{Reinforcement learning: An
  introduction}.
\newblock MIT press.

\bibitem[{Tang and Song(2016)}]{tang2016fused}
Tang, L. and Song, P.~X. (2016) Fused lasso approach in regression coefficients
  clustering--learning parameter heterogeneity in data integration.
\newblock \textit{Journal of Machine Learning Research}, \textbf{17}, 1--23.

\bibitem[{Tomkins et~al.(2021)Tomkins, Liao, Klasnja and
  Murphy}]{tomkins2021intelligentpooling}
Tomkins, S., Liao, P., Klasnja, P. and Murphy, S. (2021) Intelligentpooling:
  Practical {T}hompson sampling for m{H}ealth.
\newblock \textit{Machine learning}, \textbf{110}, 2685--2727.

\bibitem[{Uehara et~al.(2021)Uehara, Imaizumi, Jiang, Kallus, Sun and
  Xie}]{uehara2021finite}
Uehara, M., Imaizumi, M., Jiang, N., Kallus, N., Sun, W. and Xie, T. (2021)
  Finite sample analysis of minimax offline reinforcement learning:
  Completeness, fast rates and first-order efficiency.
\newblock \textit{arXiv preprint arXiv:2102.02981}.

\bibitem[{van~der Vaart and Wellner(1996)}]{van1996}
van~der Vaart, A.~W. and Wellner, J.~A. (1996) \textit{Weak convergence and
  empirical processes}.
\newblock Springer Series in Statistics. Springer-Verlag, New York.

\bibitem[{Wan et~al.(2021)Wan, Zhang, Shi, Luo and Song}]{wan2021}
Wan, R., Zhang, S., Shi, C., Luo, S. and Song, R. (2021) Pattern transfer
  learning for reinforcement learning in order dispatching.
\newblock In \textit{Proceedings of IJCAI Reinforcement Learning for
  Intelligent Transportation Systems Workshop}.

\bibitem[{Wang et~al.(2023{\natexlab{a}})Wang, Fang, Frank, Walton, Burmeister,
  Tewari, Dempsey, NeCamp, Sen and Wu}]{wang2023effectiveness}
Wang, J., Fang, Y., Frank, E., Walton, M.~A., Burmeister, M., Tewari, A.,
  Dempsey, W., NeCamp, T., Sen, S. and Wu, Z. (2023{\natexlab{a}})
  Effectiveness of gamified team competition as mhealth intervention for
  medical interns: a cluster micro-randomized trial.
\newblock \textit{NPJ Digital Medicine}, \textbf{6}, 4.

\bibitem[{Wang et~al.(2023{\natexlab{b}})Wang, Qi and Wong}]{wang2023projected}
Wang, J., Qi, Z. and Wong, R.~K. (2023{\natexlab{b}}) Projected state-action
  balancing weights for offline reinforcement learning.
\newblock \textit{The Annals of Statistics}, \textbf{51}, 1639--1665.

\bibitem[{Wang et~al.(2023{\natexlab{c}})Wang, Shi and Wu}]{wang2023robust}
Wang, J., Shi, C. and Wu, Z. (2023{\natexlab{c}}) A robust test for the
  stationarity assumption in sequential decision making.
\newblock In \textit{International Conference on Machine Learning},
  36355--36379. PMLR.

\bibitem[{Wang and Samworth(2018)}]{wang2018high}
Wang, T. and Samworth, R.~J. (2018) High dimensional change point estimation
  via sparse projection.
\newblock \textit{Journal of the Royal Statistical Society Series B},
  \textbf{80}, 57--83.

\bibitem[{Wei and Luo(2021)}]{wei2021non}
Wei, C.-Y. and Luo, H. (2021) Non-stationary reinforcement learning without
  prior knowledge: An optimal black-box approach.
\newblock In \textit{Conference on Learning Theory}, 4300--4354. PMLR.

\bibitem[{Wei et~al.(2023)Wei, Ma and Wang}]{wei2023adaptive}
Wei, W., Ma, X. and Wang, J. (2023) Adaptive experiments toward learning
  treatment effect heterogeneity.
\newblock \textit{arXiv preprint arXiv:2312.06883}.

\bibitem[{Yang et~al.(2023)Yang, Cen, Wei, Chen and Chi}]{yang2023federated}
Yang, T., Cen, S., Wei, Y., Chen, Y. and Chi, Y. (2023) Federated natural
  policy gradient methods for multi-task reinforcement learning.
\newblock \textit{arXiv preprint arXiv:2311.00201}.

\bibitem[{Yang et~al.(2022)Yang, Zhang and Zhang}]{yang2022toward}
Yang, W., Zhang, L. and Zhang, Z. (2022) Toward theoretical understandings of
  robust {Markov} decision processes: Sample complexity and asymptotics.
\newblock \textit{The Annals of Statistics}, \textbf{50}, 3223--3248.

\bibitem[{Zhang et~al.(2020)Zhang, Sun and Li}]{zhang2020mixed}
Zhang, J., Sun, W.~W. and Li, L. (2020) Mixed-effect time-varying network model
  and application in brain connectivity analysis.
\newblock \textit{Journal of the American Statistical Association},
  \textbf{115}, 2022--2036.

\bibitem[{Zhao et~al.(2023)Zhao, Jiang, Yu and Chen}]{zhao2023high}
Zhao, Z., Jiang, F., Yu, Y. and Chen, X. (2023) High-dimensional dynamic
  pricing under non-stationarity: Learning and earning with change-point
  detection.
\newblock \textit{arXiv preprint arXiv:2303.07570}.

\bibitem[{Zhong et~al.(2021)Zhong, Yang and
  Szepesv{\'a}ri}]{zhong2021optimistic}
Zhong, H., Yang, Z. and Szepesv{\'a}ri, Z. W.~C. (2021) Optimistic policy
  optimization is provably efficient in non-stationary {MDP}s.
\newblock \textit{arXiv preprint arXiv:2110.08984}.

\bibitem[{Zhou et~al.(2024)Zhou, Zhu and Qu}]{zhou2022estimating}
Zhou, W., Zhu, R. and Qu, A. (2024) Estimating optimal infinite horizon dynamic
  treatment regimes via pt-learning.
\newblock \textit{Journal of the American Statistical Association},
  \textbf{119}, 625--638.

\bibitem[{Zhou et~al.(2023)Zhou, Shi, Li and Yao}]{zhou2023testing}
Zhou, Y., Shi, C., Li, L. and Yao, Q. (2023) Testing for the {Markov} property
  in time series via deep conditional generative learning.
\newblock \textit{Journal of the Royal Statistical Society Series B},
  \textbf{85}, 1204--1222.

\bibitem[{Zhu et~al.(2020)Zhu, Lin and Zhou}]{zhu2020transfer}
Zhu, Z., Lin, K. and Zhou, J. (2020) Transfer learning in deep reinforcement
  learning: A survey.
\newblock \textit{arXiv preprint arXiv:2009.07888}.

\end{thebibliography}

\appendix

This supplement is organised as follows. We first introduce  some notations and additional technical conditions, and provide auxiliary lemmas and proofs of our major theorems. We next detail our experimental settings and implementations.
\section{Notations, conditions and proofs}\label{sec:proof:auxiliary}
\subsection{Notations}\label{sec:notation}
For each cluster $k$, let $\theta_k^0$ denote the oracle parameter of the state transition model after the most recent change. Let $T-\tau^{(k)}$ denote the cluster-specific most recent change point of the $k$th cluster. We have $\tau^{(k)}=\tau_i^*$ for any $i\in \mathcal{C}_k$. 


We use $\theta_k^{1}$ to denote the limit of 
\begin{equation*}
\argmax_{\theta}\sum_{i\in \mathcal{C}_k}\sum_{t=T-\tau+1}^{T-\tau^{(k)}}\Mean \log p(\triple,\theta),
\end{equation*}
as $\tau-\tau^{(k)}$ diverges to infinity. Such a limit exists under the ergodicity assumption imposed in Section \ref{sec:addassump}. Define the signal strength of the temporal change as $s_{cp}=\min_k \|\tk^1-\tk^0\|$. Similarly, define the signal strength of the subject heterogeneity as $s_{cl}=\min_{k_1\neq k_2}\|\theta_{k_1}^0-\theta_{k_2}^0\|$. 

Let $\mathcal{B}_k^*$ denote the Bellman operator such that for any Q-function $Q$, $\mathcal{B}_k^* Q$ denotes another function given by 
\begin{eqnarray*}
    \mathcal{B}_k^* Q(s,a)=\Mean \Big[\max_{a^*} Q(S_{i,T+1},a^*)|S_{i,T}=s,A_{i,T}=a\Big],
\end{eqnarray*}
for any $i\in \mathcal{C}_k$. Additionally, let $\mathcal{Q}$ denote the Q-function class used to model the Q-function in FQI at each iteration. 

Throughout the proof, we use $c$ and $C$ to denote some generic constants whose values are allowed to vary from place to place. Finally, we define a non-negative number $\epsilon=\epsilon(N,T)$ dependent on $N$ and $T$ such that (i) $\epsilon=0$ when $N$ diverges to infinity with $T$; (ii) $\epsilon= T^{-1}\log(NT)\sqrt{\log(\log(NT))}$ when $N$ remains finite.

\subsection{Additional Assumptions}\label{sec:addassump}
\begin{asmp}[Change point locations and cluster sizes]\label{as:tau}
$\tau^{(1)}$, $\tau^{(2)}$, $\cdots$, $\tau^{(K)}$ are proportional to $T$, and $|\mathcal{C}_1|$, $|\mathcal{C}_2|$, $\cdots$, $|\mathcal{C}_K|$ are proportional to $N$. 
\end{asmp}
\begin{asmp}[Compact space and parameter spaces]\label{as:compact}
Both the state space and the parameter space $\Theta$ are  compact. 
\end{asmp}
\begin{asmp}[Differentiability and boundedness of the transition function]\label{as:f'' contineous}
The transition function $p$ is uniformly bounded away from zero, and is twice continuously differentiable with uniformly bounded second-order derivatives on $\Theta$. In addition, there exists some constant $\lambda>0$ such that
\begin{eqnarray*}
    \min_a \min_{\theta\in \Theta} \lambda_{\min}\left[-\int_{s,s'}\frac{\partial \log p(s'|s,a,\theta)}{\partial \theta\partial \theta^\top}dsds' \right]\ge \lambda,
\end{eqnarray*}
where $\lambda_{\min}[\bullet]$ denotes minimum eigenvalue of a given matrix.
\end{asmp}

\begin{asmp}[Signal strength]\label{as:signal}
     \sloppy $s_{cl}\gg \max(T^{-1/4}, \log^{3/2}(NT)/(\sqrt{NT} s_{cp}))$ and $s_{cp}\gg N^{-1/2} T^{-1/4} \log^{3/2} (NT)$.
\end{asmp}

\begin{asmp}[Geometric ergodicity]\label{as:ergodic}
For each $k$ and any $i\in \mathcal{C}_k$, the sequence $\{S_{i,t}\}_{t\ge T-\tau^{(k)}}$ is geometrically ergodic \citep[see e.g.,][for the detailed definition]{Bradley2005}. In addition, when $N$ is finite, the sequence $\{S_{i,t}\}_{(1-\bar{\epsilon})T-\tau_i^*\le t< T-\tau_i^{*}}$ is geometrically ergodic for any $i$ as well. Finally, the behavior policy is a double homogeneous policy, whose value is bounded away from zero. 
\end{asmp}

\begin{asmp}[Sample size]\label{as:samplesize}
    $N=O(T^{\ell})$ for any sufficiently large constant $\ell>0$. 
\end{asmp}

\begin{asmp}[Number of iterations]\label{as:FQIiter}
    The number of iterations in FQI is much larger than $\log(NT)/\log(\gamma^{-1})$. 
\end{asmp}

\begin{asmp}[Bounded reward]\label{as:reward}
    There exists some constant $R_{\max}<\infty$ such that $|R_{i,t}|\leq R_{\max}$ almost surely. 
\end{asmp}

\begin{asmp}[Completeness]\label{as:complete}
    For any $1\le k\le K$ and $Q$ that belongs to the Q-function class $\mathcal{Q}$, we have $\mathcal{B}_k^* Q\in \mathcal{Q}$.
\end{asmp}

\begin{asmp}[Q-function class]\label{as:VCclass}
    The estimated Q-function belongs to the VC type class \citep[see e.g., Definition 2.1,][]{cherno2014} with a finite VC index. Additionally, its envelop function is upper bounded by $R_{\max}/(1-\gamma)$. 
\end{asmp}

\subsection{An Auxiliary Lemma}
Recall that the log-likelihood function for a given set of indices $\mathcal{C}$ and a given time interval $[t_1,t_2]$ is given by
\begin{eqnarray*}
    \ell(\theta;\mathcal{C},[t_1,t_2])=\frac{1}{|\mathcal{C}|(t_2-t_1)}\sum_{i\in \mathcal{C}}\sum_{t=t_1+1}^{t_2}\log p(\triple,\theta).
\end{eqnarray*}
We use $\ell_0$ to denote its expectation, i.e., $\ell_0(\theta;\mathcal{C},[t_1,t_2])=\ell(\theta;\mathcal{C},[t_1,t_2])$ for a given $\mathcal{C}$ and $[t_1,t_2]$. Recall that $\widehat{\theta}_{\mathcal{C}_k,[t_1,t_2]}$ denotes the conditional maximum likelihood estimator. The following lemma establishes the uniform consistency and rate of convergence of these estimators. Its proof is similar to the one for standard maximum likelihood estimators \citep[see e.g.,][]{casella2024statistical}. The difference lies in that we consider MDP settings with dependent data while allowing the number of parameters to diverge to infinity as well. For completeness, we provide its proof in the following subsection.

\begin{lemma}[Uniform rate of convergence]\label{lemma:consistency}
Suppose MA, LHE, LSE and Assumptions \ref{as:tau}-\ref{as:ergodic} hold. Then for sufficiently large $T$, the set of estimated parameters $\{\widehat{\theta}_{\mathcal{C}_k, [t_1,t_2]}:t_2-t_1>\epsilon T,t_1\ge T-\tau^{(k)}\}$ satisfies
\begin{eqnarray*}
    \|\widehat{\theta}_{\mathcal{C}_k,[t_1,t_2]}-\theta_k^0\|=O\left(\frac{\sqrt{\log(NT)}}{\sqrt{N(t_2-t_1)}}\right),
\end{eqnarray*}
uniformly in all triplets $(k,t_1,t_2)$ such that $t_2-t_1>\epsilon T$ and $t_1\ge T-\tau^{(k)}$, with probability at least $1-O(N^{-1}T^{-1})$.
\end{lemma}

\subsection{Proof of Lemma \ref{lemma:consistency}}
\begin{proof}
Notice that for any $k$, under the given assumptions, it follows from Jensen's inequality that $\ell_0(\theta;\mathcal{C}_k,[t_1,t_2])$ is uniquely maximised at $\theta_k^0$ whenever $t_1\ge T-\tau^{(k)}$ \citep[see e.g.,][]{hogg1995introduction}.  The rest of the proof is divided into three steps. The first step is to establish the uniform consistency of log-likelihood function. The second step is to show the uniform consistency of the estimated parameters. The last step derives the rate of convergence. 

\subsubsection{Proof of Step 1}
In this step, we aim to establish the uniform convergence of $\lct[\widehat{\mathcal{C}}_k][t_1][t_2]$. That is, if Assumptions \ref{as:tau}-\ref{as:ergodic} hold, then with probability at least $1-O(N^{-1}T^{-1})$, 
\begin{equation*}
\max_{k,t_1,t_2}\sup\limits_{\bstheta}|\lct[\widehat{\mathcal{C}}_k][t_1][t_2] - \lcte[\widehat{\mathcal{C}}_k][t_1][t_2]|=O\left(\frac{\log(NT)}{\sqrt{N(t_2-t_1)}}\right),
\end{equation*}
where the first maximum is taken over all triplets $(k,t_1,t_2)$ such that $t_2-t_1>\epsilon T$ and $t_1\ge T- \tau^{(k)}$.

We will first prove the point-wise convergence of $\lct[\mathcal{C}_k][t_1][t_2]$ at a given parameter value $\theta$. Notice that the likelihood function $\lct[\mathcal{C}_k][t_1][t_2]$ can be decomposed into the sum of the following three terms:
\begin{eqnarray*}
    &&\frac{1}{|\mathcal{C}_k|(t_2-t_1)}\sum_{i\in \mathcal{C}_k}\sum_{t=t_1+1}^{t_2} [f(\triple,\theta)-\Mean_{(\bullet|)} f(\triple,\theta)]\\
    &+&\frac{1}{|\mathcal{C}_k|(t_2-t_1)}\sum_{i\in \mathcal{C}_k}\sum_{t=t_1+1}^{t_2} [\Mean_{(\bullet|)} [f(\triple,\theta)-\Mean \{f(\triple,\theta)|S_{it-1}\} ]\\
    &+&\frac{1}{|\mathcal{C}_k|(t_2-t_1)}\sum_{i\in \mathcal{C}_k}\sum_{t=t_1+1}^{t_2} [\Mean \{f(\triple,\theta)|S_{it-1}\}-\Mean f(\triple,\theta)],
\end{eqnarray*}
where $f$ is a shorthand for $\log p$ and $\Mean_{(\bullet|)}$ is a shorthand for the conditional expectation of the next state given the current state-action pair. In the following, we will use concentration inequalities designed for martingales and $\beta$-mixing processes to bound the first two lines and the third line, respectively. Specifically: 
\begin{enumerate}
    \item For the first line, a key observation is that, under the Markov assumption, the first line forms a sum of martingale difference sequence \citep[see e.g., Step 3 of the proof of Theorem 1 in][for a detailed illustration]{shi2020statistical}. In addition, it follows from Assumption \ref{as:f'' contineous} that $f$ is uniformly bounded away from infinity. As such, using the Azuma-Hoeffding's inequality\footnote{see e.g., \url{https://galton.uchicago.edu/~lalley/Courses/386/Concentration.pdf}.}, we can show that with probability at least $1-O(N^{-\kappa} T^{\kappa-2})$, for any sufficiently large constant $\kappa>0$, any $t_2-t_1>\epsilon T$ and any $1\le k\le K$, the absolute value of the first line is upper bounded by $C \sqrt{\kappa N(t_2-t_1)\log (NT)}$ with proper choice of the constant $C>0$. Using Bonferroni's inequality, we can show that the above event holds uniformly for any $t_1,t_2,k$ such that $t_2-t_1>\epsilon T$, with probability at least $1-O(N^{-\kappa} T^{-\kappa})$. 
    \item Next, using similar arguments, we can show the supremum of the absolute value of the second line over the triplet $(t_1,t_2,k)$ with the constraint that $t_2-t_1>\epsilon T$ is upper bounded by $O(\sqrt{\kappa N(t_2-t_1)\log (NT)})$ with probability at least $1-O(N^{-\kappa} T^{-\kappa})$. 
    \item Finally, consider the third line. Under Assumption \ref{as:f'' contineous}, for any $i\in \mathcal{C}_k$, both the probability density function of $S_{i,T-\tau^{(k)}}$ and the stationary probability density function of $\{S_{i,t}\}_{t\ge T-\tau^{(k)}}$ are bounded away from zero and infinity; see e.g., Part 2 of the proof of Lemma E.2 of \citet{shi2020statistical}. Together with Assumption \ref{as:ergodic}, it follows from Lemma 1 of \cite{meitz2019subgeometric} that $\{S_{i,t}\}_{t\ge T-t^{(k)}}$ is exponentially $\beta$-mixing. Denote the resulting $\beta$-mixing coefficient by $\{\beta(q)\}_q$. Similar to Theorem 4.2 of \cite{ChenChristensen2015}, we can show that, for any $t\geq 0$ and integer $1<q<T$,
    \begin{align}
    &\max\limits_{\substack{t_2-t_1>\epsilon T\\ t_1\ge T- \tau^k}}\mathbb{P}\left(|\ossumt[\mathcal{C}_k][t_1+1][t_2] \mathbb{E}\{f(\triple)|S_{it-1}\} - \mathbb{E} f(\triple)| \geq 6t\right)\nonumber\\
    &= \max\limits_{\substack{t_2-t_1>\epsilon T\\ t_1\ge T- \tau^k}}\mathbb{P}\big(|\sum\limits_{(i,t) \in I_r} \mathbb{E} \{f(\triple)|S_{it-1}\}- \mathbb{E} f(\triple)|\geq t \big)\nonumber\\
    &+O(1)\frac{|\mathcal{C}_k|(t_2-t_1)}{q }\beta(q) + O(1)\exp\left(\frac{-t^2/2}{|\mathcal{C}_k|(t_2-t_1)qM^2 + qM t/3}\right),\label{eq:step2 mixing middle}
    \end{align}
    where $O(1)$ denotes some positive constant, $I_r = \{q\left \lceil{|\mathcal{C}_k|(t_2-t_1)/q}\right \rceil, q\left \lceil{|\mathcal{C}_k|(t_2-t_1)/q}\right \rceil +1, \dots, |\cC_k|(t_2-t_1+1)-1 \}$ and $2f$ is uniformly upper bounded by $M$. Suppose $t>5qM$. Notice that $|I_r|\leq q$. We have
    \begin{equation*}
    P\left(|\sum\limits_{(i,t) \in I_r} \mathbb{E} \{f(\triple)|S_{it-1}\}-\mathbb{E} f(\triple)|\geq t \right)=0.
    \end{equation*}
    Under exponential $\beta$-mixing, we have $\beta(q) = O(\rho^q)$ for some positive constant $\rho<1$. Set $q = -(\kappa+1)\log(|\mathcal{C}_k|(t_2-t_1))/\log \rho$, we obtain $|\mathcal{C}_k|(t_2-t_1)\beta(q)/q = O(N^{-\kappa}T^{-\kappa})$ under Assumption \ref{as:tau}. Set $t =\max\{\sqrt{2(\kappa+2) |\mathcal{C}_k|(t_2-t_1)qM^2\log(|\mathcal{C}_k|(t_2-t_1))}, (\kappa+2) qM \log(|\mathcal{C}_k|(t_2-t_1))\}$, we obtain that
    \begin{equation*}
    \frac{t^2}{2}\geq (\kappa+2)|\mathcal{C}_k|(t_2-t_1)qM^2\log(|\mathcal{C}_k|(t_2-t_1))\ \mathrm{and}\ \frac{t^2}{2}\geq (\kappa+2) qM\log(|\mathcal{C}_k|(t_2-t_1))\frac{t}{3} \ \mathrm{and}\ t\gg qM,
    \end{equation*}
    as either $N\rightarrow \infty$ or $T\rightarrow \infty$. Thus, it follows from \eqref{eq:step2 mixing middle} that the absolute value of the third line is upper bounded by $O(\sqrt{N(t_2-t_1)}\log (NT))$ with probability at least $1-O(N^{-\kappa-2} T^{-\kappa-2})$. By Bonferroni's inequality, we can show that this event holds uniformly for any triplet $(t_1,t_2,k)$ such that $t_2-t_1>\epsilon T$ with probability $1-O(N^{-\kappa}T^{-\kappa})$. 
\end{enumerate}

To summarize, we have shown that with probability at least $1-O(N^{-\kappa}T^{-\kappa})$, 
\begin{align}\label{eqn:proofstep1summary}
    |\lct[\mathcal{C}_k][t_1][t_2] - \lcte[\mathcal{C}_k][t_1][t_2]|=O\left(\frac{\log(NT)}{\sqrt{N(t_2-t_1)}}\right),
\end{align}
for all triplets $(k,t_1,t_2)$ such that $t_2-t_1>\epsilon T$ and $t_1\ge T-\tau^{(k)}$. 
This proves the pointwise convergence of the log-likelihood function as either $N$ or $T$ diverges to infinity. 

To establish the uniform convergence, consider open balls of radius $\delta=N^{-1}T^{-1}$ around $\bstheta \in \Theta$, i.e., $B(\bstheta, \delta) = \{\widetilde{\bstheta}:\|\bstheta - \widetilde{\bstheta}\|<\delta\}$ where $\|\bullet\|$ denotes the Euclidean norm. Let $d$ denote the dimension of $\theta$. Since $\Theta$ is a compact set, it follows from Lemma 2.2 of \citet{mendelson2008uniform} that there exist some $\Theta_{\delta}$ with number of elements upper bounded by $O(N^dT^d)$ such that the union of the open balls $\cup \{B(\bstheta, \delta): \bstheta\in \Theta_{\delta}\}$ covers $\Theta$. Denote these open balls by $\{B(\bstheta^j,\delta),j=1,\dots,J\}$. It follows from the triangle inequality that
\begin{align}
\label{eq:ball1}
&|\lct[\mathcal{C}_k][t_1][t_2][\bstheta] - \lcte[\mathcal{C}_k][t_1][t_2][\bstheta]| \nonumber\\
&\leq  | \lct[\mathcal{C}_k][t_1][t_2][\bstheta] -\lct[\mathcal{C}_k][t_1][t_2][\bstheta^j] |\\
\label{eq:ball2}
&  +|\lct[\mathcal{C}_k][t_1][t_2][\bstheta^j] -\lcte[\mathcal{C}_k][t_1][t_2][\bstheta^j]|\\
\label{eq:ball3}
& + |\lcte[\mathcal{C}_k][t_1][t_2][\bstheta^j] - \lcte[\mathcal{C}_k][t_1][t_2][\bstheta]|.
\end{align}
For a given $\theta$, set $\bstheta^j$ such that $\bstheta\in B(\bstheta^j,\delta)$. Under Assumption \ref{as:f'' contineous}, the log-likelihood function is Lipschitz continuous. As such, \eqref{eq:ball1} can be upper bounded by $L\delta$ for some constant $L>0$. Similarly, \eqref{eq:ball3} can be upper bounded by $L\delta$ as well. Finally, by setting the constant $\kappa$ to $d+1$, it follows from \eqref{eqn:proofstep1summary} and Bonferroni's inequality that \eqref{eq:ball2} is of the order $O((N(t_1-t_1))^{-1/2}\log(NT))$, with this big-O term being uniform across all $j$. 
The proof for Step 1 is hence completed.

\subsubsection{Proof of Step 2} 
In this step, we aim to show that for any positive $\varepsilon>0$ and sufficiently large $T$, the event $\max_{k,t_2-t_1>\epsilon T}\|\widehat{\theta}_{\mathcal{C}_k,[t_1,t_2]}-\theta_k^0\|\le \varepsilon$ holds with probability at least $1-O(N^{-1}T^{-1})$. 

Consider the objective function $\ell_0(\theta, \mathcal{C}_k, [t_1,t_2])$. For any $t_1\ge T-\tau^{(k)}$, under Assumption \ref{as:ergodic}, for sufficiently large $t_2$, the distribution of the state $S_{it_2}$ will converge to its limiting distribution. As discussed in Step 1 of the proof, the process $\{S_{i,t}\}_{t\ge T-\tau^{(k)}}$ is exponentially $\beta$-mixing. According to the definition of the $\beta$-mixing coefficient, we have 
\begin{eqnarray*}
    \beta(q)=\int_s \sup_{0\le \varphi\le 1} \left|\mathbb{E} [\varphi(S_{iq+t_1})|S_{it_1}=s]-\int \varphi(s)\mu(s)ds \right|\mu(s)ds,
\end{eqnarray*}
where $\mu$ denotes the density function of the limiting distribution. Since $p$ is well bounded away from zero and infinity, so are the marginal distributions of $S_{it_1}$ and $\mu$. Consequently, there exists some universal constant $C>0$ such that
\begin{eqnarray*}
    \int_s \sup_{0\le \varphi\le 1} \left|\mathbb{E} [\varphi(S_{iq+t_1})|S_{it_1}]-\int \varphi(s)\mu(s)ds \right|\le C\beta(q).
\end{eqnarray*}
Under exponentially $\beta$-mixing, this immediately implies that $\ell_0(\theta,\mathcal{C}_k,[t_1,t_2])\to \ell_0^{\infty}(\theta,\mathcal{C}_k)$ whenever $t_1\ge T-\tau^{(k)}$ and $t_2-t_1\ge \kappa T$, as $T\to \infty$. Here, $\ell_0^{\infty}(\theta,\mathcal{C}_k)=\int_s \Mean \log p(S_{it_1+1}|S_{it_1}=s,A_{it_1},\theta)\mu(s)ds$ for any $i\in \mathcal{C}_k$. Moreover, the convergence is uniform in $k$, $t_1$ and $t_2$. This together with the proof for Step 1 yields the uniform convergence of $\ell(\theta,\mathcal{C}_k,[t_1,t_2])$ to $\ell_0^{\infty}(\theta,\mathcal{C}_k)$. 

Under the regularity conditions in Assumption \ref{as:f'' contineous}, for each $k$, $\ell_0^{\infty}(\theta,\mathcal{C}_k)$ is uniquely maximised at $\theta_k^0$. In addition, it is a continuous function of $\theta$. Since the parameter space is compact, $\ell_0^{\infty}(\theta_k^0,\mathcal{C}_k)$ is strictly larger than $\sup_{\|\theta-\theta_k^0\|\le \varepsilon}\ell_0^{\infty}(\theta,\mathcal{C}_k)$. This together with the uniform consistency of $\ell(\theta,\mathcal{C}_k,[t_1,t_2])$ established in Step 1 
yields the uniform consistency of the estimated parameters. 
\end{proof}

\subsubsection{Proof of Step 3}\label{sec:proofrate}
\begin{proof}
By Taylor expansion, we obtain that
\begin{equation*}
0=\lctfirst[\mathcal{C}_k][t_1][t_2][\widehat{\bstheta}_{\mathcal{C}_k,[t_1,t_2]}] = \lctfirst[\mathcal{C}_k][t_1][t_2][\theta_k^0]+\int_0^1\ell^{''}(t\widehat{\bstheta}_{\mathcal{C}_k,[t_1,t_2]}+(1-t)\theta_k^0;\mathcal{C}_k, [t_1, t_2])dt(\widehat{\bstheta}_{\mathcal{C}_k,[t_1,t_2]} - \theta_k^0),
\end{equation*}
for some $\theta_k^*$ lying on the line segment joining $\widehat{\bstheta}_{\mathcal{C}_k,[t_1,t_2]}$ and $\theta_k^0$. It follows that 
\begin{align}
\|\widehat{\bstheta}_{\mathcal{C}_k,[t_1,t_2]} - \theta_k^*\| \le
 \left[\lambda_{\min}[-\int_0^1\ell^{''}(t\widehat{\bstheta}_{\mathcal{C}_k,[t_1,t_2]}+(1-t)\theta_k^0;\mathcal{C}_k, [t_1, t_2])dt]\right] ^{-1} \|\ell^{'}(\theta_k^0;\mathcal{C}_k, [t_1, t_2])\|.\label{eq:cluster_theta_taylor}
\end{align}
It remains to bound the two terms on the right-hand-side (RHS) of \eqref{eq:cluster_theta_taylor}. 

First, consider the first term on the RHS of \eqref{eq:cluster_theta_taylor}. Under Assumption \ref{as:f'' contineous}, both the transition function $p$ and the marginal state density function is lower bounded by some constant $c>0$. This together with the minimum eigenvalue assumption in Assumption \ref{as:f'' contineous} implies that
\begin{eqnarray*}
    \min_{\theta}\lambda_{\min}\left[-\Mean \frac{\partial^2 \log p(\triple,\theta)}{\partial \theta\theta^\top}\right]\ge \min_{a,\theta} \lambda_{\min}\left[-\Mean \frac{\partial^2 \log p(S_{i,t}|S_{it-1},a,\theta)}{\partial \theta\theta^\top}\right]\\
    \ge c^2 \min_{a,\theta} \lambda_{\min}\left[-\int_{s,s'} \frac{\partial^2 \log p(s'|s,a,\theta)}{\partial \theta\theta^\top}dsds'\right]
\end{eqnarray*}
is well bounded away from zero. Similarly, the minimum eigenvalue of $-\Mean \ell^{''}(\theta;\mathcal{C}_k, [t_1, t_2])$ is uniformly bounded away from zero as well. In addition, similar to Lemma \ref{lemma:consistency}, we can show that the set of difference $\{\ell^{''}(\theta;\mathcal{C}_k, [t_1, t_2])-\Mean \ell^{''}(\theta;\mathcal{C}_k, [t_1, t_2]):t_1\ge T-\tau^{(k)},t_2,k,\theta\}$ converge uniformly to $0$ with probability at least $1-O(N^{-1}T^{-1})$. As such, the minimum eigenvalue of $-\int_0^1\ell^{''}(t\widehat{\bstheta}_{\mathcal{C}_k,[t_1,t_2]}+(1-t)\theta_k^0;\mathcal{C}_k, [t_1, t_2])dt$ is uniformly bounded away from zero, with probability at least $1-O(N^{-1}T^{-1})$. Equivalently, the first term on the RHS of \eqref{eq:cluster_theta_taylor} is upper bounded by some positive constant. 

As for the second term, notice that 
\begin{eqnarray*}
    \Mean_{(\bullet|)}\frac{\partial \log p(\triple,\theta_k^0)}{\partial \theta}=0,
\end{eqnarray*}
whenever $t>T-\tau^{(k)}$ and $i\in \mathcal{C}_k$. As such, the second term forms a sum of martingale difference sequence. Using similar arguments in Step 1 of the proof of  Lemma \ref{lemma:consistency}, it follows from the martingale concentration inequality that the supermum of the second term on the RHS of \eqref{eq:cluster_theta_taylor} over all triplets $(t_1,t_2,k)$ such that $t_1\ge T-\tau^{(k)}, t_2-t_1> \epsilon T$ is upper bounded by $O(N^{-1/2}(t_2-t_1)^{-1/2}\sqrt{\log (NT)})$, with probability $1-O(N^{-1}T^{-1})$. This together with the uniform upper bound for the first term yields the desired uniform rate of convergence. 
\end{proof}

\subsection{Proof of Theorem \ref{thmrate}}
Notice that Theorem \ref{thmrate} is automatically implied by the following three lemmas. We focus on proving these lemmas one by one in this section. 
\begin{lemma}\label{lemma1}
Suppose MA, LSE, LHE, Assumptions \ref{as:tau} -- \ref{as:f'' contineous}, \ref{as:ergodic} hold and $s_{cp}\gg (NT)^{-1/2} \log^{3/2} (NT)$. For sufficiently large $T$, when using the oracle cluster memberships as input, the estimated change points computed by the proposed most recent change point detection subroutine satisfy 
\begin{equation*} 
    \max_i \frac{|\widehat{\tau}_i^*-\tau_i^*|}{\tau_i^*}=O\left[\frac{\log^3(NT)}{NT s_{cp}^2}\right], \tag{\ref{eqn:cperr}}
\end{equation*}
with probability $1-O(N^{-1}T^{-1})$. 
\end{lemma}
\begin{lemma}\label{lemma2}
Suppose MA, LSE, LHE and Assumptions \ref{as:tau} -- \ref{as:f'' contineous}, \ref{as:ergodic} and \ref{as:samplesize} hold. Suppose the initial estimators satisfy $\max_i [\tau_i^0-\tau_i^*]_+/\tau_i^*\ll s_{cl}$,  $s_{cl}\gg T^{-1/2}\sqrt{\log (NT)}$, $\min_i \tau_i^0\ge \kappa T$ for some constant $\kappa>0$, and the number of cluster $K$ is correctly specified. Then for sufficiently large $T$, the estimated cluster memberships based on the proposed clustering subroutine achieves a zero clustering error, with probability at least $1-O(N^{-1}T^{-1})$. 
\end{lemma}
\begin{lemma}\label{lemma3}
Suppose MA, LSE, LHE and Assumptions \ref{as:tau} -- \ref{as:f'' contineous}, \ref{as:ergodic} and \ref{as:samplesize} hold. Suppose the initial estimators satisfy $\max_i [\tau_i^0-\tau_i^*]_+/\tau_i^*\ll T^{-1/2}\sqrt{\log(NT)}$, $s_{cl}\gg \max(T^{-1/4},\log^{3/2}(NT)/(\sqrt{NT}s_{cp}))$, $\min_i \tau_i^0\ge \kappa T$ for some constant $\kappa>0$.  
Then the proposed IC correctly identifies $K$, with probability at least $1-O(N^{-1}T^{-1})$. 
\end{lemma}
\subsubsection{Proof of Lemma \ref{lemma1}}
We first prove Lemma \ref{lemma1}. Notice that the clustering error equals exactly zero, with probability at least $1-O(N^{-1}T^{-1})$. Lemma \ref{lemma1} thus implies that at each iteration, the estimated $\{\widehat{\tau}_i^*\}_i$ will converge at a rate of \eqref{eqn:cperr},  with probability at least $1-O(N^{-1}T^{-1})$. Additionally, the condition on $s_{cp}$ is automatically implied by Assumption \ref{as:signal}. 

\begin{proof}
The proof is divided into two steps. In the first step, we aim to show that for all $\tau<\tau^{(k)}$ and $k$, the threshold used in the likelihood ratio test, as a function of the sample size $|\cC_k|\tau$, is greater than the maximum log-likelihood ratio statistics with probability $1-O(N^{-1}T^{-1})$. This implies that our method will not underestimate $\tau^{(k)}$.   

Recall that for a given candidate change point location $u$, the loglikelihood is given by 
\begin{equation*}
\begin{aligned}
&\mathrm{LR}(\cC_k, [T-\tau, T],u) = -\ossumo f(\triple, \widehat{\bstheta}_{\cC_k, [T-\tau, T]})\\
&+ \ossumo[\cC_k][T-\tau+1][u] f(\triple, \widehat{\bstheta}_{\cC_k,[T-\tau,u-1]}) + \ossumo[\cC_k][u+1] f(\triple, \widehat{\bstheta}_{\cC_k,[u,T]}). 
\end{aligned}
\end{equation*}

To simplify the notation, let $\widehat{\bstheta}_k^{null}= \widehat{\theta}_{\mathcal{C}_k, [T-\tau, T]}$, 
$\widehat{\bstheta}_k^1=\widehat{\theta}_{\mathcal{C}_k, [T-\tau, u-1]}$, and $\widehat{\bstheta}_k^2=\widehat{\theta}_{\mathcal{C}_k, [u, T]}$. Using Taylor expansion, we obtain that
\begin{equation*}
\begin{aligned}
\mathrm{LR}(\cC_k, [T-\tau, T],u)& = \ossumo[\cC_k][T-\tau+1][u]f(\triple,\tkt) \\
&- (\widehat{\bstheta}_k^1-\tkt)^\top \ossumo[\cC_k][T-\tau+1][u]f^{''}(\triple,\tk^{1,*})(\widehat{\bstheta}_k^1-\tkt)\\
&+\ossumo[\cC_k][u+1]f(\triple,\tkt)\\
& - (\widehat{\bstheta}_k^2-\tkt)^\top\ossumo[\cC_k][u+1]f^{''}(\triple,\tk^{2,*})(\widehat{\bstheta}_k^2-\tkt)\\
& - \ossumo f(\triple,\tkt)\\
&+  (\widehat{\bstheta}_k^{null}-\tkt)^\top\ossumo[\cC_k] f^{''}(\triple,\tk^{n,*})(\widehat{\bstheta}_k^{null}-\tkt),\\
\end{aligned}
\end{equation*}
for some $\tk^{1,*}$, $\tk^{2,*}$, $\tk^{n,*}$ that lie on the line segments joining $\theta_k^0$ and $\widehat{\theta}_k^1$, $\theta_k^0$ and $\widehat{\theta}_k^2$, $\theta_k^0$ and $\widehat{\theta}_k^{null}$, respectively. It suffices to analyse the second, fourth and last lines in the above expression. Below, we analyse them one by one:
\begin{itemize}
    \item Under Assumption \ref{as:tau}, it follows from Lemma \ref{lemma:consistency} that with probability at least $1-O(N^{-1}T^{-1})$, the difference between $\widehat{\theta}_k^{null}$ and $\theta_k^0$ is upper bounded by $O(N^{-1/2}T^{-1/2}\sqrt{\log (NT)})$. Under the boundedness assumption on the second-order derivatives, the last line is upper bounded by $O(\log (NT))$, with probability at least $1-O(N^{-1}T^{-1})$. 
    \item Similarly, for any candidate change point location $u$ such that $u+\tau-T\ge \epsilon T$, the difference between $\widehat{\theta}_k^2$ and $\theta_k^0$ is upper bounded by $O(N^{-1/2}(u+\tau-T)^{-1/2}\sqrt{\log (NT)})$. Hence, the second term is upper bounded by $O(\log(NT))$. In cases where $N$ is finite and $u+\tau-T<\epsilon T$, it follows from the boundedness of the parameter space in Assumption \ref{as:compact} and the definition of $\epsilon$ that the second line is upper bounded by $O(\epsilon^2T^2)=O(\log^2(NT)\log(\log(NT)))$. 
    \item Using similar arguments in the second bullet point, we can show that the fourth term is upper bounded by $O(\log^2(NT)\log(\log(NT)))$ as well. 
\end{itemize}
To summarize, we have shown that the likelihood ratios are uniformly upper bounded by $O(\log^2 (NT)\log(\log (NT)))$, with probability at least $1-O(N^{-1}T^{-1})$. According to Assumption \ref{as:tau}, $|\mathcal{C}_k|\tau$ approaches infinity as $NT\to \infty$. It follows from Section \ref{sec:sim:implementation:cp} that the threshold is much larger than the maximum likelihood ratio. This completes the proof for the first step. 

In the second step, we show that the test statistics would exceed the threshold if \begin{equation*}
\tau= \tau^{(k)}+N^{-1}\log^3(NT)s_{cp}^{-2}.
\end{equation*}
Consider the log-likelihood ratio LR$(\mathcal{C}_k, [T-\tau, T], \tau^{(k)})$. Similarly, it follows from Taylor expansion that $\mathrm{LR}(\mathcal{C}_k, [T-\tau, T], \tau^{(k)})$ equals
\begin{align}
& \ossumo[\cC_k][T-\tau+1][T-\tau^{(k)}]f(\triple,\tk^{1})
- \frac{1}{2}(\widehat{\bstheta}_k^1 - \tk^{1})^\top\ossumo[\cC_k][T-\tau+1][T-\tau^{(k)}] f^{''}(\triple, \tk^{1,*})(\widehat{\bstheta}_k^1 - \tk^{1})\nonumber\\
&
+\ossumo[\cC_k][T-\tau^{(k)}+1]f(\triple,\tk^0)
- \frac{1}{2}(\widehat{\bstheta}_k^2 - \tk^0)^\top\ossumo[\cC_k][T-\tau^{(k)}+1] f^{''}(\triple, \tk^{0,*})(\widehat{\bstheta}_k^2 - \tk^0)\nonumber \\
&-\ossumo[\cC_k]f(\triple,\tk^0)+\frac{1}{2}(\widehat{\bstheta}_k^{null} - \tk^0)^\top\ossumo[\cC_k] f^{''}(\triple, \tk^{n,*})(\widehat{\bstheta}_k^{null} - \tk^0)], \label{eq:thm4.1 h0}
\end{align}
for some $\tk^{1,*}$, $\tk^{2,*}$, $\tk^{n,*}$ that lie on the line segments joining $\theta_k^{1}$ and $\widehat{\theta}_k^1$, $\theta_k^0$ and $\widehat{\theta}_k^0$, $\theta_k^0$ and $\widehat{\theta}_k^{null}$, respectively, all converging to $\theta_k^0$ as  $\theta_k^1$ is asymptotically equivalent to $\theta_k^0$.

Meanwhile, using similar arguments to the proof of Lemma \ref{lemma:consistency}, we obtain 
that the minimum eigenvalues of $\ossumo[\cC_k][T-\tau+1][T-\tau^{(k)}] f^{''}(\triple, \tk^{1,*})/(|\cC_k|(\tau-\tau^{(k)}))$ and $\ossumo[\cC_k][T-\tau^{(k)}+1][T] f^{''}(\triple, \tk^{0,*})/(|\cC_k|\tau^{(k)})$ are  bounded away from zero, with probability at least $1-O(N^{-1}T^{-1})$. This together with the boundedness of $f''$ yields that
\begin{eqnarray}\label{eqn:someinequality}
\eqref{eq:thm4.1 h0}\geq  \ossumo[\cC_k][T-\tau+1][T-\tau^{(k)}]\log \frac{p(\triple, \tk^{1})}{p(\triple, \bstheta_k^{0})} 
-C|\mathcal{C}_k|\tau\|\widehat{\bstheta}_k^{null} - \bstheta_k^0\|^2,
\end{eqnarray}
for some constant $C>0$. 

Using similar arguments in the proof of Lemma \ref{lemma:consistency}, we can show that the the convergence rate $\|\widehat{\bstheta}_k^{null} - \bstheta_k^0\|_2^2$ is proportional to the order of magnitude of 
\begin{eqnarray}\label{eqn:someinequality1}
    \left\|\frac{1}{|\mathcal{C}_k|\tau}\sum_{i\in \mathcal{C}_k}\sum_{t=T-\tau+1}^{T-\tau^{(k)}} f'(\triple,\theta_k^0) \right\|_2^2+\left\|\frac{1}{|\mathcal{C}_k|\tau}\sum_{i\in \mathcal{C}_k}\sum_{t=T-\tau^{(k)}+1}^T f'(\triple,\theta_k^0) \right\|_2^2,
\end{eqnarray}
by Cauchy-Schwarz inequality. Notice that according to the Azuma Hoeffding's inequality, the second term in \eqref{eqn:someinequality1} is $O(N^{-1}T^{-1}\log (NT))$, with probability $1-O(N^{-1}T^{-1})$. As for the first term, using similar arguments in the proof for Step 1 of Lemma \ref{lemma:consistency}, we can show that it is of the order of magnitude of
\begin{eqnarray*}
    \frac{1}{N^2T^2}\left[\ossumo[\cC_k][T-\tau+1][T-\tau^{(k)}]\Mean f'(\triple, \tk^{0})\right]^2 +\frac{(\tau-\tau^{(k)})\log^2 (NT)}{N^2T^2}\\=O\left(\frac{(\tau-\tau^{(k)})^2}{T^2}\|\theta_k^1-\theta_k^0\|^2\right)+\frac{(\tau-\tau^{(k)})\log^2 (NT)}{N^2T^2},
\end{eqnarray*}
with probability $1-O(N^{-1}T^{-1})$. It follows from \eqref{eqn:someinequality} that the log-likelihood ratio is larger than or equal to
\begin{eqnarray*}
\ossumo[\cC_k][T-\tau+1][T-\tau^{(k)}]\log \frac{p(\triple, \tk^{1})}{p(\triple, \bstheta_k^{0})}-\frac{c(\tau-\tau^{(k)})\log^2 (NT)}{NT}-\frac{cN(\tau-\tau^{(k)})^2}{T}\|\theta_k^1-\theta_k^0\|^2,
\end{eqnarray*}
for some constant $c>0$.

Next, using similar arguments to Step 1 of the proof of  Lemma \ref{lemma:consistency}, we can show that the first term in the above expression is larger than or equal to
\begin{eqnarray*}
 \left[\ossumo[\cC_k][T-\tau+1][T-\tau^{(k)}]\Mean\log \frac{p(\triple, \tk^{1})}{p(\triple, \bstheta_k^{0})}\right]-O\left(\|\tk^{1}-\theta_k^0\|\sqrt{N(\tau-\tau^{(k)})}  \log(NT)\right),
\end{eqnarray*}
with probability $1-O(N^{-1}T^{-1})$. 

Moreover, under Assumption \ref{as:f'' contineous}, using a second order Taylor expansion and similar arguments in bounding the first term on the RHS of \eqref{eq:cluster_theta_taylor}, we can show that,
\begin{eqnarray*}
\ossumo[\cC_k][T-\tau+1][T-\tau^{(k)}]\Mean\log \frac{p(\triple, \tk^{1})}{p(\triple, \bstheta_k^{0})}\ge c\lambda N(\tau-\tau^{(k)}) \|\tk^{1}-\theta_k^0\|^2,
\end{eqnarray*}
for some constant $c>0$. Combining these results together, it is immediate to see that under the given specification on $\tau$ and the signal strength condition that $s_{cp}\gg (NT)^{-1/2}\log^{3/2}(NT)$, the likelihood ratio is strictly larger than $0$ and is strictly larger than the threshold with probability at least $1-O(N^{-1}T^{-1})$. This completes the proof for the second step. Therefore, the change point detection procedure will stop as long as $\tau\ge \tau^{(k)}+N^{-1}s_{cp}^{-2}\log^3(NT)$. This yields the desired rate of convergence. 
\end{proof}
\subsubsection{Proof of Lemma \ref{lemma2}}
We next prove Lemma \ref{lemma2}. Suppose Lemma \ref{lemma2} is proven. Under the conditions that $s_{cl}\gg (NT)^{-1/2}s_{cp}^{-1}\log^{3/2}(NT)$ and $s_{cp}\gg N^{-1/2}T^{-1/4}\log^{3/2}(NT)$ in Assumption \ref{as:signal}, it follows that $s_{cl}$ is much larger than the change point detection error in \eqref{eqn:cperr}. Consequently, when $K$ is correctly specified, it follows that the clustering error will be zero with probability at least $1-O(N^{-1}T^{-1})$ during each iteration --- not just at the initial iteration. 
\begin{proof}
We use $k(\bullet)$ to denote a given mapping from the indices of subjects $\{1,\cdots,N\}$ to the indices of clusters $\{1,\cdots,K\}$. Let $\mathcal{K}$ denote the set $\{k(i)\}_i$. For a given set of parameters $\theta=\{\theta_k\}_k$, define 
\begin{equation*}
Q(\bstheta, \mathcal{K}) = \frac{1}{N}\sum_{i =1}^N \frac{1}{\widehat{\tau}_i} \sum_{t=T-\widehat{\tau}_i+1}^T \log p(S_{i, t}|A_{i, t-1}, S_{i, t-1} ; \theta_{k(i)}).
\end{equation*}
Let $(\widehat{\bstheta}, \widehat{\mathcal{K}})=\argmax Q(\bstheta, \mathcal{K})$ where $\widehat{\mathcal{K}}=\{\widehat{k}_i\}_i$. For a given value of $\bstheta$, define the optimal group assignment for each unit as
\begin{equation*}
\widehat{k}_i(\bstheta) = \argmax_{k \in \{1,\dots,K\}} \sum\limits_{t=T-\widehat{\tau}_i^*+1}^{T} \log p(\triple, \bstheta_k). 
\end{equation*}
For conciseness, we write $\widehat{k}_i(\widehat{\bstheta})$ as $\widehat{k}_i $, let $k_i^0$ denote the oracle group assignment for the $i$th unit and $\mathcal{K}^0=\{k_i^0\}_k$. Let $\theta^0=\{\theta_k^0\}_k$ denote the set of oracle parameters. 

In the first step, we establish the rate of convergence of the estimated parameters $N^{-1}\sum \|\widehat{\theta}_{\widehat{k}_i}-\theta^0_{k_i^0}\|^2$. 
It follows that 
\begin{eqnarray}\label{eqn:taylor}
\begin{aligned}
0 & \ge Q(\bstheta^0, \mathcal{K}^0) - Q(\widehat{\bstheta},\widehat{\mathcal{K}}) \\
& = \cssum[\tau^0_i][\min(\tau_i^*,\tau_i^0)+1][][f(\triple,\btgit^0)-f(\triple,\btgih)] \\
& + \cssum[\tau^0_i][\tau^0_i+1][-\min(\tau_i^*,\tau_i^0)][f(\triple,\btgit^0)-f(\triple,\btgih)]\mathbb{I}(\tau_i^0> \tau_i^*).
\end{aligned}
\end{eqnarray}
By Taylor expansion, the second line equals
\begin{eqnarray*}
\cssum[\tau_i^0][\min(\tau^*_i,\tau_i^0)+1][] \left[- f^{'}(\triple,\btgit^0)^\top (\btgih-\btgit^0)\right.\\
\left. - \frac{1}{2}(\btgit^0-\btgih)^\top f^{''}(\triple,\bstheta_i^*)(\btgit^0-\btgih)\right],
\end{eqnarray*}
for some $\theta_i^*$ that lies on the line segment joining $\btgit^0$ and $\btgih$. 

Under the given conditions, both $\tau_i^*$ and $\tau_i^0$ are proportional to $T$. As $T\to 
\infty$, under the minimum eigenvalue condition in Assumption \ref{as:f'' contineous}, using similar arguments to Step 1 of the proof of Lemma \ref{lemma:consistency}, we can show that the minimum eigenvalues of the matrices $\{-(\tau_i^0)^{-1}\sum_{t=T-\min(\tau_i^*,\tau_i^0)+1}^T f^{''}(\triple,\bstheta_i^*)\}_i$ are uniformly bounded away from zero with probability $1-O(N^{-1}T^{-1})$. In addition, using Azuma Hoeffding's inequality, sums of the scores $\{\|\sum_{t=T-\min(\tau_i^*,\tau_i^0)+1}^T f'(\triple,\theta_{k_i^0}^0)\|\}_i$ can be uniformly upper bounded by $\sqrt{T\log(NT)}$, with probability $1-O(N^{-1}T^{-1})$. To summarise, we have shown that the second line of \eqref{eqn:taylor} is lower bounded by 
\begin{eqnarray*}
    \frac{c}{N}\sum_{i=1}^N \|\btgih-\btgit^0\|^2-\frac{C\sqrt{\log (NT)}}{N\sqrt{T}} \sum_{i=1}^N \|\btgih-\btgit^0\|.
\end{eqnarray*}
with probability at least $1-O(N^{-1}T^{-1})$, for some constants $c,C>0$. 

Next, consider the third line of \eqref{eqn:taylor}. Under Assumption \ref{as:f'' contineous}, the derivative $f'$ is uniformly bounded. Under (A1), $\tau_i^0$ is proportional to $\tau_i^*$ for any $i$. As such, the third line can be lower bounded by
\begin{eqnarray*}
    -\frac{O(1)}{N}\sum_{i=1}^N \frac{[\tau_i^0-\tau_i^*]_+}{\tau_i^*}\|\btgih-\btgit^0\|,
\end{eqnarray*}
where $O(1)$ denotes some positive constant whose value is allowed to vary from place to place. 

It follows from \eqref{eqn:taylor} that with probability $1-O(N^{-1}T^{-1})$,
\begin{eqnarray*}
    \frac{c}{N}\sum_{i=1}^N \|\btgih-\btgit^0\|^2\le \frac{O(1)}{N}\sum_{i=1}^N \left(\frac{[\tau_i^0-\tau_i^*]_+}{\tau_i^*}+\frac{\sqrt{\log (NT)}}{\sqrt{T}}\right)\|\btgih-\btgit^0\|,
\end{eqnarray*}
for some positive constant denoted by $O(1)$. Using Cauchy-Schwarz inequality, it is immediate to see that with probability $1-O(N^{-1}T^{-1})$, we have
\begin{eqnarray}\label{eqn:step1assertion}
    \frac{1}{N}\sum_{i=1}^N \|\btgih-\btgit^0\|^2\le O(1)\frac{\log (NT)}{T}+O(1)\frac{1}{N}\sum_{i=1}^N \frac{[\tau_i^0-\tau_i^*]_+^2}{(\tau_i^*)^2}. 
\end{eqnarray}
This completes the proof of the first step. 

In the second step, we aim to show that the clustering algorithm achieves a zero clustering error, with probability $1-O(N^{-1}T^{-1})$. Toward that end, we notice that under the current conditions, the signal strength $s_{cl}$ is much larger than the square root of the RHS of \eqref{eqn:step1assertion}. Since $K$ is correctly specified, using similar arguments in the proof of Lemma B.3 in \cite{ECTA11319}, we can show the existence of a permutation $\sigma(\bullet): \{1,\cdots,K\}\rightarrow \{1,\cdots,K\}$ such that for each $k$,
\begin{eqnarray}\label{eqn:step2assertion1}
    \|\widehat{\theta}_{\sigma(k)}-\theta_k^0\|^2\le O(1)\frac{\log (NT)}{T}+O(1)\frac{1}{N}\sum_{i=1}^N \frac{[\tau_i^0-\tau_i^*]_+^2}{(\tau_i^*)^2}.
\end{eqnarray}
Without loss of generality, assume $\sigma$ is an identity function such that $\sigma(k)=k$ for any $k$. Notice that at this point, we have shown that the clustering error decays to zero. Below, we show that it is exactly zero with probability $1-O(N^{-1}T^{-1})$. A key observation is that, since the estimated cluster membership maximises the log-likelihood function, we have for each $i$ that
\begin{eqnarray*}
    \sum_{t=T-\tau_i^0+1}^T f(\triple, \widehat{\theta}_{\widehat{k}_i})\ge \sum_{t=T-\tau_i^0+1}^T f(\triple, \widehat{\theta}_{k_i^0}).
\end{eqnarray*}
Similar to Step 1 of the proof, this implies that with probability at least $1-O(N^{-1}T^{-1})$, we have for any $i$ that
\begin{eqnarray}\label{eqn:step2assertion2}
    \|\widehat{\theta}_{\widehat{k}_i}-\widehat{\theta}_{k_i^0}\|^2\le C\frac{(\tau_i^0-\tau_i^*)_+^2}{(\tau_i^*)^2}+C \left\| \frac{1}{\tau_i^0}\sum_{t=T-\min(\tau_i^0, \tau_i^*)+1 }^Tf^{'}(\triple, \widehat{\theta}_{k_i^0}) \right\|^2,
\end{eqnarray}
for some positive constant $C>0$. Using Taylor expansion and Cauchy-Schwarz inequality, the second term on the RHS can be upper bounded by
\begin{eqnarray*}
    2C \left\| \frac{1}{\tau_i^0}\sum_{t=T-\min(\tau_i^0, \tau_i^*)+1 }^Tf^{'}(\triple, \theta_{k_i^0}^0) \right\|^2+O(1) \|\widehat{\theta}_{k_i^0}-\theta_{k_i^0}^0\|^2,
\end{eqnarray*}
where $O(1)$ denotes some positive constant. Similar to Step 1 of the proof, with probability $1-O(N^{-1}T^{-1})$, the first term of the above expression can be upper bounded by $T^{-1 } \log (NT)$ and the bound is uniform in $i$. Meanwhile, the second term can be upper bounded based on \eqref{eqn:step2assertion1}. As such, it follows from \eqref{eqn:step2assertion2} that 
\begin{eqnarray}\label{eqn:step2assertion3}
    \|\widehat{\theta}_{\widehat{k}_i}-\widehat{\theta}_{k_i^0}\|^2\le O(1)\max_i \frac{[\tau_i^0-\tau_i^*]_+^2}{(\tau_i^*)^2}+O(1)\frac{\log (NT)}{T},
\end{eqnarray}
for some positive constant $O(1)$. 

Given that $s_{cl}\gg \max_i (\tau_i^*)^{-1} (\tau_i^0-\tau_i^*)_++T^{-1/2}\sqrt{\log(NT)}$, according to \eqref{eqn:step2assertion1}, the difference $\|\widehat{\theta}_{k_1}-\widehat{\theta}_{k_2}\|$ is at least $s_{cl}/2$ whenever $k_1\neq k_2$. As such, \eqref{eqn:step2assertion3} holds only when $\widehat{k}_i=k_i^0$. This completes the proof for the second step.
\end{proof}

\subsubsection{Proof of Lemma \ref{lemma3}}\label{sec:proof:rateIC_old}
Before proving Lemma \ref{lemma3}, we remark that to guarantee the results in Lemma \ref{lemma3} hold for later iterations in addition to the first iteration, we require the estimated change point to converge at a rate of $T^{-1/2}\log(NT)$. However, this is achieved by the condition $s_{cp}\gg N^{-1/2}T^{-1/4}\log^{3/2}(NT)$ in Assumption \ref{as:signal}. Thus, under Assumption \ref{as:signal}, Lemma \ref{lemma3} implies the consistency of the proposed IC at every iteration, not just the first iteration. 
\begin{proof}
Firstly, consider the case where $K>K^0$ where $K^0$ is the true number of clusters. For a given $K$, we rewrite the clustering objective function as $Q(\theta, \mathcal{K}|K)$ and denote $(\widehat{\theta}(K), \widehat{\mathcal{K}}(K)) = \argmax Q(\theta, \mathcal{K}|K)$ to highlight their dependencies upon $K$.

Using similar arguments in the proof of Lemma \ref{lemma2}, we can show that $\widehat{\theta}_{K}$ also satisfies the rate of convergence in \eqref{eqn:step1assertion} with probability $1-O(N^{-1}T^{-1})$. Meanwhile, it is easy to show that the upper error bound therein can be refined by replacing the $\log(NT)$ term with on the RHS with $\log(1/\alpha)$ for any fixed $\alpha\in (0,1]$. However, this refinement comes at the cost of changing the high probability bound from the previous $1-O(N^{-1}T^{-1})$ to $1-\alpha-o(1)$.
Therefore, 
the difference between the proposed IC with $K^0$ clusters and that with $K$ many clusters equals
\begin{equation*}
\begin{aligned}
    &IC(\widehat{\theta}_{K^0}, \widehat{\mathcal{K}}_{K^0}) -IC(\widehat{\theta}_{K}, \widehat{\mathcal{K}}_{K})=N[Q(\widehat{\theta}_{K^0}, \widehat{\mathcal{K}}_{K^0}|K^0) - Q(\widehat{\theta}_{K }, \widehat{\mathcal{K}}_{K }|K)] + (K-K^0) \frac{N\log(NT)}{T}\\
    =&\cssumn[\tau_i^0][\min(\tau_i^*, \tau_i^0)] f^{'}(\triple, \theta_{k_i^0}^0)^\top (\widehat{\theta}_{\widehat{k}_{i}(K^0)}(K^0) - \widehat{\theta}_{\widehat{k}_{i}(K)}(K))\\
    +&\frac{1}{2}\cssumn[\tau_i^0][\min(\tau_i^*, \tau_i^0)](\theta^0_{k_i^0} - \widehat{\theta}_{\widehat{k}_i(K^0)}(K^0))^\top f^{''}(\triple, \theta_i^*(K^0)) (\theta^0_{k_i^0} - \widehat{\theta}_{\widehat{k}_i(K^0)}(K^0))\\ -&\frac{1}{2}\cssumn[\tau_i^0][\min(\tau_i^*, \tau_i^0)](\theta^0_{k_i^0} - \widehat{\theta}_{\widehat{k}_i(K)}(K))^\top f^{''}(\triple, \theta_i^*(K)) (\theta^0_{k_i^0} - \widehat{\theta}_{\widehat{k}_i(K)}(K))\\
    -&\cssum[\tau^0_i][\tau^0_i+1][-\min(\tau_i^*,\tau_i^0)][f(\triple,\widehat{\theta}_{\widehat{k}_i(K^0)}(K^0))-f(\triple,\widehat{\theta}_{\widehat{k}_{i}(K)}(K))]\mathbb{I}(\tau_i^0> \tau_i^*)\\
   + &(K-K^0)\frac{N\log (NT)}{T}, 
\end{aligned}
\end{equation*}
for some $\theta_i^*(K^0)$ and $\theta_i^*(K)$ that lie between the oracle parameter and the estimator. Below, we analyse the terms on the RHS one by one:
\begin{itemize}
    \item Using similar arguments to the proof of Lemma \ref{lemma2}, the second line can be lower bounded by $-O(T^{-1/2}\sqrt{\log(1/\alpha)})\sum_{i=1}^N \|\widehat{\theta}_{\widehat{k}_{i}(K^0)}(K^0) - \widehat{\theta}_{\widehat{k}_{i}(K)}(K)\|$, with probability $1-\alpha$. Based on the established convergence rates for $\sum_{i=1}^N \|\widehat{\theta}_{\widehat{k}_{i}(K^0)}(K^0)-\theta^0_{k_i^0}\|^2$ and $\sum_{i=1}^N \|\widehat{\theta}_{\widehat{k}_{i}(K)}(K)-\theta^0_{k_i^0}\|^2$, it follows from Cauchy-Schwarz inequality that the second line can be further lower bounded by
    \begin{eqnarray*}
        -O(1)\frac{N\log(1/\alpha)}{T}-O(1)\sum_{i=1}^N \frac{(\tau_i^0-\tau_i^*)_+^2}{(\tau_i^0)^2},
    \end{eqnarray*}
    with probability $1-\alpha$. 
    \item Similarly, based on the established convergence rates for $\sum_{i=1}^N \|\widehat{\theta}_{\widehat{k}_{i}(K^0)}(K^0)-\theta^0_{k_i^0}\|^2$, the third line can be lower bounded by
    \begin{eqnarray*}
        -O(1)\frac{N\log(1/\alpha)}{T}-O(1)\sum_{i=1}^N \frac{(\tau_i^0-\tau_i^*)_+^2}{(\tau_i^0)^2},
    \end{eqnarray*}
    with probability $1-\alpha$.
    \item Using similar arguments to Step 1 of the proof of Lemma \ref{lemma:consistency}, we can show that the minimum eigenvalues of the matrices $\{-(\tau_i^0)^{-1}\sum_{t=T-\min(\tau_i^*,\tau_i^0)+1}^T f^{''}(\triple,\bstheta_i^*(K))\}_i$ are positive semi-definite with probability at least $1-O(N^{-1}T^{-1})$. Consequently, the fourth line is non-negative with probability at least $1-O(N^{-1}T^{-1})$. 
    \item Similar to the proof of Lemma \ref{lemma2}, the fifth line can be lower bounded by $-[\max_i (\tau_i^0-\tau_i^*)_+/\tau_i^0] \sum_{i=1}^N \|\widehat{\theta}_{\widehat{k}_{i}(K^0)}(K^0) - \widehat{\theta}_{\widehat{k}_{i}(K)}(K)\|$, which, according to the convergence rates of $\widehat{\theta}_{\widehat{k}_{i}(K^0)}(K^0)$ and $\widehat{\theta}_{\widehat{k}_{i}(K)}(K)$, can be further lower bounded by
    \begin{eqnarray*}
        -O(1)\frac{N\log(1/\alpha)}{T}-O(1)\sum_{i=1}^N \frac{(\tau_i^0-\tau_i^*)_+^2}{(\tau_i^0)^2},
    \end{eqnarray*}
    with probability $1-\alpha$.
    \item Finally, the last line is lower bounded by $N\log(NT)/T$, as $K>K^0$. 
\end{itemize}
To summarize, we have shown that $IC(\widehat{\theta}_{K^0}, \widehat{\mathcal{K}}_{K^0}) -IC(\widehat{\theta}_{K},\widehat{\mathcal{K}}_{K})$ is lower bounded by
\begin{eqnarray*}
    \frac{N\log(NT)}{T}-O(1)\frac{N\log(1/\alpha)}{T}-O(1)\sum_{i=1}^N \frac{(\tau_i^0-\tau_i^*)_+^2}{(\tau_i^0)^2},
\end{eqnarray*}
with probability $1-\alpha-O(N^{-1}T^{-1})$. The above expression is strictly positive, under the given conditions on the initial change point estimator. This suggests that the proposed IC will not over-select the number of clusters with probability $1-\alpha-O(N^{-1}T^{-1})$.

Next, consider the case where $K<K^0$. We claim that there are $\Omega(N)$ many subjects being wrongly clustered into the same group. This is because 
as $K$ is smaller than $K^0$, there will be at least two true clusters, say the $k_1$-th and $k_2$-th clusters, with over $|\mathcal{C}_{k_1}|/K^0$ and $|\mathcal{C}_{k_2}|/K^0$ many subjects, respectively, being assigned to the same cluster. Under Assumption \ref{as:tau}, then number of subjects in this wrongly formed cluster is proportional to $N$. Let $\Tilde{\theta}$ denote the estimated parameter using data from this cluster. 
The difference between the proposed IC with $K^0$ clusters and that with $K$ clusters is given by
\begin{equation*}
\begin{aligned}
    &IC(\widehat{\theta}_{K^0}, \widehat{\mathcal{K}}_{K^0}) -IC(\widehat{\theta}_{K}, \widehat{\mathcal{K}}_{K})=N[Q(\widehat{\theta}_{K^0}, \widehat{\mathcal{K}}_{K^0}|K^0) - Q(\widehat{\theta}_{K }, \widehat{\mathcal{K}}_{K }|K)] +  (K-K^0) \frac{N\log(NT)}{T}\\
 =&\cssumn[\tau_i^0][\min(\tau_i^*, \tau_i^0)] f^{'}(\triple, \theta_{k_i^0}^0)^\top (\widehat{\theta}_{\widehat{k}_{i}(K^0)}(K^0) - \widehat{\theta}_{\widehat{k}_{i}(K)}(K))\\
    +&\frac{1}{2}\cssumn[\tau_i^0][\min(\tau_i^*, \tau_i^0)](\theta^0_{k_i^0} - \widehat{\theta}_{\widehat{k}_i(K^0)}(K^0))^\top f^{''}(\triple, \theta_i^*(K^0)) (\theta^0_{k_i^0} - \widehat{\theta}_{\widehat{k}_i(K^0)}(K^0))\\ -&\frac{1}{2}\cssumn[\tau_i^0][\min(\tau_i^*, \tau_i^0)](\theta^0_{k_i^0} - \widehat{\theta}_{\widehat{k}_i(K)}(K))^\top f^{''}(\triple, \theta_i^*(K)) (\theta^0_{k_i^0} - \widehat{\theta}_{\widehat{k}_i(K)}(K))\\
    -&\cssum[\tau^0_i][\tau^0_i+1][-\min(\tau_i^*,\tau_i^0)][f(\triple,\widehat{\theta}_{\widehat{k}_i(K^0)}(K^0))-f(\triple,\widehat{\theta}_{\widehat{k}_{i}(K)}(K))]\mathbb{I}(\tau_i^0> \tau_i^*)\\
   + &(K-K^0)\frac{N\log (NT)}{T}.
\end{aligned}
\end{equation*}
Again, we analyse the above expression line by line:
\begin{itemize}
	\item Similarly, the second line can be lower bounded by $-O(NT^{-1/2}\sqrt{\log(1/\alpha)})$, with probability $1-\alpha$;
	\item with probability $1-\alpha$, the third line is again lower bounded by
    \begin{eqnarray*}
        -O(1)\frac{N\log(1/\alpha)}{T}-O(1)\sum_{i=1}^N \frac{(\tau_i^0-\tau_i^*)_+^2}{(\tau_i^0)^2}.
    \end{eqnarray*}
	\item The fourth line is lower bounded by $c T^{-1}\sum_{i=1}^N \|\theta^0_{k_i^0} - \widehat{\theta}_{\widehat{k}_i(K)}(K)\|^2$ for some constant $c>0$ with probability at least $1-O(N^{-1}T^{-1})$. Considering the $\Omega(N)$ many subjects who originally belong to the $k_1$-th and $k_2$-th clusters but are wrongly clustered together, this term is at least $CN\|\theta^0_{k_1}-\theta^0_{k_2}\|^2\ge CNs_{cl}^2$. 
	\item The fifth line can be similarly lower bounded by $-O(1)N[\max_i (\tau_i^0-\tau_i^*)_+/\tau_i^0]$. 
	\item The last line is $-O(T^{-1} N\log (NT))$.
\end{itemize}
To summarize, we have shown that $IC(\widehat{\theta}_{K^0}, \widehat{\mathcal{K}}_{K^0}) -IC(\widehat{\theta}_{K},\widehat{\mathcal{K}}_{K})$ is lower bounded by
\begin{eqnarray*}
    CNs_{cl}^2-O(1)\frac{\sqrt{\log(1/\alpha)}}{\sqrt{T}}--O(1)N\max_i \frac{(\tau_i^0-\tau_i^*)_+}{\tau_i^0}.
\end{eqnarray*}
Under the given signal strength conditions in Assumption \ref{as:signal}, it is strictly positive, with probability at least $1-\alpha-O(N^{-1}T^{-1})$. 

Consequently, we have shown that the proposed IC is maximized at the true number of clusters $K^0$, with probability $1-\alpha-O(N^{-1}T^{-1})$. Since $\alpha$ can be made arbitrarily small, it follows that the proposed IC is consistent. This completes the proof. 

\end{proof}

\subsection{Proof of Corollary \ref{thmrate2}}
\begin{proof}
The proof of Corollary \ref{thmrate2} is straightforward. Based on the results in Theorem 4.1 and the condition on $N$, at each iteration, we can show that the change point detection error is smaller than $1/T$ with probability at least $1-O(N^{-1}T^{-1})$. Since the change point detection error can only take values $0$, $1/T$, $2/T$, etc., it implies that the change point detection error equals exactly $0$ with probability at least $1-O(N^{-1}T^{-1})$. The proof is hence completed. 
\end{proof}

\subsection{Proof of Theorem \ref{thmregret}}
Theorem \ref{thmregret} is concerned with the regrets of various estimated optimal policies. Below, we first derive the regret bound for the proposed algorithm. We next prove the inconsistencies of the Homogeneous, Stationary and Doubly Homogeneous algorithms. 
\subsubsection{The proposed algorithm}\label{thm:regretproposed}
\begin{proof}
The proof is very similar to that of \cite{chen2019information}, who established the regret bound of the FQI algorithm in standard doubly homogeneous environments. Consequently, we provide a sketch of the proof only, focusing on highlighting the difference from that of \cite{chen2019information}. 

Our proof is divided into three parts. In Part 1, we define another regret by assuming the transition never change after time $T$, and bound the difference between this regret and the original definition. In Part 2, we apply the performance difference lemma to upper bound our newly defined regret using the estimation error of the Q-function. Finally, in Part 3, we derive the Q-function's estimation error. 

\smallskip

\noindent \textbf{Part 1}. We begin with a new definition of the cumulative reward. Specifically, for a given policy $\pi$, let $\Mean_s^{\pi}$ denote the expectation by assuming the transition function $p_{i,t}$ remains stationary after time $T$. This allows us to define the following expected cumulative reward
\begin{eqnarray*}
    J_s(\pi)=\frac{1}{N}\sum_{i=1}^N \sum_{t=T+1}^{\infty} \gamma^{t-T-1} \Mean_s^{\pi} (R_{i,t}),
\end{eqnarray*}
and its associated regret $\sup_{\pi^*} J_s(\pi^*)-J_s(\pi)$. In this part, we focus on providing an upper bound for $\sup_{\pi^*} J(\pi^*)-J(\pi)-[\sup_{\pi^*} J_s(\pi^*)-J_s(\pi)]$. 

Recall that $T^*$ corresponds to the most recent change point after $T$. By definition, $\Mean_s^{\pi} (R_{i,t})$ is equal to $\Mean^{\pi} (R_{i,t})$ for any $T<t< T+T^*$. Let $\pi^{**}$ denote the argmax of $J(\pi^*)$, we have
\begin{eqnarray*}
    \sup_{\pi^*} J(\pi^*)-J(\pi)-[\sup_{\pi^*} J_s(\pi^*)-J_s(\pi)]\le J(\pi^{**})-J(\pi)-J_s(\pi^{**})+J_s(\pi)\\
    \le \sup_{\pi} \frac{2}{N}\sum_{i=1}^N \sum_{t=T^*}^{\infty} \gamma^{t-T-1} |\Mean_s^{\pi} (R_{i,t})-\Mean^{\pi} (R_{i,t})|\le 2R_{\max} \frac{\gamma^{T^*-T-1}}{1-\gamma},
\end{eqnarray*}
where the last inequality follows from the bounded reward assumption in Assumption \ref{as:reward}. Under the assumption that $T^*-T\gg \log(T)/\log(\gamma^{-1})$ and $N$ is at most proportional to $T R$, this term is of the order $T^{-C} R_{\max}/(1-\gamma)$, or equivalently $O(N^{-c}T^{-c} R_{\max}/(1-\gamma))$ for any sufficiently large constant $c>0$. Notice that this term is negligible as the constant $c$ can be made arbitrarily large. Without this condition, it will incur an additional term in the regret bound, given by
\begin{eqnarray*}
     \frac{2R_{\max}\Mean (\gamma^{T^*-T})}{\gamma(1-\gamma)}.
\end{eqnarray*}
Part 1 of the proof is thus completed. 

\smallskip

\noindent \textbf{Part 2}. Based on the results in Part 1, it suffices to upper bound the newly defined regret $\sup_{\pi^*} J_s(\pi^*)-J_s(\pi)$. Under LHE, $J_s(\pi)$ can be represented by $N^{-1}\sum_{k=1}^K |\mathcal{C}_k| J_{k,s}(\pi)$ where 
\begin{eqnarray*}
    J_{k,s}(\pi)=\sum_{t=T+1}^{\infty} \gamma^{t-T-1} \Mean_s^{\pi} (R_{i,t}),
\end{eqnarray*}
for any $i\in \mathcal{C}_k$. Since $K$ is finite, it suffices to show that for each $k$, the regret $\sup_{\pi^*} J_{s,k}(\pi^*)-J_{s,k}(\widehat{\pi})$ is of the order of magnitude specified in Theorem \ref{thmregret}. 

By the definition of $J_{s,k}$, this result can be established using the arguments from proofs in doubly homogeneous environments; see e.g., the proof of Theorem 11 of \citet{chen2019information}. We summarise the main steps below. 
\begin{enumerate}
    \item First, using the performance difference lemma \citep[see e.g., Lemma 13 of][]{chen2019information}, we can upper bound $\sup_{\pi^*} J_{s,k}(\pi^*)-J_{s,k}(\widehat{\pi})$ by $(1-\gamma)^{-1}[\|\widehat{Q}_{s,k}-Q_{s,k}^*\|_{\eta^{\widehat{\pi}}\times \widehat{\pi}}+\|\widehat{Q}_{s,k}-Q_{s,k}^*\|_{\eta^{\widehat{\pi}}\times \pi_s^*}]$ where $Q_{s,k}^*$ denotes the optimal Q-function for the $k$-th cluster assuming their transition function remains stationary after time point $T$, $\widehat{Q}_{s,k}$ denotes its estimator and for any policies $\pi_1,\pi_2$ and function $f(S,A)$, 
    \begin{eqnarray*}
        \|f\|_{\eta^{\pi_1}\times \pi_2}:=\sqrt{\Mean_{S\sim \eta^{\pi_1},A\sim \pi_2} f^2(S,A)},
    \end{eqnarray*}
    where the expectation $\Mean_{S\sim \eta^{\pi_1},A\sim \pi_2}$ is defined by assuming the state follows the discounted visitation distribution under $\pi_1$ and the action follows $\pi_2$.
    \item Next, notice that under Assumptions \ref{as:f'' contineous} and \ref{as:ergodic}, both the transition function and the behavior policy are bounded away from zero. This implies that Assumption 1 of \citet{chen2019information} is automatically satisfied with a finite concentratability coefficient. Now, using Lemmas 14 \& 15 and the proof of Theorem 11 of \citet{chen2019information}, we can further upper bound $(1-\gamma)^{-1}[\|\widehat{Q}_{s,k}-Q_{s,k}^*\|_{\eta^{\widehat{\pi}}\times \widehat{\pi}}+\|\widehat{Q}_{s,k}-Q_{s,k}^*\|_{\eta^{\widehat{\pi}}\times \pi_s^*}]$ by 
    \begin{eqnarray}\label{eqn:somebound}
        O\Big(\frac{\gamma^J R_{\max}}{(1-\gamma)^2}\Big)+O\Big(\frac{\displaystyle \max_{1\le j\le J} \|\widehat{Q}_{s,k}^{(j)}-\mathcal{B}_k^* \widehat{Q}_{s,k}^{(j-1)}\|_{\mu_k}}{(1-\gamma)^2}\Big),
    \end{eqnarray}
    where recall that $J$ denotes the number of iterations in FQI, and $\widehat{Q}_{s,k}^{(j)}$ denotes the estimated Q-function computed at the $j$th iteration. Under the condition $J\gg \log(NT)/\log(\gamma^{-1})$ in Assumption \ref{as:FQIiter}, the first term in \eqref{eqn:somebound} becomes negligible. It remains to upper bound the second term in \eqref{eqn:somebound}. We derive this upper bound in Part 3. 
\end{enumerate}

\smallskip

\noindent \textbf{Part 3}. As commented in Step 2 of the proof, we aim to upper bound $\max_{1\le j\le J} \|\widehat{Q}_{s,k}^{(j)}-\mathcal{B}_k^* \widehat{Q}_{s,k}^{(j-1)}\|_{\mu_k}$ in this step. The proof is again, similar to that of Lemma 16 of \citet{chen2019information}. Below, we focus on highlighting their differences. 

For any functions $Q_1,Q_2,Q_3\in \mathcal{Q}$, define a function $g(s,a,r,s';Q_1,Q_2,Q_3)=[r+\gamma \max_a \gamma Q_1(s',a)-Q_2(s,a)]^2-[r+\gamma \max_a \gamma Q_1(s',a)-Q_3(s,a)]^2$. 
A key step in the proof is to establish a uniform upper bound the following empirical process
\begin{eqnarray}\label{eqn:emp}
\begin{split}
    \frac{1}{|\widehat{\mathcal{C}}_k| \widehat{\tau}^{(k)}} \sum_{i\in \widehat{\mathcal{C}}_k}\sum_{t=T-\widehat{\tau}^{(k)}}^{T-1}  \Big[g(S_{i,t},A_{i,t},R_{i,t},S_{i,t+1};Q_1,Q_2,Q_3)\\-\Mean [g(S_{i,t},A_{i,t},R_{i,t},S_{i,t+1};Q_1,Q_2,Q_3)]\Big],
\end{split}
\end{eqnarray}
indexed by $g\in \mathcal{G}=\{g:Q_1,Q_2,Q_3\in \mathcal{Q}\}$. 

We next summarise the differences between our setting and the setting considered in Lemma 16 of \citet{chen2019information}:
\begin{enumerate}
    \item  The error bound in  \citet{chen2019information} is derived in stationary environments. To the contrary, we consider potentially non-stationary environments where the transition function can be non-stationary when $\widehat{\tau}^{(k)}$ over-estimates its oracle value $\tau^{(k)}$.
    \item \citet{chen2019information} imposed a finite hypothesis class assumption on $\mathcal{Q}$ whereas our VC-class condition in Assumption \ref{as:VCclass} is more reasonable as it allows $Q$ to be an infinite hypothesis class. 
    \item \citet{chen2019information} required the state-action-reward-next-state tuples to i.i.d. whereas we consider the more realistic setting by taking their temporal dependence into account.
\end{enumerate}
To handle non-stationary environments, we decompose \eqref{eqn:emp} into two terms, given by
\begin{eqnarray*}
    \frac{1}{|\widehat{\mathcal{C}}_k| \widehat{\tau}^{(k)}} \sum_{i\in \widehat{\mathcal{C}}_k}\sum_{t=T-\widehat{\tau}^{(k)}}^{T-\min(\widehat{\tau}^{(k)},\tau^{(k)})}  \Big[g(S_{i,t},A_{i,t},R_{i,t},S_{i,t+1};Q_1,Q_2,Q_3)\\-\Mean [g(S_{i,t},A_{i,t},R_{i,t},S_{i,t+1};Q_1,Q_2,Q_3)|S_{i,t},A_{i,t}]\Big]
\end{eqnarray*}
and
\begin{eqnarray}\label{eqn:empsecondterm}
\begin{split}
    \frac{1}{|\widehat{\mathcal{C}}_k| \widehat{\tau}^{(k)}} \sum_{i\in \widehat{\mathcal{C}}_k}\sum_{t=T-\min(\widehat{\tau}^{(k)},\tau^{(k)})}^{T-1}  \Big[g(S_{i,t},A_{i,t},R_{i,t},S_{i,t+1};Q_1,Q_2,Q_3)\\-\Mean [g(S_{i,t},A_{i,t},R_{i,t},S_{i,t+1};Q_1,Q_2,Q_3)|S_{i,t},A_{i,t}]\Big].
\end{split}
\end{eqnarray}
Under Assumptions \ref{as:tau} and \ref{as:VCclass}, it follows from the conclusions in Theorem \ref{thmrate} that the first term can be upper bounded by
\begin{eqnarray}\label{eqn:firstermorder}
    O\Big(\frac{\log^3(NT)R_{\max}^2}{NTs_{cp}^2(1-\gamma)^2}\Big),
\end{eqnarray}
with probability at least $1-O(N^{-1}T^{-1})$. 

To handle infinite hypothesis classes, we notice that under Assumption \ref{as:VCclass}, the composite function $g$ is Lipschitz as a function of $Q_1$, $Q_2$ and $Q_3$, with the Lipschitz constant upper bounded by $O(R_{\max}/(1-\gamma))$. According to e.g., Lemma A.6 of \citet{cherno2014}, the function class $\mathcal{G}$ belongs to the VC type class as well, with the envelop function upper bounded by $O(R_{\max}^2/(1-\gamma)^2)$. Consequently, we can find an $\epsilon$-net of $\mathcal{G}$, denoted by $\mathcal{G}_0$, with $\epsilon$ proportional to $(NT)^{-1}R_{\max}^2/(1-\gamma)^2$, such that restricting to the finite hypothesis class $\mathcal{G}_0$ provides an reasonable approximation for $\mathcal{G}$ with the approximation error upper bounded by 
\begin{eqnarray}\label{eqn:approx}
    O\Big(\frac{R_{\max}^2}{(1-\gamma)^2NT}\Big).
\end{eqnarray} 
Meanwhile, the number of elements in $\mathcal{G}_0$ is of the order $O(N^V T^V)$ for some constant $V>0$. 

Finally, to handle non i.i.d data, we notice that the sequence $\{S_{i,t}\}_{t>T}$ is exponentially $\beta$-mixing; see Step 1 of the proof of Lemma \ref{lemma:consistency}. Consequently, we can invoke the Bernstein's inequality designed for martingales \citep[see e.g.,][]{dzhaparidze2001bernstein} and exponentially $\beta$-mixing time series \citep[Theorem 4.2]{ChenChristensen2015} to obtain a uniform upper bound for \eqref{eqn:emp}, when restricting to the class of functions $g\in \mathcal{G}_0$. Other arguments are similar to those in the proof of Lemma 16 of \citet{chen2019information} and the proof of Step 1 of Lemma \ref{lemma:consistency}. More specifically, it can be shown that with probability at least $1-O(N^{-1}T^{-1})$, \eqref{eqn:empsecondterm} can be upper bounded by
\begin{eqnarray}\label{eqn:esterror}
    O\Big(\frac{R_{\max}^2\log^2(NT)}{(1-\gamma)^2NT}\Big)+O\Big(\frac{R_{\max}\log(NT)\|Q_2-Q_3\|_{\mu_k}}{(1-\gamma)\sqrt{NT}}\Big),
\end{eqnarray}
uniformly in $g\in \mathcal{G}_0$. This together with \eqref{eqn:firstermorder} and \eqref{eqn:approx} yields that, when considering the unrestricted function class $\mathcal{G}$, \eqref{eqn:emp} can be upper bounded by
\begin{eqnarray}\label{eqn:esterrorfinal}
    O\Big(\frac{\log^3(NT)R_{\max}^2}{NTs_{cp}^2(1-\gamma)^2}\Big)+O\Big(\frac{R_{\max}^2\log^2(NT)}{(1-\gamma)^2NT}\Big)+O\Big(\frac{R_{\max}\log(NT)\|Q_2-Q_3\|_{\mu_k}}{(1-\gamma)\sqrt{NT}}\Big),
\end{eqnarray}
with probability at least $1-O(N^{-1}T^{-1})$. 

Under the completeness assumption in Assumption \ref{as:complete}, we have $\mathcal{B}_k^* \widehat{Q}_{s,k}^{(j-1)}\in \mathcal{Q}$ for any $j$. Let $Q_1$ denote $\widehat{Q}_{s,k}^{(j-1)}$, $Q_2$ denote $\widehat{Q}_{s,k}^{(j)}$ and $Q_3$ denote $\mathcal{B}_k^*\widehat{Q}_{s,k}^{(j-1)}$. Since $\widehat{Q}_{s,k}^{(j)}$ is the empirical risk minimiser, it follows that
\begin{eqnarray*}
    &&\frac{1}{|\widehat{\mathcal{C}}_k| \widehat{\tau}^{(k)}} \sum_{i\in \widehat{\mathcal{C}}_k}\sum_{t=T-\widehat{\tau}^{(k)}}^{T-1} \Big[R_{i,t}+\gamma \max_a \widehat{Q}_{s,k}^{(j-1)}(S_{i,t+1},a)-\widehat{Q}_{s,k}^{(j)}(S_{i,t},A_{i,t})\Big]^2\\
    &\le& \frac{1}{|\widehat{\mathcal{C}}_k| \widehat{\tau}^{(k)}} \sum_{i\in \widehat{\mathcal{C}}_k}\sum_{t=T-\widehat{\tau}^{(k)}}^{T-1} \Big[R_{i,t}+\gamma \max_a \widehat{Q}_{s,k}^{(j-1)}(S_{i,t+1},a)-\mathcal{B}_k^*\widehat{Q}_{s,k}^{(j-1)}(S_{i,t},A_{i,t})\Big]^2.
\end{eqnarray*}
This together with the established uniform upper error bound implies that 
\begin{eqnarray*}
    -\frac{1}{|\widehat{\mathcal{C}}_k| \widehat{\tau}^{(k)}} \sum_{i\in \widehat{\mathcal{C}}_k}\sum_{t=T-\widehat{\tau}^{(k)}}^{T-1}  g_{i,t}^*(\widehat{Q}_{s,k}^{(j-1)},\widehat{Q}_{s,k}^{(j)},\mathcal{B}_k^*\widehat{Q}_{s,k}^{(j-1)})
\end{eqnarray*}
where $g_{i,t}^*(Q_1,Q_2,Q_3)=\Mean [g(S_{i,t},A_{i,t},R_{i,t},S_{i,t+1};Q_1,Q_2,Q_3)]$, 
is upper bounded by \eqref{eqn:esterrorfinal}, with probability at least $1-O(N^{-1}T^{-1})$.

The above expression can be decomposed into two terms, given by 
\begin{eqnarray*}
    -\frac{1}{|\widehat{\mathcal{C}}_k| \widehat{\tau}^{(k)}} \sum_{i\in \widehat{\mathcal{C}}_k}\sum_{t=T-\widehat{\tau}^{(k)}}^{T-\min(\tau^{(k)},\widehat{\tau}^{(k)})}  g_{i,t}^*(\widehat{Q}_{s,k}^{(j-1)},\widehat{Q}_{s,k}^{(j)},\mathcal{B}_k^*\widehat{Q}_{s,k}^{(j-1)})
\end{eqnarray*}
and 
\begin{eqnarray*}
    -\frac{1}{|\widehat{\mathcal{C}}_k| \widehat{\tau}^{(k)}} \sum_{i\in \widehat{\mathcal{C}}_k}\sum_{t=T-\min(\tau^{(k)},\widehat{\tau}^{(k)})}^{T-1} g_{i,t}^*(\widehat{Q}_{s,k}^{(j-1)},\widehat{Q}_{s,k}^{(j)},\mathcal{B}_k^*\widehat{Q}_{s,k}^{(j-1)}).
\end{eqnarray*}
Under similar arguments in proving \eqref{eqn:firstermorder}, the first term can be lower bounded by
\begin{eqnarray}\label{eqn:lowerbound}
    -O\Big(\frac{\log^3(NT)R_{\max}^2}{NTs_{cp}^2(1-\gamma)^2}\Big).
\end{eqnarray}
As for the second term, notice that $\mathcal{B}_k^*\widehat{Q}_{s,k}^{(j-1)}$ is the argmin of 
$g_{i,t}^*(\widehat{Q}_{s,k}^{(j-1)},\widehat{Q}_{s,k}^{(j)},\bullet)$ 
whenever $i\in \mathcal{C}_k$ and $t\ge \tau^{(k)}$. According to Theorem \ref{thmrate}, we have $\widehat{\mathcal{C}}_k=\mathcal{C}_k$ with probability at least $1-O(N^{-1}T^{-1})$. It follows that the second term can be represented by
\begin{eqnarray*}
    \frac{1}{|\mathcal{C}_k| \widehat{\tau}^{(k)}} \sum_{i\in \mathcal{C}_k}\sum_{t=T-\min(\tau^{(k)},\widehat{\tau}^{(k)})}^{T-1} \Mean \Big[Q_3(S_{i,t},A_{i,t})-Q_2(S_{i,t},A_{i,t})\Big]^2,
\end{eqnarray*}
with $Q_2$ being $\widehat{Q}_{s,k}^{(j)}$ and $Q_3$ being $\mathcal{B}_k^*\widehat{Q}_{s,k}^{(j-1)}$. Under the boundedness assumption on the transition function in Assumption \ref{as:f'' contineous}, the density function $S_{i,t}$ is bounded away from zero. It follows from the change of measure theorem that the second term can be lower bounded by $c\|\widehat{Q}_{s,k}^{(j)}-\mathcal{B}_k^*\widehat{Q}_{s,k}^{(j-1)}\|_{\mu_k}^2$ for some constant $c>0$. 

Combining these results together with \eqref{eqn:esterrorfinal} yields that 
\begin{eqnarray*}
    c\|\widehat{Q}_{s,k}^{(j)}-\mathcal{B}_k^*\widehat{Q}_{s,k}^{(j-1)}\|_{\mu_k}^2\le \frac{CR_{\max}^2\log^{3}(NT)}{(1-\gamma)^2 NTs^2_{cp}} +\frac{CR_{\max}\log(NT)\|\widehat{Q}_{s,k}^{(j)}-\mathcal{B}_k^*\widehat{Q}_{s,k}^{(j-1)}\|_{\mu_k}}{(1-\gamma)\sqrt{NT}},
\end{eqnarray*}
for some constant $C>0$. Using Cauchy-Schwarz inequality, the last term can be upper bounded by
\begin{eqnarray*}
    \frac{c\|\widehat{Q}_{s,k}^{(j)}-\mathcal{B}_k^*\widehat{Q}_{s,k}^{(j-1)}\|_{\mu_k}^2}{2}+\frac{C^2 R_{\max}^2\log^2(NT)}{2c(1-\gamma)^2NT}.
\end{eqnarray*}
Consequently, $\|\widehat{Q}_{s,k}^{(j)}-\mathcal{B}_k^*\widehat{Q}_{s,k}^{(j-1)}\|_{\mu_k}$ can be upper bounded by
\begin{eqnarray*}
    O\Big(\frac{R_{\max}\log^{3/2}(NT)}{(1-\gamma)\sqrt{NT}s_{cp}}\Big).
\end{eqnarray*}
This together with \eqref{eqn:somebound} yields the desired upper error bound. 

Finally, under the assumption that $N\gg s_{cp}^{-2}\log^3(NT)$, the change point error equals exactly zero, as proven in Corollary \ref{thmrate2}. In this case, both the first term in \eqref{eqn:esterrorfinal} and the lower bound in \eqref{eqn:lowerbound} become zero. Using the same arguments, it can be shown that $\|\widehat{Q}_{s,k}^{(j)}-\mathcal{B}_k^*\widehat{Q}_{s,k}^{(j-1)}\|_{\mu_k}$ can be upper bounded by
\begin{eqnarray*}
    O\Big(\frac{R_{\max}\log(NT)}{(1-\gamma)\sqrt{NT}}\Big),
\end{eqnarray*}
instead, with probability at least $1-O(N^{-1}T^{-1})$. The proof is hence completed. 
\end{proof}
\subsection{Inconsistencies of the Homogeneous and Doubly Homogeneous algorithms}\label{sec:MDPinconsistent}
Consider an MDP where the states are i.i.d. over time and population, independent of the rewards and actions. Consider settings with binary reward. Assume there exist two subgroups. For subjects belonging to the first subgroup, their reward equals 2 if they select action 1, and 0 otherwise. For those in the second subgroup, their reward equals 1 if they select action 0, and 0 otherwise. By definition, it is immediate to see that the optimal policy will assign all subjects within the first subgroup action 1, and all subjects within the second subgroup action 0. This yields an expected cumulative reward of
\begin{eqnarray*}
    \frac{1}{2}\sum_{t=0}^{\infty} \gamma^t \times 2+\frac{1}{2}\sum_{t=0}^{\infty} \gamma^t \times 1=\frac{3}{2(1-\gamma)}. 
\end{eqnarray*}
Since the transition function does not change, no change point will be identified by the Homogeneous algorithm. As such, both the Homogeneous and the Doubly Homogeneous algorithms will apply FQI to learn the optimal policy based on the entire offline dataset. Suppose the initial Q-function is a zero function. At the first iteration, its learned Q-function will converge to 
\begin{eqnarray*}
    Q^{(1)}(a,s)=\left\{\begin{array}{ll}
        \frac{1}{2}\times 2+\frac{1}{2}\times 0=1 & a=1 \\
        \frac{1}{2}\times 1+\frac{1}{2}\times 0=\frac{1}{2} & a=0
    \end{array}
    \right.
\end{eqnarray*}
as either $N$ or $T\to \infty$. At the second iteration, its learned Q-function will converge to 
\begin{eqnarray*}
    Q^{(2)}(a,s)=\left\{\begin{array}{ll}
        \frac{1}{2}\times 2+\frac{1}{2}\times 0+\gamma\max(1,1/2)=1+\gamma & a=1 \\
        \frac{1}{2}\times 1+\frac{1}{2}\times 0+\gamma\max(1,1/2)=\frac{1}{2}+\gamma & a=0
    \end{array}
    \right.
\end{eqnarray*}
Following the same logic, it can be shown that the learned optimal Q-function will converge to 
\begin{eqnarray*}
    \left\{\begin{array}{ll}
        \frac{1}{1-\gamma} & a=1 \\
        \frac{1}{1-\gamma}-\frac{1}{2} & a=0
    \end{array}\right.
\end{eqnarray*}
Consequently, the learned policy will assign action 1 to all subjects, as it ignores heterogeneity over population. Apparently, this policy is sub-optimal, which would incur a regret of
\begin{eqnarray*}
    \frac{1}{2}\sum_{t=0}^{\infty} \gamma^t \times 1-\frac{1}{2}\sum_{t=0}^{\infty} \gamma^t \times 0=\frac{1}{2(1-\gamma)}.
\end{eqnarray*}
This completes the proof. 
\subsection{Inconsistency of the Stationary algorithm}
The MDP can be similarly constructed to that in Section \ref{sec:MDPinconsistent}. Specifically, we assume the action is binary and all states are i.i.d. over time and population. Additionally, suppose all subjects share the same transition function. However, there exists a single change point at location $T/2+1$. Specifically, prior to the change, the reward equals $2$ if action 1 is selected, and $0$ otherwise.  After the change, the reward equals $1$ if action 0 is selected, and $0$ otherwise. 

Suppose there is no change point after time $T$. Then the optimal policy will assign action 0 to all subjects, leading to an expected cumulative reward of 
\begin{eqnarray*}
    \sum_{t=0}^{\infty} \gamma^t=\frac{1}{1-\gamma}. 
\end{eqnarray*}
Since all subjects share the same data distribution, the Stationary algorithm will only identify one cluster in the offline data. Similar to Section \ref{sec:MDPinconsistent}, the  optimal policy selected by FQI will assign action 0 to all subjects. Again, this policy is apparently sub-optimal, with a regret of $(1-\gamma)^{-1}$. This completes the proof. 
\section{Semi-synthetic data simulation} \label{sec:sim:real:additional}



\subsection{Offline Data Generating Process} \label{sec:sim:real:setting}
The semi-synthetic data setup is designed based on the analysis of IHS conducted in Section \ref{sec:realdata},
which identified two distinct clusters of interns and associated change points
at $T_{train}=82$. 
At each time point $t = 0, \ldots, T$, the binary action for the $i$-th individual is randomly generated with $P(A_{i,t} = 1) = 1 - P(A_{i,t} = -1) = 0.5$. The state vector $S_{i,t}$ comprises three variables. The state variables are initiated at $t = 0$ as independent normal distributions with $S_{i,0,1} \sim \cN(0, 1)$ for $i=1,2,3$.
The $k$th transition dynamic takes this form:
$S_{i,t+1} = \left(\bB + \delta \bG_{k}\right) (A_{i,t}, S_{i,t}^\top, A_{i,t}*S_{i,t}^\top)^\top + \epsilon_{i,t}$. Here, $\bB$ represents the effect of the current state on the next state in a reference dynamic (in our example, the $\mathcal{P}_1$), $\bG_{k}$ represents the difference between the effect of the current state in dynamic $k$ and the reference dynamic,
and $\epsilon_{i,t} \sim \cN_3(0, \mathrm{diag}(0.25,0.25,0.25))$. Here, $\mathrm{diag}(0.25,0.25,0.25)$ denotes a diagonal matrix with diagonal elements 0.25, 0.25, and 0.25. In addition, $\delta \in \{2, 1, 0.6\}$ is a factor that controls strong, moderate, and weak signal in the change of transition dynamics, respectively.

The base transition matrix is 
\begin{equation*}\mathbf{B}=
\begin{pmatrix}
  0.107 &  0.005 &  0.372 &  0.025 & -0.002 & -0.005 &  0.013 &  0.028 \\
 -0.099 & -0.014 &  0.038 &  0.1   &  0.005 & -0.006 & -0.007 &  0.008 \\
  0.005 & -0.007 &  0.002 & -0.015 &  0.475 & -0.002 &  0.005 & -0.002
\end{pmatrix}
\end{equation*}
and 
\begin{equation*}
\mathbf{G}_{2}=
\begin{pmatrix}
  0.048 &  0.015 &  0.053 &  0.037 & -0.000 & -0.036 & -0.048 & -0.013 \\
 -0.127 &  0.030 &  0.085 & -0.039 &  0.016 &  0.024 &  0.024 &  0.018 \\
 -0.033 &  0.021 &  0.037 & -0.004 & -0.012 & -0.038 & -0.000 &  0.041
\end{pmatrix},
\end{equation*}

\begin{equation*}\mathbf{G}_3=
\begin{pmatrix}
 -0.234 & -0.010 & -0.094 & -0.011 &  0.002 &  0.021 & -0.044 & -0.038 \\
  0.198 &  0.010 & -0.026 & -0.001 &  0.019 & -0.005 &  0.037 & -0.023 \\
 -0.011 &  0.006 &  0.000 & -0.001 &  0.053 & -0.016 &  0.006 &  0.008
\end{pmatrix},
\end{equation*}

\begin{equation*}
\mathbf{G}_4=
    \begin{pmatrix}
 -0.358 & -0.092 & -0.187 &  0.012 &  0.003 &  0.020 &  0.112 & -0.042 \\
  0.427 &  0.058 &  0.018 & -0.035 & -0.004 &  0.007 & -0.143 & -0.006 \\
  0.044 & -0.022 & -0.006 & -0.023 &  0.035 &  0.007 &  0.040 &  0.017
\end{pmatrix}.
\end{equation*}

The transition functions for all subjects of the are specified as the followings.
For the first 20 subjects, they follow $\mathcal{P}_1$ from $t=0$ to $t=49$ and switch to $\mathcal{P}_2$ after $t=49$:
$$
S_{i,t+1} = \left\{
\begin{aligned}
    & \mathbf{B}(A_{i,t}, S_{i,t}^\top, A_{i,t}*S_{i,t}^\top)^\top + \epsilon_{i,t} \quad \text{if}\ t \in [0, 49]\\
    & (\mathbf{B}+\delta\mathbf{G}_2)(A_{i,t}, S_{i,t}^\top, A_{i,t}*S_{i,t}^\top)^\top + \epsilon_{i,t} \quad \text{if}\ t \in [50, 99]
\end{aligned}
\right.
$$
Subjects 21 to 40 follow $\mathcal{P}_3$ from $t=0$ to $t=49$ and switch to $\mathcal{P}_4$ after $t=49$: 
$$
S_{i,t+1} = \left\{
\begin{aligned}
    & (\mathbf{B}+\delta\mathbf{G}_3)(A_{i,t}, S_{i,t}^\top, A_{i,t}*S_{i,t}^\top)^\top + \epsilon_{i,t} \quad \text{if}\ t \in [0, 49]\\
    & (\mathbf{B}+\delta\mathbf{G}_4)(A_{i,t}, S_{i,t}^\top, A_{i,t}*S_{i,t}^\top)^\top + \epsilon_{i,t} \quad \text{if}\ t \in [50, 99]
\end{aligned}
\right.
$$
The remaining 20 subjects follow $\mathcal{P}_4$ throughout the entire period:
$$
S_{i,t+1} = (\mathbf{B}+\mathbf{G}_4)(A_{i,t}, S_{i,t}^\top, A_{i,t}*S_{i,t}^\top)^\top + \epsilon_{i,t} \quad \text{for}\ t \in [0, 99]
$$

\subsection{Implementation of change point detection and clustering}\label{sec:sim:implementation:cp}

To construct the log-likelihood ratio test (LRT) statistic proposed in Section \ref{sec:method}, we need to estimate the conditional distribution $p(\triple)$.
We parameterised $p(\triple)$ using a Gaussian distribution, by fitting linear regression model $S_{i,t+1} = \bbeta \bm X_{i,t} + \epsilon_{i,t}$, where $\bm X_{i,t} = (A_{i,t},S_{i,t}^\top,   A_{i,t}S_{i,t}^\top)^\top$ consists of the state, action, and their interaction, and $\bbeta$ is the regression coefficient. The variance of the residual $\epsilon_{i,t}$ is assumed constant in time. 

When calculating the rejection threshold of the LRT, we noticed that the asymptotic distribution of the CUSUM statistic depends on the data generating process only through its degree of freedom, thanks to the Markov assumption. As such, to approximate the threshold, we first sampled i.i.d. $d$-dimensional standard multivariate Gaussian vectors $\{Z_{i,t}\}_{i\in \widehat{\mathcal{C}}_k, \tau\le t\le T-1}$, where $d$ equals the degrees of freedom, or equivalently the number of nonzero coefficients in $\bbeta$. For a given $\tau$, we next computed the maximum of the squared weighted Euclidean distances: $\max \left\{ (\bar{Z}_{[T-\tau, u)} - \bar{Z}_{[u, T]} )^2; T - \tau < u < T \right\}$, where $\bar{Z}_{I}$ is the sample mean of $Z_{i,t}$'s on the time interval $I$ over all $i$'s. 
We repeated this process to generate 10000 samples of the asymptotic reference distribution. We then used the 0.01 empirical upper quantile of the distribution as the rejection threshold. For theoretical analysis, we set the threshold to be proportional to $\bar{C}\log^2(NT)\log(\log(NT))$ for some sufficiently constant $\bar{C}>0$ so that the resulting test is consistent theoretically. 

To implement the clustering algorithm, we adopt the proposal by \cite{ECTA11319} which minimises a least square objective function. 
Given a set of initial change points and a number of clusters, we iterate between the two subroutines twice to update the estimated change points and cluster membership.

\subsection{Implementation Details of Online Evaluation}\label{sec:sim:real:evaluation:additional}

To estimate the optimal policy, we coupled FQI with decision tree regression to compute the Q-estimator. The discounted factor $\gamma$ is set to be $0.9$. The hyperparameters in the decision tree model, the maximum tree depth and the minimum number of samples on each leaf node, were selected using 5-fold cross validation from $\{3,5,6\}$ and $\{50,60,80\}$, respectively. 
We considered an online setting and simulated potential change points start from $t = 100$ from a Poisson process with rate $1/60$. Accordingly, a new change point was expected to occur every 40 time points. We set the terminal time to 400, yielding 4 to 5 potential change points in most replications. 

The online data were simulated in the following manner. We consider the most recent two transition dynamics described above with 
$S' = \left( \bB + \bG_{k} \right) (A_{i,t}, S_{i,t}^\top, A_{i,t}*S_{i,t}^\top)^\top  + \epsilon_{i,t}$ for $k \in\{2, 4\}$. 
The subjects $i\in [1,20]$, $i\in [21,40]$ and $i\in[41,60]$ are considered to belong to three underlying clusters within which all cluster members share the same dynamic at any time during the whole online data generating process. Whenever a new change point occurs in a cluster, the first two underlying clusters separately decide to change its transition dynamics  change into another dynamics with probability 0.5, or otherwise remain constant. The last underlying cluster adopt dynamic $\mathcal{P}_2$ for all time.
This yields a total of five possible scenarios, namely, merge, split, switch, evolution and constancy (see Figure \ref{fig:dynamics_illustration} in the main text). 
In addition, the reward function is assumed doubly homogeneous and equals $R_{i,t} = S_{i,t+1,1}$, i.e., the first dimension of the state variable for all individuals and time points.

Finally, we assumed that online data came in batches regularly at every $L = 25$ time points starting from $t = 100$. This yielded a total of $12$ batches of data. The first data batch was generated according to
the estimated optimal policy $\widehat{\pi}_k$ computed based on the data subset $\{O_{i,t}:i\in\widehat{\cC}_k, t\in[ T-\widehat{\tau}_{i}^*,T]\}, k=1,\dots,K$.
Let $T^*_0=\max_i T-\widehat{\tau}^*_i$. 
Suppose $b$ batches of data were received. We first applied the proposed change point detection and clustering method on the data subset in $[T^*_{b-1}, T + bL]$ to identify new change points and clusters. If there was at least one change point, we set $T^*_b = \max_i \{T-\widehat{\tau}^*_{i,b}\}$. If no changes were detected, we set $T^*_b = T^*_{b-1}$. We next updated the optimal policy using the data subset $\{O_{i,t}: i \in \widehat{\cC}_k, t \in [ T-\widehat{\tau}^{*}_{i,b}, T +bL] \}, k=1, \dots, K$ for all the current clusters and used the updated optimal policy (combined with the $\epsilon$-greedy algorithm) to generate the
$(b +1)$-th data batch. We repeated this procedure until all $12$ batches of data were received. Finally, we computed the average rewards obtained from time 100 to 400 for the 60 subjects. 

\section{Additional implementation details for analysing the IHS dataset}\label{sec:sim:realIHS2020}
All data is standardized to have a mean of zero and unit variance before training. The implementation of the proposed algorithm and fitted-Q iterations follows the specifications in Section \ref{sec:sim:real:additional}. For off-policy evaluation (OPE), we assume that cluster membership in the testing set remains the same as detected by the proposed method in the training set, and there is no change point in the testing set. Therefore, we conduct separate OPE for each cluster in the testing set and report the average OPE result, weighted by the number of subjects in each cluster. We implement the fitted-Q evaluation algorithm using decision tree regression to compute the Q-estimator. The hyperparameters and tuning method are the same as those used for the FQI in Section \ref{sec:sim:real:additional}. 
The average value reported in Table \ref{tab:realdata_opevalues} is calculated by multiplying the discounted cumulative step counts by $1-\gamma=0.1$.

\end{document}